%% file: main.tex
\documentclass{article} 
\usepackage{in_token_report,times}

\input{math_commands.tex}

\usepackage{hyperref}
\hypersetup{pdftitle={In-Token Learning for High-Fidelity Image Restoration via Diffusion Transformers},pdfauthor={Xingfu Yi and Xiaoxue Yu},hidelinks}
\usepackage{url}

\usepackage{booktabs,multirow,adjustbox,siunitx,makecell}

\newcolumntype{C}[1]{>{\centering\arraybackslash}m{#1}}

\usepackage{siunitx}
\usepackage{booktabs,adjustbox}
\usepackage{pifont}
\newcommand{\cmark}{\ding{51}} 
\newcommand{\xmark}{\ding{55}} 
\usepackage{caption}

\usepackage{graphicx}
\usepackage{subcaption}
\newcommand{\best}[1]{\textcolor{red}{\textbf{#1}}}     
\newcommand{\secbest}[1]{\underline{\textcolor{blue}{\textbf{#1}}}}  

\renewcommand{\best}[1]{\textcolor{red}{\bfseries \num{#1}}} 
\renewcommand{\secbest}[1]{\textcolor{blue}{\bfseries\underline{\num{#1}}}} 
\newcommand{\mch}[1]{\multicolumn{1}{c}{#1}} 

\usepackage[table]{xcolor}
\renewcommand{\best}[1]{{\bfseries\underline{\num{#1}}}}  
\renewcommand{\secbest}[1]{{\bfseries \num{#1}}}          

\title{In-Token Learning for High-Fidelity Image Restoration via Diffusion Transformers}

\author{\textbf{Xingfu Yi}$^{1}$ \quad \textbf{Xiaoxue Yu}$^{2}$\\[3pt]
  $^{1}$Independent Researcher, Hangzhou, China\\
  $^{2}$Zhejiang University, Hangzhou, China\\
  \texttt{yixingfu.research@gmail.com}}

\begin{document}

\maketitle

{\small\noindent\textbf{Author note.} This technical report preserves the earlier, broader study that preceded \emph{Fill2SR}, including super-resolution, denoising, and automatic colorization experiments. Fill2SR subsequently developed the real-world super-resolution direction. Section~\ref{sec:relation} explains the shared foundations and subsequent developments.\par}

\begin{abstract}
We present \textbf{In-Token Learning}, an image restoration framework that adapts a pretrained diffusion transformer using conditional rectified flow matching. Clean targets paired with degraded inputs supervise transport from Gaussian noise to restored images. Spatially aligned degraded-image tokens are fused with evolving latent tokens along the channel dimension, preserving the image-token count at a given resolution. \textbf{Direct Low-Quality Guidance (DLG)} combines frozen degraded-image embeddings with a fixed task prompt through the native conditioning pathway, without a trainable ControlNet-style branch or image captioning.
We evaluate super-resolution and denoising on DIV2K, LSDIR, FFHQ, RealLQ250, and RealPhoto60, and automatic colorization on DIV2K and LSDIR. The tasks use separately trained checkpoints under the same framework. Results show competitive fidelity and perceptual quality under the evaluated protocols, with weaker generalization on RealLQ250. We report full-image QHD ($2560{\times}1440$) inference and a tiled $12$K restoration demonstration of \textit{Along the River During the Qingming Festival}. Attention cost still increases with resolution. This technical report preserves the early broader study underlying Fill2SR, which subsequently developed the real-world super-resolution direction.
\end{abstract}

\section{Introduction}

Recent advances in diffusion models~\cite{DDPM,SD,SDXL} have significantly improved image restoration quality, especially in super-resolution (SR) and real-world deblurring. Diffusion Transformers (DiTs)~\cite{DiT,SD3} combine strong generative priors with long-range attention, enabling high-fidelity detail synthesis. However, despite their perceptual strength, two key limitations remain.

First, current diffusion-based restoration systems~\cite{StableSR,DiffBIR} often fail under severe degradations such as strong noise, motion blur, or low-resolution compression. These conditions frequently lead to geometric drift, identity loss, or texture hallucination—especially when training is restricted to $\le 1024^2$ pixels~\cite{SUPIR,DreamClear}. Second, although no-reference perceptual scores (e.g., CLIPIQA~\cite{CLIPIQA}, MUSIQ~\cite{MUSIQ}, MANIQA~\cite{MANIQA}) are often high, full-reference distortion metrics (e.g., PSNR, SSIM) remain low—commonly referred to as the perception-distortion gap~\cite{Blau_2018_CVPR}.
Moreover, high-resolution inference is limited by quadratic attention and tight architecture coupling, making tile-consistent scaling difficult without modification.

We study \textbf{In-Token Learning}, an adaptation of a pretrained diffusion transformer for conditional restoration using rectified flow matching (RFM)~\cite{liu2022flow}. Sampling starts from pure noise; the degraded observation supplies spatially aligned evidence rather than the initial evolving latent. Unlike sequence-level conditioning~\cite{ICLoRA,Kontext}, evolving latent tokens and degraded-input tokens are fused along their channel dimension. This preserves the number of image tokens at a fixed resolution, not a resolution-independent attention budget. We report full-image QHD ($2560{\times}1440$) inference and use overlapping tiles for 4K, 8K, and 12K outputs. The pretrained generative prior remains part of the model, and conditioning does not guarantee recovery of lost information or prevention of hallucination.
To further enhance learning stability and semantic conditioning, we introduce a lightweight mechanism: \textbf{Direct Low-Quality Guidance (DLG)}. DLG injects a fused embedding of the degraded image and a per-task system prompt into the text-conditioning pathway. Unlike prior works (e.g., SUPIR~\cite{SUPIR}, FaithDiff~\cite{FaithDiff}) that rely on external vision-language models (VLMs)~\cite{LLAVA} or ControlNet-style side branches~\cite{ControlNet}, DLG provides compact, task-aware guidance at minimal cost.

We evaluate super-resolution and denoising on synthetic (DIV2K~\cite{DIV2K}, LSDIR~\cite{LSDIR}, FFHQ~\cite{FFHQ}) and real-world (RealLQ250~\cite{DreamClear}, RealPhoto60~\cite{SUPIR, FaithDiff}) benchmarks. Performance varies across metrics and datasets, with an important synthetic-to-real generalization limitation on RealLQ250. A \emph{separately trained} automatic colorization checkpoint uses the same backbone and conditioning framework and performs strongly among the evaluated baselines on DIV2K and LSDIR. This is evidence for applicability of the framework to different tasks, not zero-shot transfer or a single joint-task model.

\subsection{Relationship to Fill2SR}
\label{sec:relation}
This report archives the earlier broad study from which \emph{Fill2SR: Repurposing Inpainting Diffusion Transformers for Real-World Super-Resolution}~\cite{Yi2026Fill2SR} developed. The works share the FluxFill backbone, conditional noise-to-clean flow supervision, channel-aligned low-quality evidence, and Redux-based semantic conditioning. Both evaluate on DIV2K, LSDIR, RealLQ250, and RealPhoto60; some benchmark scenes used for comparison recur, including the phone example in Fig.~\ref{fig:RealWorld}, and both use \emph{Along the River During the Qingming Festival} for a high-resolution demonstration. These shared evaluation materials are not independent corroboration, nor are all restored outputs or evaluation protocols identical. The present report preserves the early numerical results and comparison images, including the colorization study, without importing the later Fill2SR experiments.
Fill2SR subsequently focused on real-world super-resolution. It formalized the \emph{Inpainting-Interface Evidence Adapter} (IIEA), interpreting evidence injection through the native masked-image slot under a full-image mask, and introduced \emph{Reference-Conditioned Degradation Transfer} (RCDT) for offline synthesis of paired real-world-like training data. The RCDT pipeline and its later results are not contributions or experiments of this earlier report.

\noindent\textbf{Contributions.}
\begin{itemize}
    \item \textbf{In-Token Learning:} 
    A conditional restoration framework that adapts a pretrained inpainting diffusion transformer with paired rectified flow supervision, in-token alignment, and DLG.
    \item \textbf{Direct Low-Quality Guidance:} 
    A lightweight guidance mechanism that injects task-aware information by fusing degraded-image and prompt embeddings through the native text-conditioning pathway, without external VLMs or ControlNet-style branches.
    \item \textbf{Restoration evaluation:}
    Full-reference and no-reference comparisons across multiple benchmarks and degradation levels, reporting both competitive results and generalization limitations.
    \item \textbf{Ultra-high resolution:} 
    Channel fusion avoids adding another spatial token sequence at a given resolution. We report direct QHD inference and overlapping-tile 4K/8K/12K inference.
    We demonstrate this by successfully restoring a $12$K classical scroll painting.
    \item \textbf{Applicability across tasks:}
    The same framework supports super-resolution/denoising and automatic colorization with two separately trained checkpoints; no shared multi-task checkpoint or zero-shot transfer is claimed.
\end{itemize}

\section{Related Work}

\subsection{Early Methods for Image Restoration}
Early image restoration methods (e.g., BSRGAN~\cite{zhang2021bsrgan}, Real-ESRGAN~\cite{wang2021realesrgan}) are trained as deterministic feed-forward regressors with pixel- or perceptual-loss objectives. While achieving relatively high scores in full-reference metrics, they under-model the conditional target distribution. Consequently, under severe degradations or large upscales, they often exhibit structural distortions or identity drift, leading to degraded no-reference perceptual quality.

\subsection{Diffusion and DiT for Restoration}
Diffusion priors~\cite{DDPM} significantly improve perceptual quality in super-resolution and other restoration tasks~\cite{SR3,DiffBIR}. Evolving from U-Nets~\cite{UNet}, DiT~\cite{DiT,SD3} variants enable global attention by tokenizing latent features. Although perceptual quality improves, the perception-distortion gap persists, limiting fidelity under severe degradations~\cite{Blau_2018_CVPR,FluxIR}. Moreover, they are typically limited to training resolutions $\leq 1024^2$ pixels, restricting scalability.

\subsection{Conditioning and Token Alignment}
Preserving structural and identity information in diffusion restoration~\cite{DDPM} remains challenging when observations lose information or conditioning is imperfect. Several methods~\cite{SUPIR,DreamClear,FaithDiff,FluxIR} use external semantic guidance and/or ControlNet-like side branches~\cite{ControlNet}. These designs add conditioning costs and exhibit fidelity/perceptual trade-offs; iterative sampling alone does not explain all such errors.

More recently, in-context concatenation approaches~\cite{ICLoRA,Kontext} attempt to align degraded inputs with denoising latents by expanding token sequences. While helpful, sequence-level fusion inflates attention cost quadratically and complicates high-resolution scaling.

\section{Method}
\label{sec:Method}

\subsection{Overview}
We adapt a pretrained inpainting diffusion transformer to sample restored images from noise via conditional RFM, using in-token alignment and DLG.
We explicitly define a degradation model $\mathcal{D}_\phi$ to synthesize paired training data; the clean endpoint provides supervision while the pretrained generative prior is retained.
Our approach is then trained to learn a conditional velocity field that maps noise to the clean latent, conditioned by the degraded input generated by this degradation model.

\begin{figure}[t]
    \centering
    \includegraphics[width=0.8\linewidth]{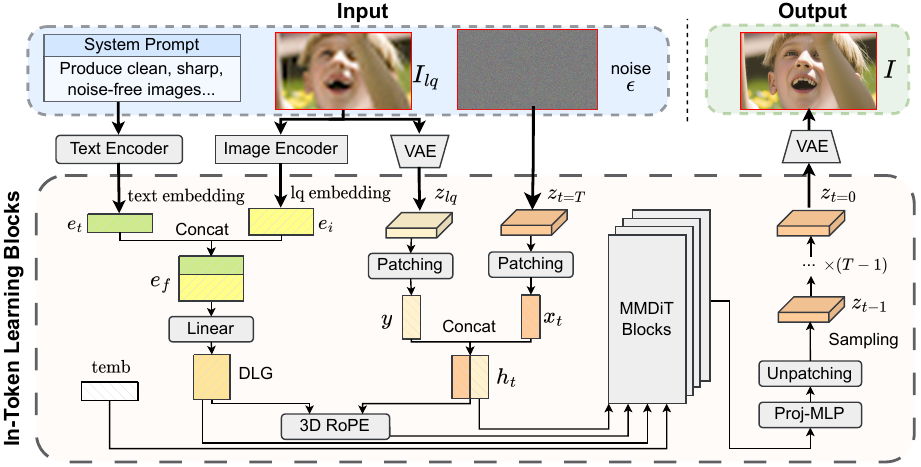}
    \caption{\textbf{In-Token Learning overview.} Paired clean targets supervise a conditional velocity field sampled from noise toward restored images. In-token alignment preserves the spatial token count at a fixed resolution; DLG supplies task and image embeddings. SR/denoising and colorization use separately trained checkpoints. Full-image QHD and tiled 4K/8K/12K inference are distinct settings.}
    \label{fig:method_overview}
\end{figure}

As shown in Fig.~\ref{fig:method_overview}, our method generates restored images from pure noise during inference.
At each step $t$, noisy-latent tokens $x_t$ and low-quality tokens $y$ (from degraded image input) are concatenated channel-wise into $h_t=[x_t;y]$, which is processed by MMDiT blocks. 
In parallel, DLG injects a concatenation fusion of the degraded image input and a task prompt into the text-conditioning pathway. 
The paired supervision and two conditioning paths encourage structural fidelity without trainable auxiliary ControlNets or VLM-generated captions; frozen image and text encoders are still used.

\subsection{Degradation Model and Synthetic Supervision}
\label{sec:degradation}

Given a clean image $I$, $\mathcal{D}_\phi$ degrades it into a low-quality counterpart $I_{\mathrm{lq}}$. This process enables supervised restoration training by defining the ground-truth velocity field from pure noise to $I$, conditioned on $I_{\mathrm{lq}}$.
For super-resolution and denoising,
\begin{equation}
I_{\mathrm{lq}} = \mathcal{D}_\phi(I) 
= \mathrm{JPEG}_q\!\big(( I * \kappa_{\sigma_b} ) \downarrow_s \big) + \eta, 
\quad \eta \sim \mathcal{N}(0,\sigma_n^2),
\end{equation}
where $s$ is the downsampling factor, $\kappa_{\sigma_b}$ is a Gaussian blur kernel with standard deviation $\sigma_b$, $q$ denotes JPEG quality, and $\eta$ is additive Gaussian noise with standard deviation $\sigma_n$.  
For colorization, $\mathcal{D}_\phi$ converts RGB images to grayscale:
\begin{equation}
I_{\mathrm{lq}} = \mathcal{D}_\phi(I) = \mathcal{G}(I),
\end{equation}
where $\mathcal{G}$ denotes grayscale conversion. This unified formulation ensures controllable coverage, and enables systematic stress-testing across degradations.

\subsection{Noise-to-Clean Mapping via Rectified Flow}
\label{sec:noise2clean}

Let $E,D$ denote the encoder and decoder of a pretrained VAE~\cite{SD}.
Given a training pair $(I_{\mathrm{lq}}, I)$, with $I_{\mathrm{lq}}=\mathcal{D}_\phi(I)$, we obtain the clean target $z_0=E(I)$ and degraded latent $z_{\mathrm{lq}}=E(I_{\mathrm{lq}})$. Downsampling, compression, and grayscale conversion lose information: the paired target is known during training but is not a unique inverse of the observation.

We formulate restoration as learning a conditional velocity field~\cite{liu2022flow} $v_\theta$ that transports a noisy latent
\begin{equation}
z_t = (1-t)z_0 + t\epsilon, \quad \epsilon\sim\mathcal{N}(0,\mathrm{I}),
\end{equation}
towards the clean latent $z_0$.  
At inference, the ODE is integrated from $t=1$ to $t=0$, starting from Gaussian noise. This separates sampling initialization from the degraded observation; artifacts can still influence the output through conditioning. Spatial alignment via in-token fusion and semantic guidance via DLG provide complementary evidence.

\subsection{Rectified Flow Matching Objective}
\label{sec:rfm}

We adopt rectified flow matching~\cite{lipman2022flow} to train the model.  
The training loss is:
\begin{equation}
\mathcal{L}_{\text{RFM}} = \mathbb{E}_{(I_{\mathrm{lq}},I),\epsilon,t}\Big[\;\|v_\theta(z_t,t,h_t,e_f) - (\epsilon-z_0)\|_2^2\;\Big].
\end{equation}
Here $x_t$ and $y$ are aligned tokens of $z_t$ and $z_{\mathrm{lq}}$, and $h_t=[x_t;y]$. The paired clean target supervises the path velocity; this loss alone does not establish faster or more stable convergence than other objectives. More theoretical notes are provided in Appendix~\ref{app:theory}.

\subsection{In-Token Alignment}
\label{sec:inToken}

To efficiently inject degraded features, we concatenate latent tokens $x_t$ and $y$ along the channel dimension to form in-token fusion:
\begin{equation}
h_t = [x_t;y] \in \mathbb{R}^{N \times 2d},
\end{equation}
with $N$ tokens and pre-fusion feature width $d$.
As shown in Fig.~\ref{fig:method_overview}, $y$ is concatenated to the channel dimension of $x_t$ at each denoising step.
The fused input is projected into the pretrained backbone; its width $2d$ does not imply that all Transformer hidden widths double. No additional spatial token sequence is introduced. See Section~\ref{sec:complexity} for the scope of the complexity comparison.

\subsection{Direct Low-Quality Guidance}
\label{sec:dlg}

To let our proposed method learn the reverse process of the degradation model more directly, we use a system prompt to provide semantic guidance for the reverse process. 
We also calculate the low-quality embedding $e_i$ from $I_{\mathrm{lq}}$ via a frozen image encoder~\cite{flux1_redux_dev}.
We provide the final guidance by fusing the fixed system prompt embedding $e_t$ with degraded low-quality embedding $e_i$:
\begin{equation}
e_f=[e_t;e_i]\in\mathbb{R}^{(n_t+n_i)\times c},
\end{equation}
where $n_t,n_i$ are token counts and $c$ the channel width. $e_f$ is fed into the native conditioning pathway at every step. DLG uses frozen encoders, but does not add a trainable ControlNet-style branch or VLM captioner. The fixed task prompts are ``produce a clean high-quality image'' for SR/denoising and ``colorize this grayscale input'' for colorization; image-specific prompt engineering is not required.

\subsection{Inference Details}
\label{sec:inference}
For resolutions up to QHD ($2560\times1440$), each sampling step is performed on the full image in a single forward pass. For 4K/8K/12K outputs, we tile the evolving latent, $z_{\mathrm{lq}}$, and $I_{\mathrm{lq}}$, compute tile-local image embeddings, and blend overlapping predictions. This limits the per-pass spatial token budget, but does not constitute native full-image 12K attention or guarantee global consistency and seam-free outputs. Details are provided in Appendix~\ref{app:inference}.

\subsection{Complexity and Scalability}
\label{sec:complexity}
Dense self-attention has a quadratic spatial-token term. Appending a second image stream would increase $N$ to $2N$; channel fusion instead preserves $N$ before projection into the backbone. At matched backbone width, the idealized attention term scales as $(2N+n_f)^2$ versus $(N+n_f)^2$, where $n_f$ counts semantic tokens. This is a token-count comparison, not a measured fourfold reduction in total memory or a twofold runtime improvement. Projection, MLP, encoder, and implementation costs remain. Increasing image resolution still increases $N$ and attention cost. See Appendix~\ref{app:complexity}.

\section{Experiments}
\label{sec:experiments}

We evaluate two separately trained checkpoints: one for super-resolution and denoising and one for automatic colorization. They share the framework but not the trained task-specific parameters.
We also report a tiled 12K restoration demonstration of \textit{Along the River During the Qingming Festival}, using the SR/denoising checkpoint. This demonstration is distinct from full-image QHD inference and is not evidence for quantitatively validated recovery of historical details. Material availability is discussed in the reproducibility statement.

\subsection{Setup}

\textbf{Test Datasets.}
For synthetic SR and denoising, we use a subset of FFHQ~\cite{FFHQ} and the validation sets of DIV2K~\cite{DIV2K}, LSDIR~\cite{LSDIR} with three degradation levels (D1/D2/D3: $\times2/\times4/\times8$ downsampling with increasing blur, noise, and JPEG compression);
For real-world SR, we use RealLQ250~\cite{DreamClear} and RealPhoto60~\cite{SUPIR,FaithDiff}.
For automatic colorization, different from prior works which mainly test on ImageNet~\cite{ImageNet} at $256^2$ resolution, we convert the validation sets of DIV2K and LSDIR to grayscale images and evaluate at their original resolution.

\textbf{Implementation Details.}
We build on the pretrained FluxFill~\cite{Flux} inpainting backbone, repurposing its channel-aligned image conditioning for restoration rather than user-specified masked editing. The original study described this as disabling the mask mechanism; the later Fill2SR work formalized the native-slot interpretation as IIEA. We retain the early in-token formulation here rather than retroactively claiming IIEA or RCDT as new components of this report.
We fine-tune the DiT at bf16 precision with LoRA~\cite{Lora} (rank 384) for 14k steps on 4$\times$RTX 5880 Ada, using a cosine learning-rate schedule with peak $2.5{\times}10^{-4}$, 2.5k warm-up steps, a global batch size of 16, and 28 sampling steps at inference.

\textbf{Metrics.}
For super-resolution and denoising, we report full-reference (PSNR/SSIM/LPIPS) and no-reference (MANIQA~\cite{MANIQA}, CLIPIQA~\cite{CLIPIQA}, MUSIQ~\cite{MUSIQ}) metrics, following prior works~\cite{SUPIR,FaithDiff}.
For automatic image colorization, we include the colorfulness score (CF)~\cite{CF}, and $\Delta$CF, the absolute deviation between the output CF and the ground-truth CF, following DDColor~\cite{DDColor}.

\subsection{Super-Resolution and Denoising}

\begin{figure}[h]
    \centering

    \begin{subfigure}{0.9\textwidth}
        \centering
        \begin{subfigure}{0.3\textwidth}
            \includegraphics[width=\linewidth]{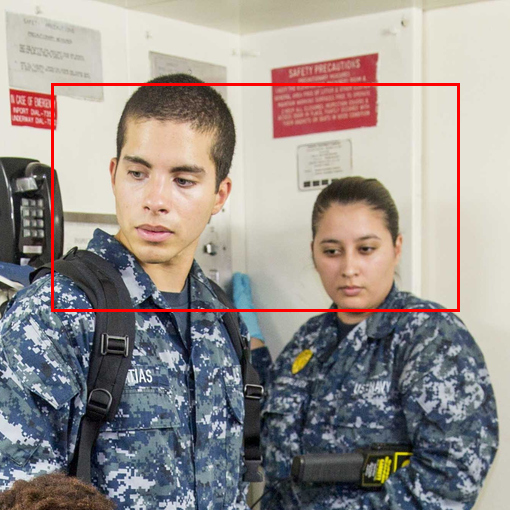}
            \caption*{GT}
        \end{subfigure}
        \begin{subfigure}{0.69\textwidth}
            \centering
            \begin{subfigure}{0.32\textwidth}
                \includegraphics[width=\linewidth]{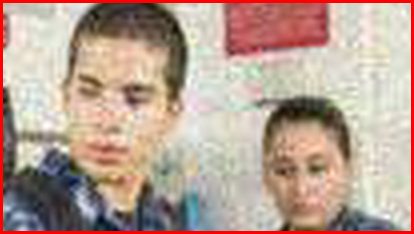}
                \caption*{LQ Input (D2)}
            \end{subfigure}
            \begin{subfigure}{0.32\textwidth}
                \includegraphics[width=\linewidth]{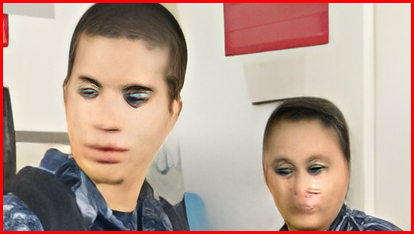}
                \caption*{Real-ESRGAN}
            \end{subfigure}
            \begin{subfigure}{0.32\textwidth}
                \includegraphics[width=\linewidth]{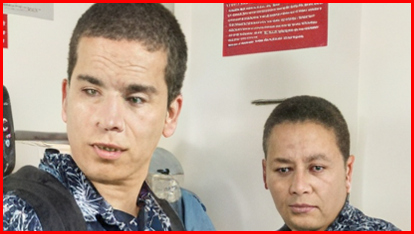}
                \caption*{SUPIR}
            \end{subfigure}
            \\
            \begin{subfigure}{0.32\textwidth}
                \includegraphics[width=\linewidth]{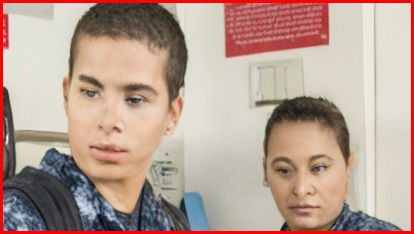}
                \caption*{FaithDiff}
            \end{subfigure}
            \begin{subfigure}{0.32\textwidth}
                \includegraphics[width=\linewidth]{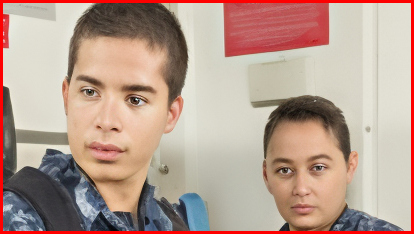}
                \caption*{SeeSR}
            \end{subfigure}
            \begin{subfigure}{0.32\textwidth}
                \includegraphics[width=\linewidth]{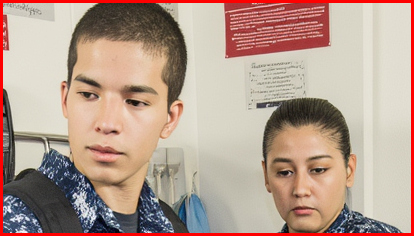}
                \caption*{\textbf{Ours}}
            \end{subfigure}
        \end{subfigure}
    \end{subfigure}

\caption{\textbf{Qualitative comparisons on \textit{DIV2K-Val} (synthetic).}
Instead of denoising from a degraded latent, our method generates from pure noise using a conditional velocity field guided by in-token alignment and DLG.
The selected crops illustrate differences in eye shape, gaze direction, thin edges, and textures; they do not establish a universal fidelity ranking.}
\label{fig:Synthetic}
\end{figure}

\begin{table}[tbp]
  \centering
\caption{Quantitative comparison on the validation set of \textit{DIV2K}, \textit{LSDIR}, \textit{FFHQ} (synthetic), \textit{RealLQ250}, and \textit{RealPhoto60} (real-world).
For synthetic datasets, no-reference metrics (CLIPIQA/MUSIQ/MANIQA) are averaged over D1--D3. Full-reference metrics (PSNR/SSIM/LPIPS) are reported for D1 only.
For real-world sets, we only apply $4{\times}$ upscale on \textit{RealLQ250} and $2{\times}$ upscale on \textit{RealPhoto60}.
Best is \textbf{\underline{bold+underline}} and second-best is \textbf{bold}.}
  \label{tab:main_summary}
  \renewcommand{\arraystretch}{1}        
  \setlength{\tabcolsep}{3.6pt}             
  \small                                    
  \begin{adjustbox}{max width=\linewidth}   
      \begin{tabular}{l l *{7}{S[table-format=2.4]}>{\columncolor{gray!8}}S[table-format=2.4]}
      \toprule
      \multirow{2}{*}{Datasets} & \multirow{2}{*}{Metrics} & \multicolumn{8}{c}{Methods} \\
      \cmidrule(lr){3-10}
       &  & \mch{Real-ESRGAN} & \mch{StableSR} & \mch{DiffBIR} & \mch{SeeSR} & \mch{SUPIR} & \mch{DreamClear} & \mch{FaithDiff} & \mch{\textbf{Ours}} \\
      \midrule

      \multirow{6}{*}{\textit{DIV2K-Val}}
        & CLIPIQA$\uparrow$ & 0.48753443 & 0.38011043 & 0.57374693 & \secbest{0.60352331} & 0.50508837 & 0.53098785 & 0.57633828 & \best{0.62134138} \\
        & MUSIQ$\uparrow$   & 58.969925  & 49.0826152 & 66.1573225 & \best{69.0463833}    & 64.6141652 & 65.9251214 & \secbest{68.1122318} & 66.3780277 \\
        & MANIQA$\uparrow$  & 0.52521121 & 0.44444642 & 0.57094952 & 0.60154404          & 0.58579387 & 0.57434695 & \secbest{0.61842447} & \best{0.62182545} \\
        & PSNR$\uparrow$    & \best{26.0668788} & 24.1885308 & \secbest{25.3999493} & 24.51775 & 25.3268562 & 23.3814183 & 24.372315 & 25.2706279 \\
        & SSIM$\uparrow$    & \best{0.76776759} & 0.70437463 & 0.70363412           & 0.67381084 & 0.70254461 & 0.64135033 & 0.66090321 & \secbest{0.7057489} \\
        & LPIPS$\downarrow$    & \best{0.2997} & 0.3186 & 0.3226 & 0.3434 & 0.3107 & 0.3464 & 0.3302 & \secbest{0.3087} \\      
        \midrule

      \multirow{6}{*}{\textit{LSDIR-Val}}
        & CLIPIQA$\uparrow$ & 0.54238515 & 0.41985749 & 0.63091531 & \secbest{0.64131011} & 0.57035871 & 0.61209618 & 0.63066873 & \best{0.72174122} \\
        & MUSIQ$\uparrow$   & 64.5600666 & 52.8769694 & 70.1139612 & \best{72.6831251}    & 67.9507897 & 70.3541404 & 71.6307486 & \secbest{72.6591677} \\
        & MANIQA$\uparrow$  & 0.56744843 & 0.47876757 & 0.61533503 & 0.6387702            & 0.62336608 & 0.6148953  & \secbest{0.66528687} & \best{0.68796813} \\
        & PSNR$\uparrow$    & \best{23.2872128} & 21.3458313 & 22.3081724 & 21.5545021 & 22.2405802 & 21.0995096 & 21.1641823 & \secbest{22.733831} \\
        & SSIM$\uparrow$    & \best{0.72201729} & 0.62410819 & 0.63545957           & 0.58594149 & 0.64313763 & 0.60291261 & 0.57021125 & \secbest{0.680924} \\
        & LPIPS$\downarrow$    & \best{0.2805} & 0.3164 & 0.3052 & 0.3349 & 0.3017 & 0.3190 & 0.3255 & \secbest{0.2819} \\
        \midrule

      \multirow{6}{*}{\textit{FFHQ}}
        & CLIPIQA$\uparrow$ & 0.42974361 & 0.48268006 & \best{0.62825903} & 0.54929475 & 0.53004989 & 0.48071575 & 0.57186803 & \secbest{0.57788007} \\
        & MUSIQ$\uparrow$   & 63.419872  & 66.7730908 & 73.9354306       & 73.384249  & 73.6045764 & 70.758564  & \best{76.1941224} & \secbest{74.1304778} \\
        & MANIQA$\uparrow$  & 0.49202478 & 0.50293169 & 0.59246225        & 0.58748612 & 0.59698592 & 0.56536722 & \best{0.64055634} & \secbest{0.61507941} \\
        & PSNR$\uparrow$    & \best{30.9152003} & 28.8964509 & 29.8402588 & 29.4370996 & 30.2578438 & 28.2803803 & 28.6318853 & \secbest{30.3367874} \\
        & SSIM$\uparrow$    & \best{0.84478184} & 0.79108477 & 0.78658425 & 0.78514645 & 0.79014311 & 0.74665857 & 0.74339401 & \secbest{0.79323577} \\
        & LPIPS$\downarrow$    & 0.3196 & \best{0.2981} & 0.3426 & 0.3181 & \secbest{0.2991} & 0.3309 & 0.3201 & 0.3162 \\
        \midrule

      \multirow{3}{*}{\textit{RealLQ250}}
        & CLIPIQA$\uparrow$ & 0.45856981 & 0.35537396 & 0.54099948 & \best{0.566769} & 0.48623623 & 0.50847115 & \secbest{0.54658837} & 0.4791711 \\
        & MUSIQ$\uparrow$   & 57.4012455 & 46.1955007 & 62.7923406 & \best{67.1608968} & 60.9576432 & 61.814557 & \secbest{66.5082069} & 54.34269 \\
        & MANIQA$\uparrow$  & 0.53329032 & 0.45783277 & 0.58831226 & \secbest{0.60262616} & 0.58160334 & 0.5748596 & \best{0.62982667} & 0.53968051 \\
    \midrule

    \multirow{3}{*}{\textit{RealPhoto60}}
    & CLIPIQA$\uparrow$ & 0.50682169 & 0.363178347 & 0.546462551 & 0.595883103 & \secbest{0.597248018} & 0.538348669 & 0.571801438 & \best{0.602827467} \\
    & MUSIQ$\uparrow$   & 59.02964811 & 50.26701819 & 61.66456555 & \best{71.80520865} & 69.6325875 & 68.83532995 & \secbest{71.58525817} & 65.71394881 \\
    & MANIQA$\uparrow$  & 0.479661382 & 0.455356112 & 0.561709506 & 0.607913569 & 0.6116412 & 0.590508775 & \best{0.651286211} & \secbest{0.616781467} \\
    
      \bottomrule
    \end{tabular}
  \end{adjustbox}
\end{table}

\begin{figure}[h]
    \centering
    \begin{subfigure}{0.85\textwidth}
        \centering
        \begin{subfigure}{0.3\textwidth}
            \includegraphics[width=\linewidth]{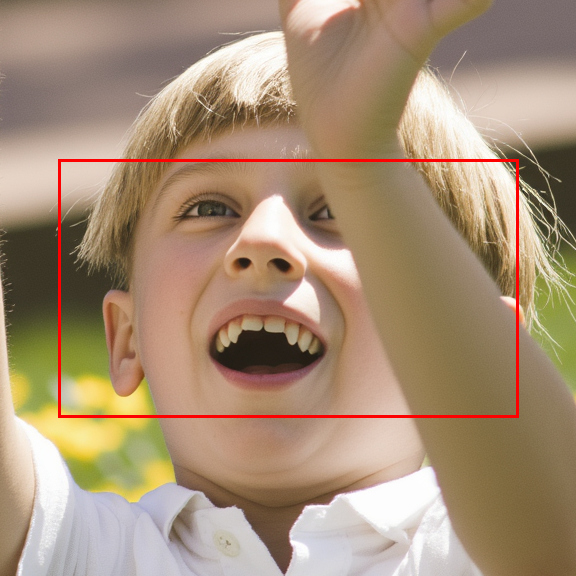}
            \caption*{\textbf{Ours}}
        \end{subfigure}
        \begin{subfigure}{0.69\textwidth}
            \centering
            \begin{subfigure}{0.32\textwidth}
                \includegraphics[width=\linewidth]{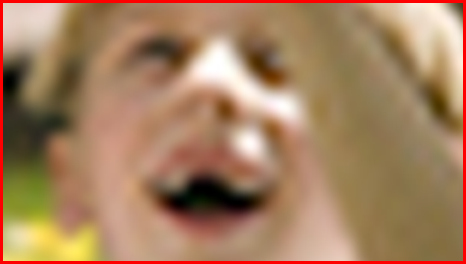}
                \caption*{LQ Input}
            \end{subfigure}
            \begin{subfigure}{0.32\textwidth}
                \includegraphics[width=\linewidth]{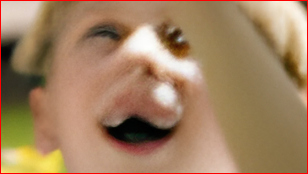}
                \caption*{Real-ESRGAN}
            \end{subfigure}
            \begin{subfigure}{0.32\textwidth}
                \includegraphics[width=\linewidth]{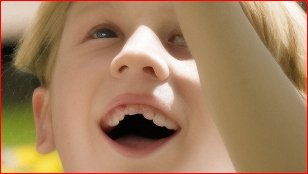}
                \caption*{DiffBIR}
            \end{subfigure}
            \\
            \begin{subfigure}{0.32\textwidth}
                \includegraphics[width=\linewidth]{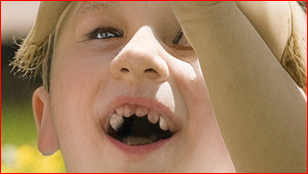}
                \caption*{FaithDiff}
            \end{subfigure}
            \begin{subfigure}{0.32\textwidth}
                \includegraphics[width=\linewidth]{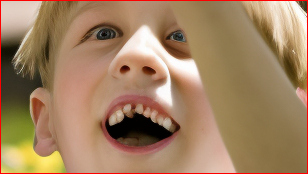}
                \caption*{SeeSR}
            \end{subfigure}
            \begin{subfigure}{0.32\textwidth}
                \includegraphics[width=\linewidth]{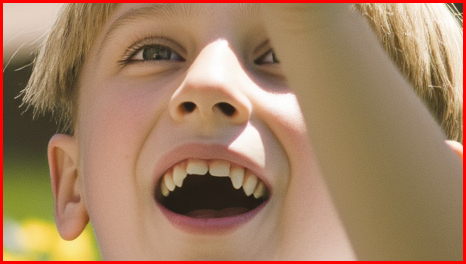}
                \caption*{\textbf{Ours}}
            \end{subfigure}
        \end{subfigure}
    \end{subfigure}
    \begin{subfigure}{0.85\textwidth}
      \centering
      \begin{subfigure}{0.3\textwidth}
          \includegraphics[width=\linewidth]{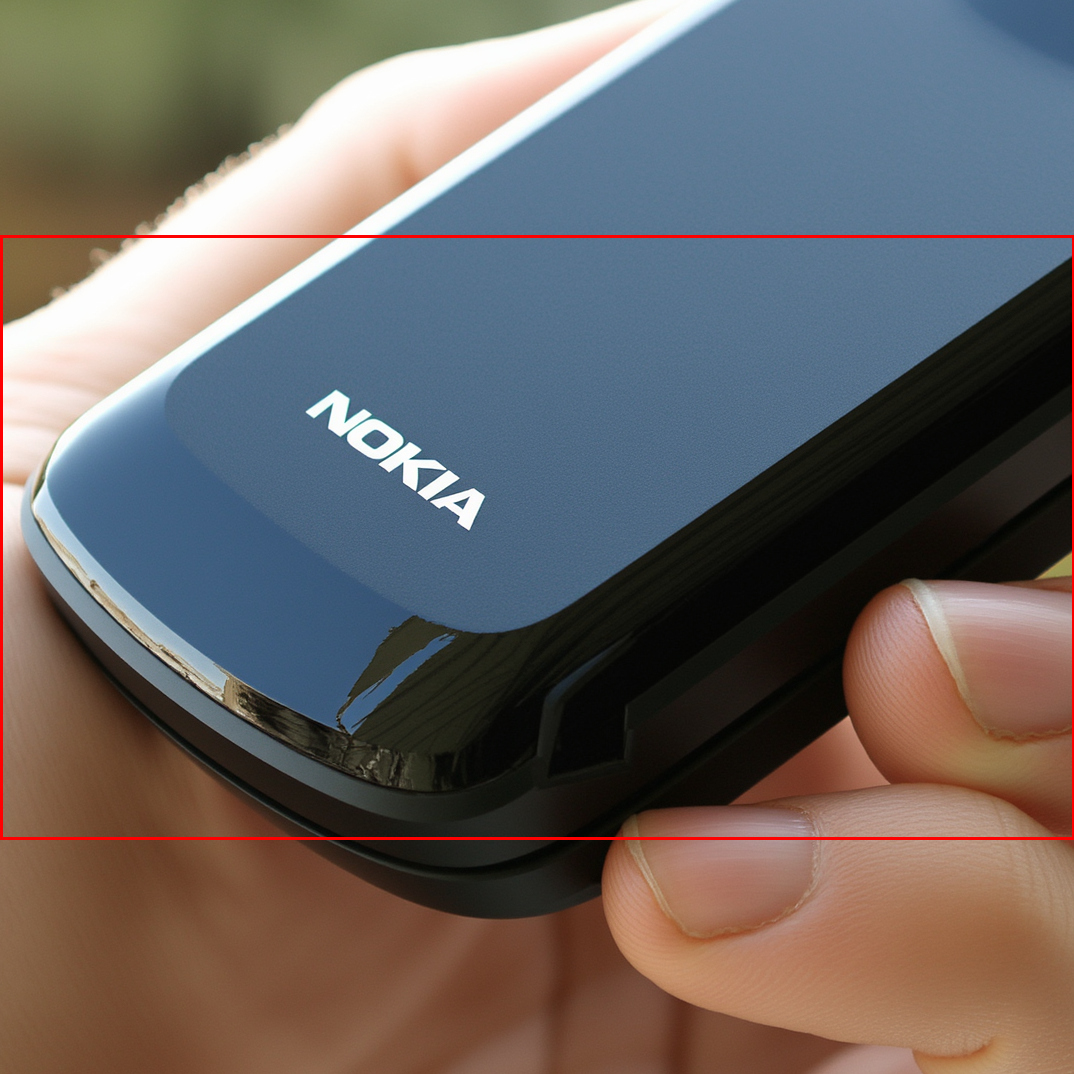}
          \caption*{\textbf{Ours}}
      \end{subfigure}
      \begin{subfigure}{0.69\textwidth}
          \centering
          \begin{subfigure}{0.32\textwidth}
              \includegraphics[width=\linewidth]{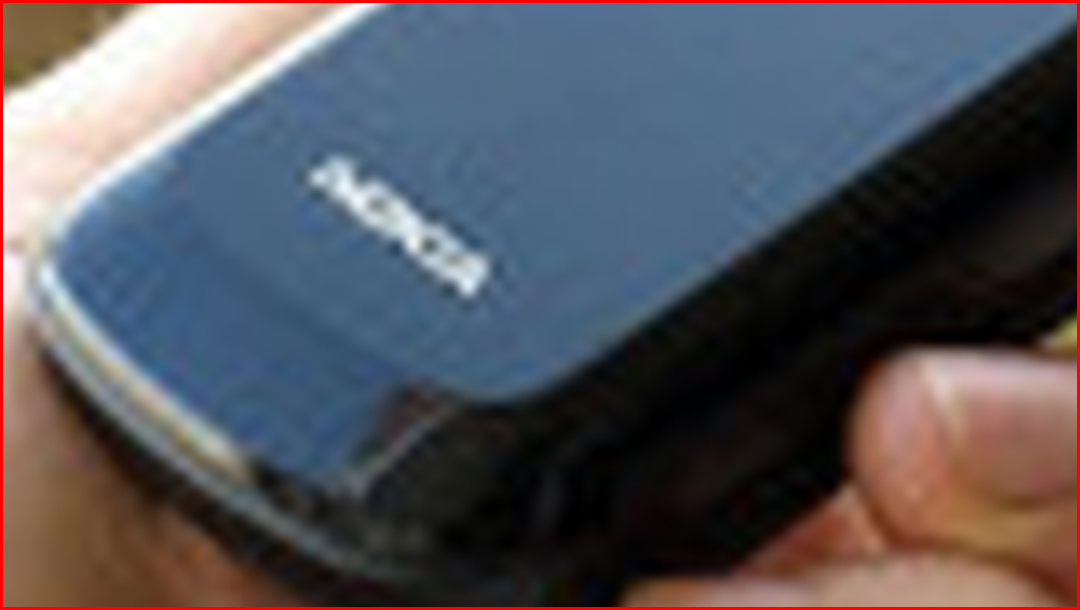}
              \caption*{LQ Input}
          \end{subfigure}
          \begin{subfigure}{0.32\textwidth}
              \includegraphics[width=\linewidth]{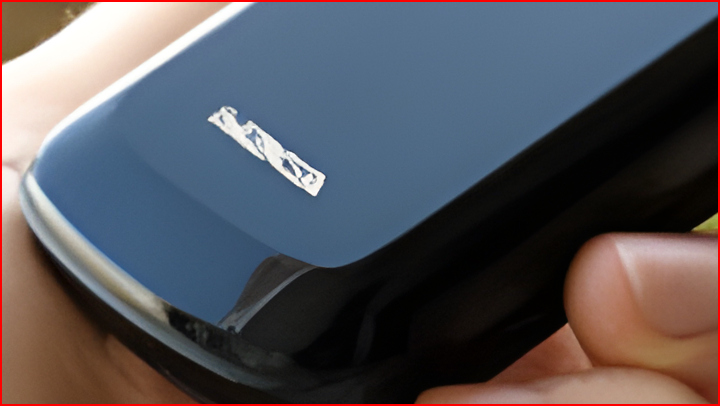}
              \caption*{Real-ESRGAN}
          \end{subfigure}
          \begin{subfigure}{0.32\textwidth}
              \includegraphics[width=\linewidth]{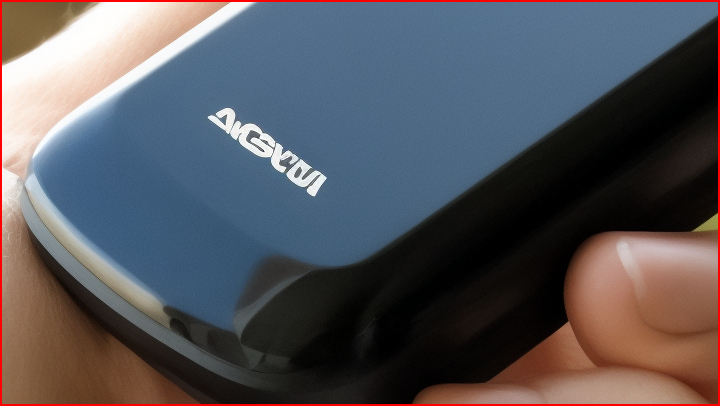}
              \caption*{DiffBIR}
          \end{subfigure}
          \\
          \begin{subfigure}{0.32\textwidth}
              \includegraphics[width=\linewidth]{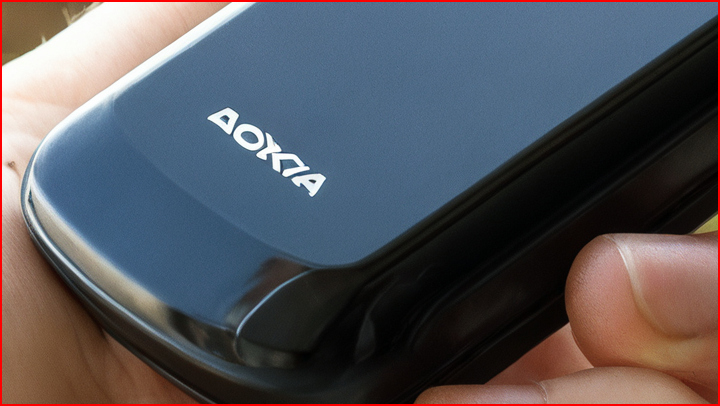}
              \caption*{FaithDiff}
          \end{subfigure}
          \begin{subfigure}{0.32\textwidth}
              \includegraphics[width=\linewidth]{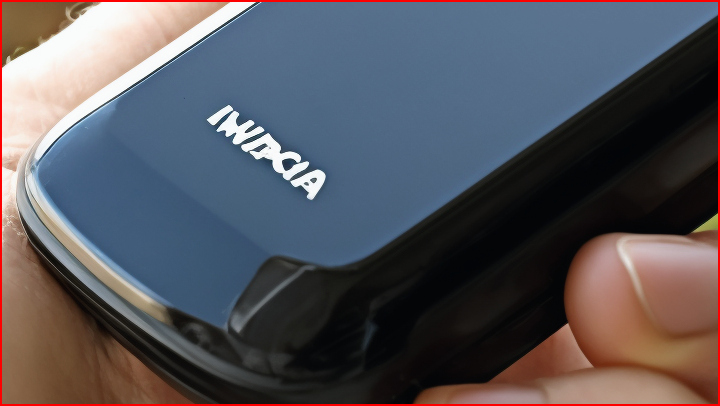}
              \caption*{SeeSR}
          \end{subfigure}
          \begin{subfigure}{0.32\textwidth}
              \includegraphics[width=\linewidth]{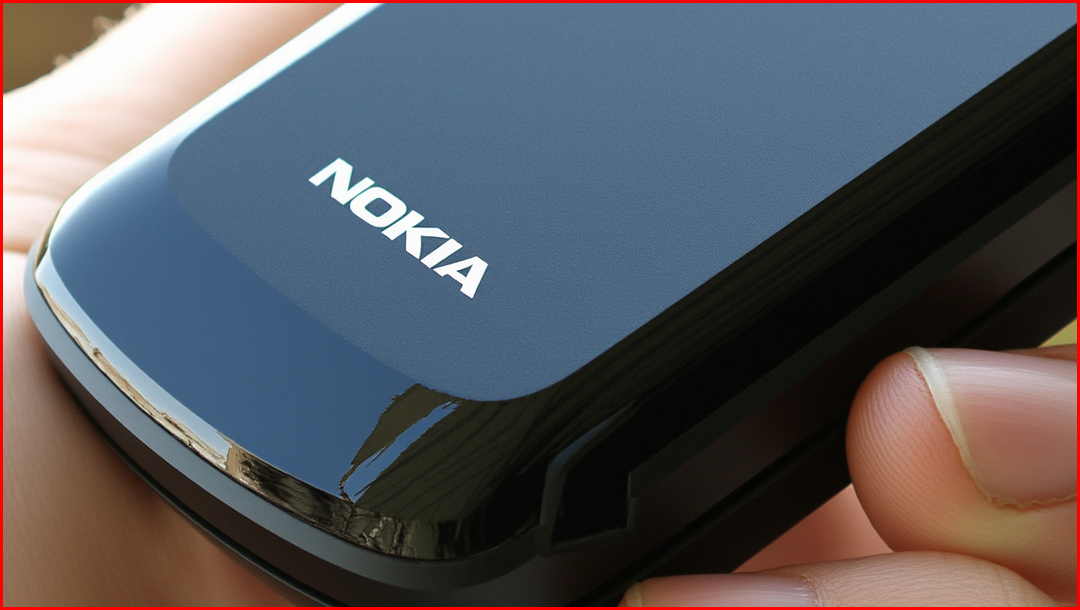}
              \caption*{\textbf{Ours}}
          \end{subfigure}
      \end{subfigure}
  \end{subfigure}

\caption{\textbf{Qualitative comparisons on \textit{RealLQ250} (real-world).}
Left: our full-resolution results with crop locations marked (red boxes).
Right: low-quality input and outputs from Real-ESRGAN, DiffBIR, FaithDiff, SeeSR, and our method on the corresponding crops.
These selected examples complement the aggregate scores and failure cases; real-world performance is not uniformly best.}
\label{fig:RealWorld}
\end{figure}

\noindent\textbf{Quantitative.}
Table~\ref{tab:main_summary} reports results on DIV2K, LSDIR, FFHQ, RealLQ250, and RealPhoto60, against GAN baseline (Real-ESRGAN~\cite{wang2021realesrgan}) and recent diffusion-based methods (StableSR~\cite{StableSR}, DiffBIR~\cite{DiffBIR}, SeeSR~\cite{SeeSR}, SUPIR~\cite{SUPIR}, DreamClear~\cite{DreamClear}, FaithDiff~\cite{FaithDiff}).

On synthetic datasets, our method provides a competitive trade-off between distortion and perceptual metrics, but does not rank first or second in every dataset/metric combination. Real-ESRGAN often obtains higher PSNR/SSIM, while several diffusion methods perform better on individual no-reference scores. Table~\ref{tab:main_summary} reports synthetic full-reference metrics at D1 and no-reference metrics averaged over D1--D3; Appendix~\ref{app:main_summary} gives per-level results. These different aggregation protocols should not be conflated, and no-reference quality is not a direct measure of fidelity to the clean target.

On RealPhoto60, our method has the highest CLIPIQA among the evaluated methods, but not the highest MUSIQ or MANIQA. On RealLQ250, its aggregate no-reference scores are below several baselines, revealing a substantial limitation in generalization from the synthetic degradation recipe to real observations. Selected qualitative examples cannot establish superiority for a majority of images, and residual blur or closeness to the input is not proof of an uncertainty-aware reliability strategy. Figure~\ref{app:RealHQ250Fail} documents failures, while the additional blur experiment shows sensitivity to preprocessing rather than disproving overfitting. Outputs, including historical-artwork demonstrations, must not be treated as verified recovery of details absent from the input.

\noindent\textbf{Qualitative.}
Figures~\ref{fig:Synthetic} and \ref{fig:RealWorld} show selected synthetic and real-world comparisons, complementing rather than replacing aggregate evaluation.
Existing diffusion-based methods often struggle to balance strong denoising or large upscale with faithful structure/identity preservation, occasionally introducing geometric drift or texture hallucination under severe degradations.
Our outputs preserve several visible structures in these examples (e.g., eye shape, thin edges, and textures), but can also lose details or retain artifacts; the examples do not establish freedom from over-smoothing or hallucination.
Appendix~\ref{app:result} includes per-degradation comparisons (Fig.~\ref{app:Degradation_LSDIR}), DIV2K results (Fig.~\ref{app:Synthetic}), RealPhoto60 results (Fig.~\ref{app:RealWorld}), and RealLQ250 failure cases (Fig.~\ref{app:RealHQ250Fail}).

\subsection{Application to Automatic Image Colorization}
\label{sec:colorization}

\begin{figure}[h]
    \centering
    \begin{subfigure}{0.85\textwidth}
        \centering
        \begin{subfigure}{0.3\textwidth}
            \includegraphics[width=\linewidth]{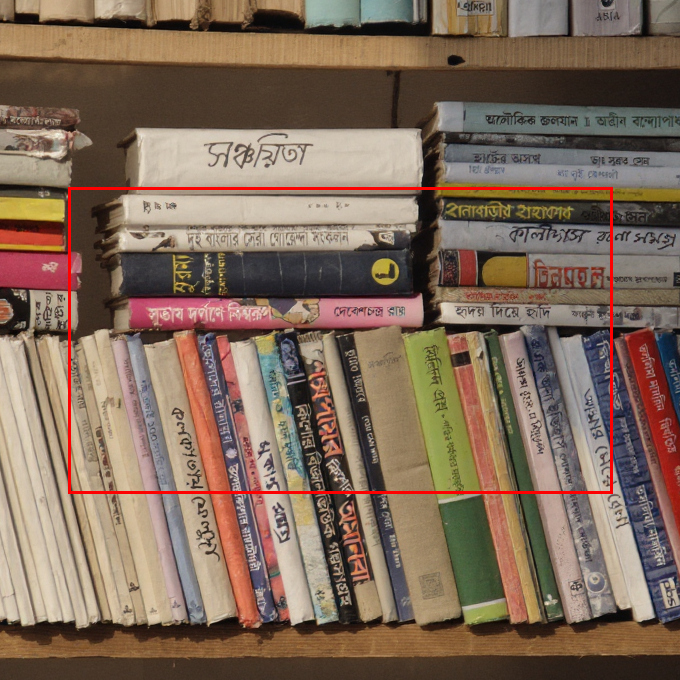}
            \caption*{\textbf{Ours}}
        \end{subfigure}
        \begin{subfigure}{0.69\textwidth}
            \centering
            \begin{subfigure}{0.32\textwidth}
                \includegraphics[width=\linewidth]{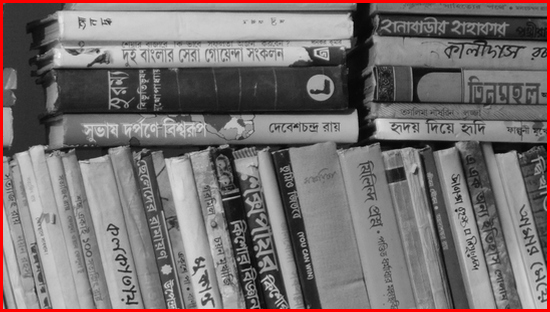}
                \caption*{LQ Input}
            \end{subfigure}
            \begin{subfigure}{0.32\textwidth}
                \includegraphics[width=\linewidth]{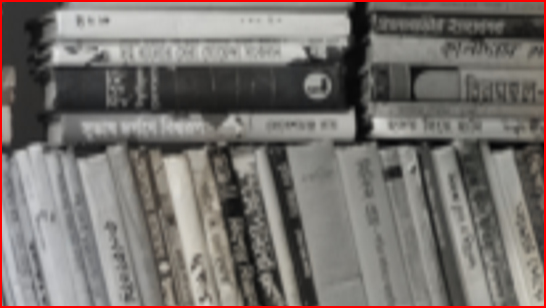}
                \caption*{InstColor}
            \end{subfigure}
            \begin{subfigure}{0.32\textwidth}
                \includegraphics[width=\linewidth]{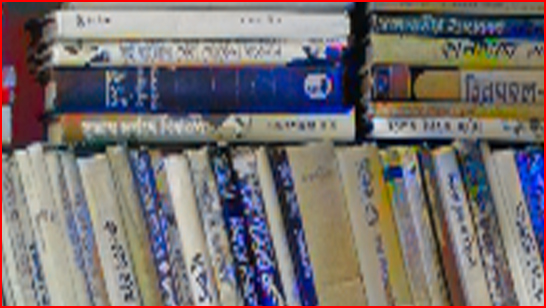}
                \caption*{ColorFormer}
            \end{subfigure}
            \\
            \begin{subfigure}{0.32\textwidth}
                \includegraphics[width=\linewidth]{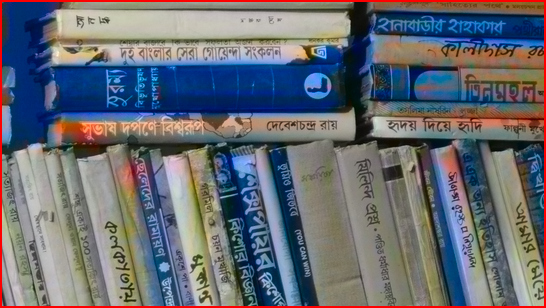}
                \caption*{BigColor}
            \end{subfigure}
            \begin{subfigure}{0.32\textwidth}
                \includegraphics[width=\linewidth]{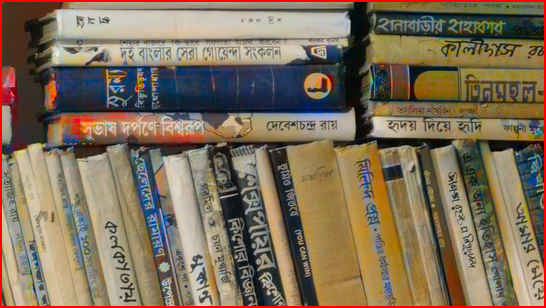}
                \caption*{DDColor}
            \end{subfigure}
            \begin{subfigure}{0.32\textwidth}
                \includegraphics[width=\linewidth]{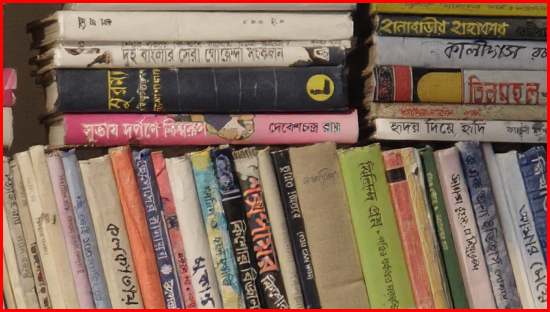}
                \caption*{\textbf{Ours}}
            \end{subfigure}
        \end{subfigure}
    \end{subfigure}
    \begin{subfigure}{0.85\textwidth}
        \centering
        \begin{subfigure}{0.3\textwidth}
            \includegraphics[width=\linewidth]{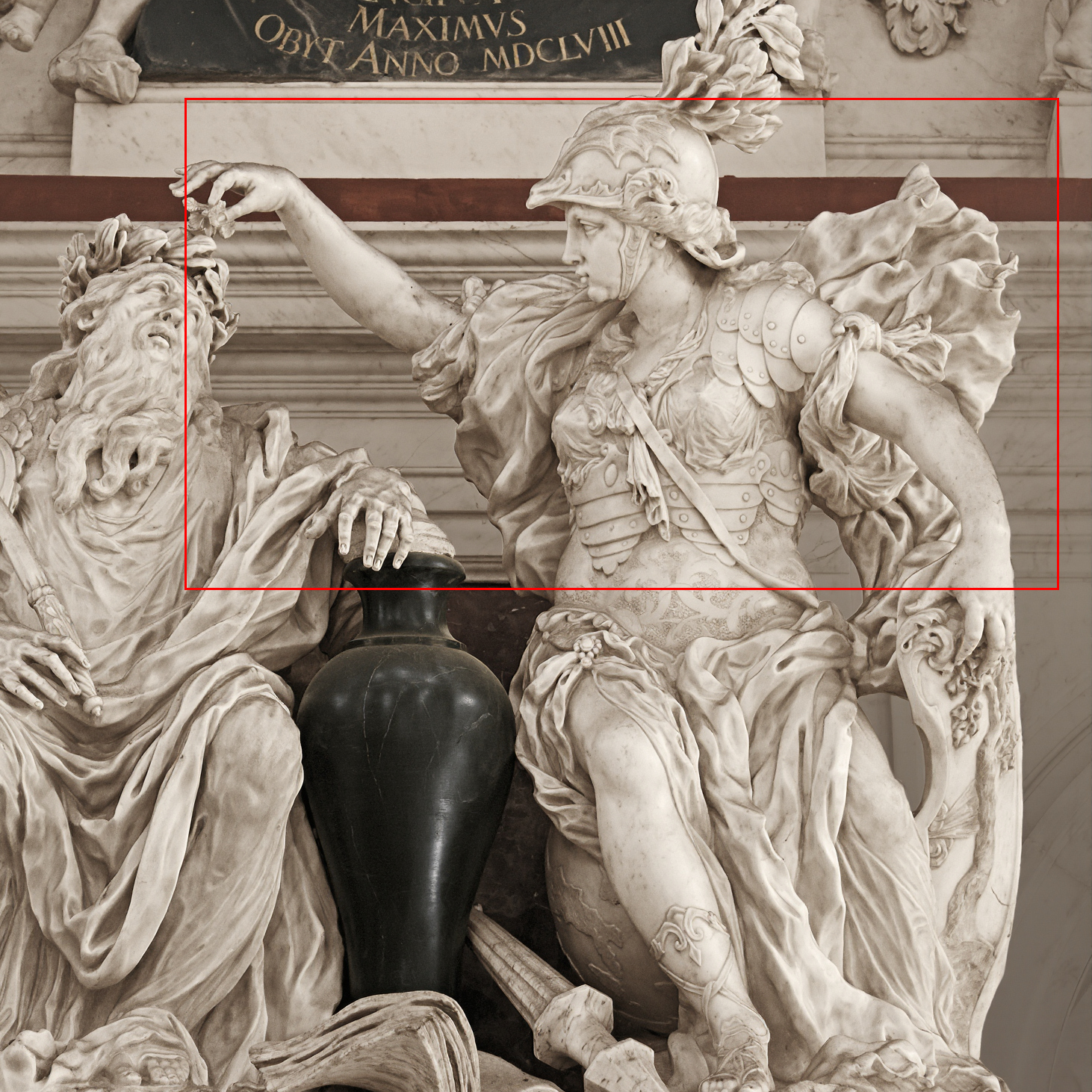}
            \caption*{\textbf{Ours}}
        \end{subfigure}
        \begin{subfigure}{0.69\textwidth}
            \centering
            \begin{subfigure}{0.32\textwidth}
                \includegraphics[width=\linewidth]{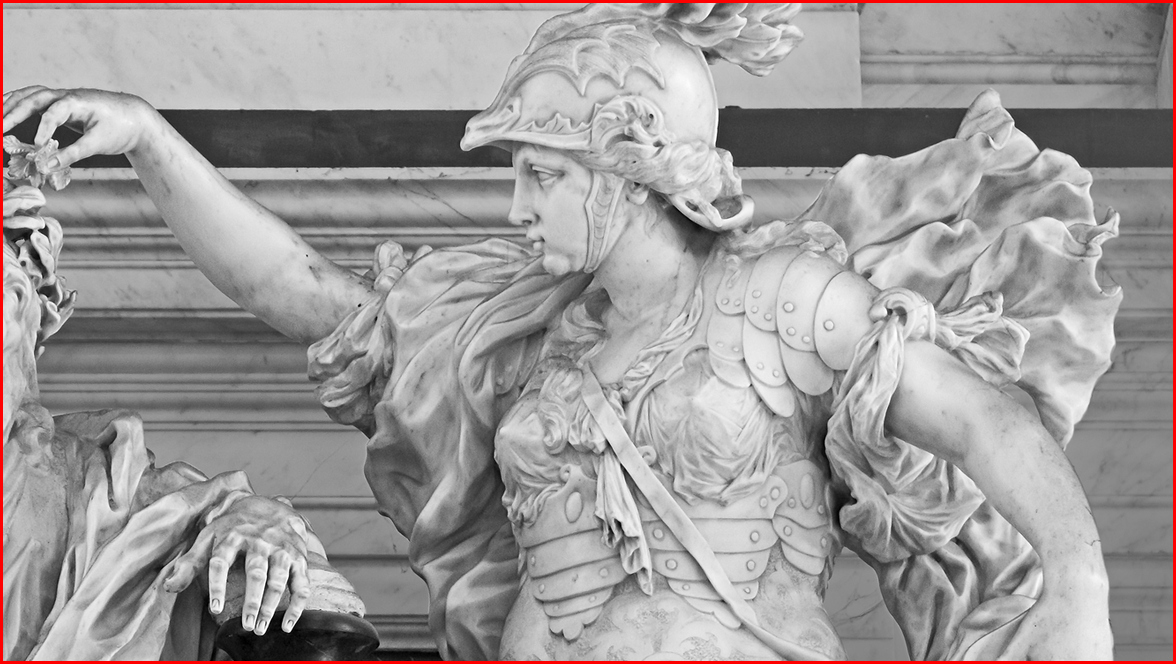}
                \caption*{LQ Input}
            \end{subfigure}
            \begin{subfigure}{0.32\textwidth}
                \includegraphics[width=\linewidth]{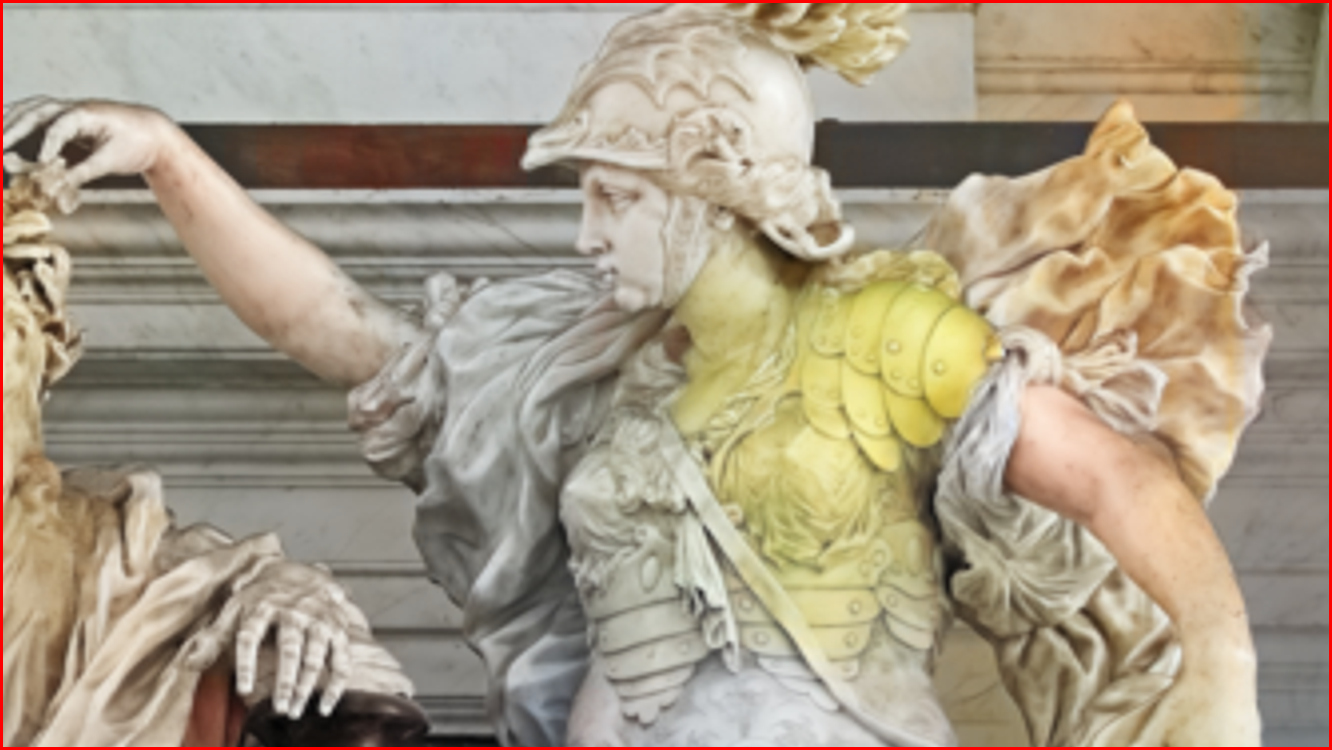}
                \caption*{InstColor}
            \end{subfigure}
            \begin{subfigure}{0.32\textwidth}
                \includegraphics[width=\linewidth]{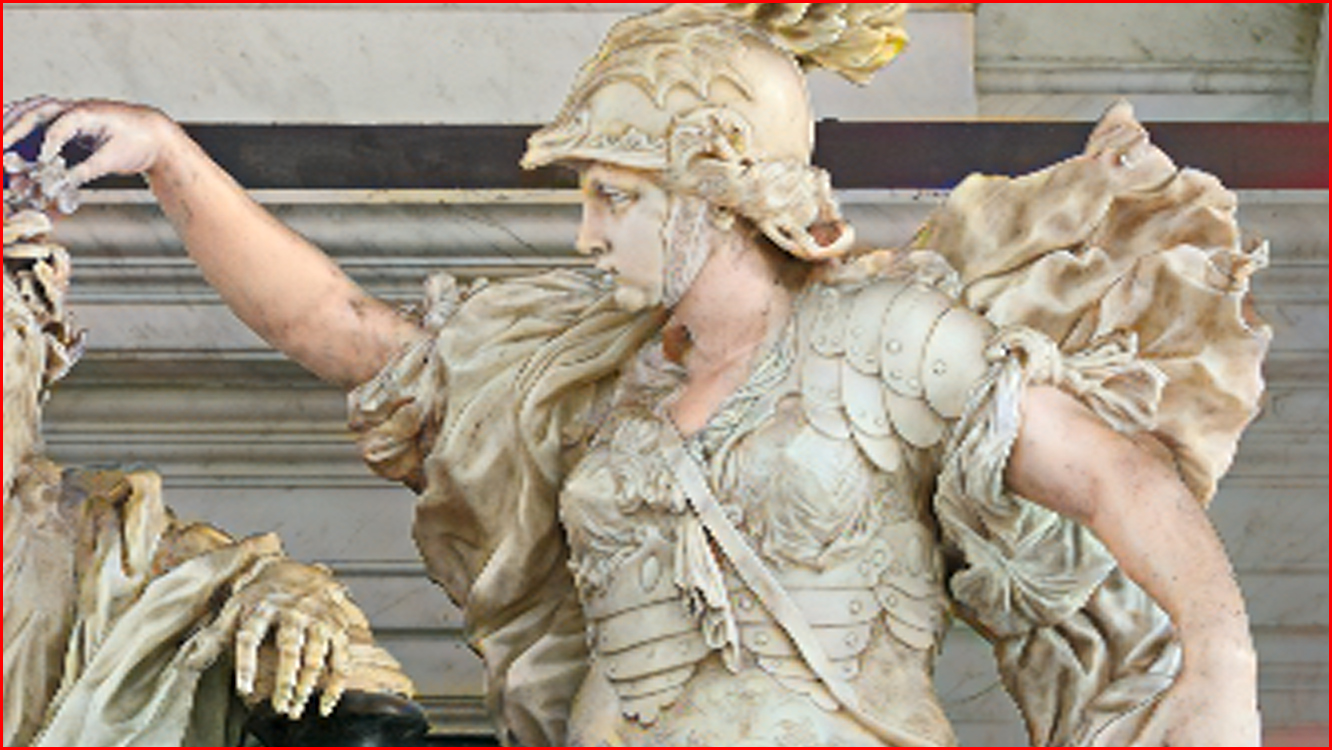}
                \caption*{ColorFormer}
            \end{subfigure}
            \\
            \begin{subfigure}{0.32\textwidth}
                \includegraphics[width=\linewidth]{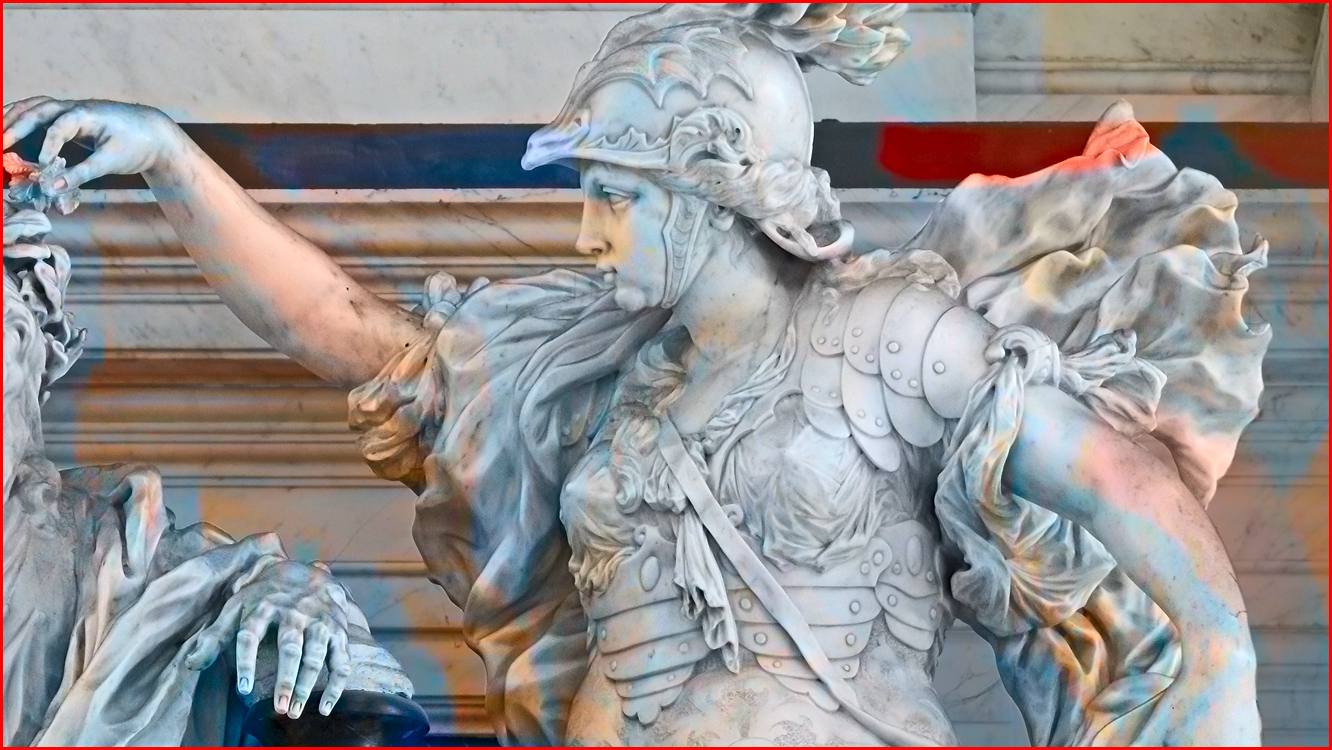}
                \caption*{BigColor}
            \end{subfigure}
            \begin{subfigure}{0.32\textwidth}
                \includegraphics[width=\linewidth]{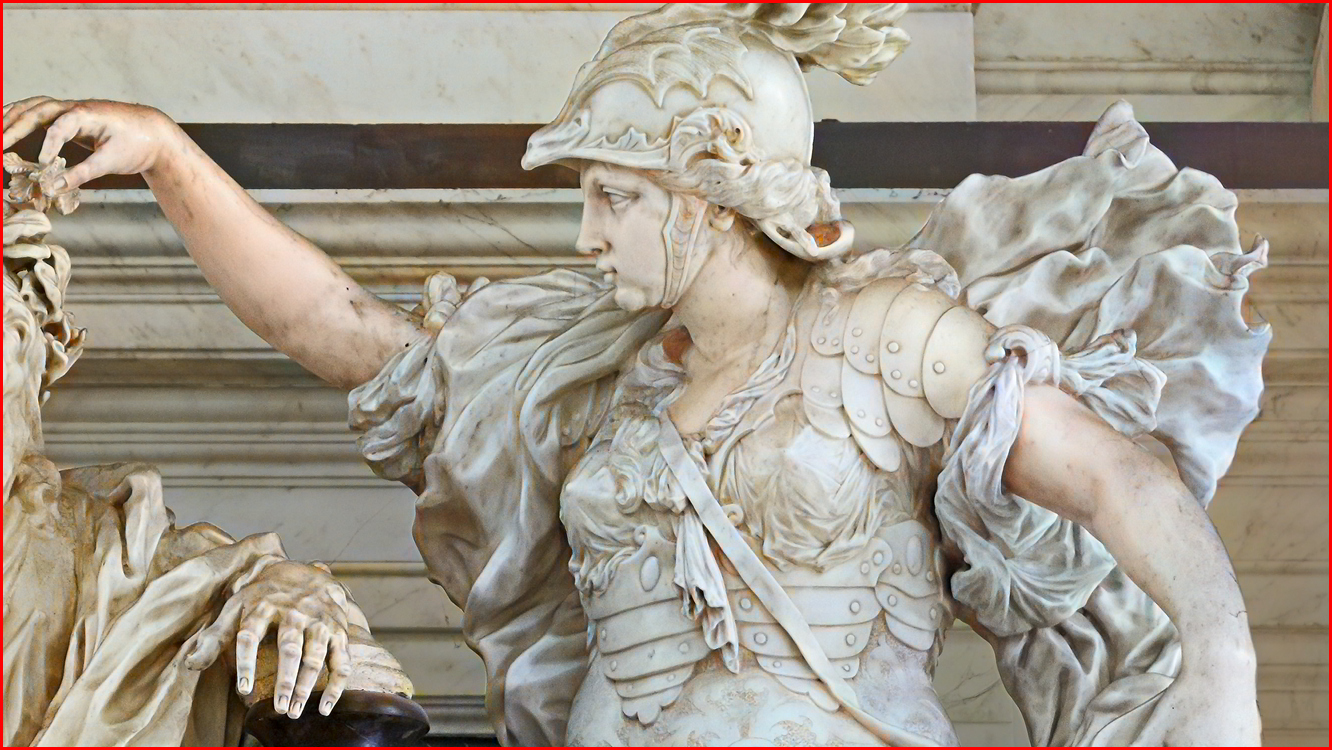}
                \caption*{DDColor}
            \end{subfigure}
            \begin{subfigure}{0.32\textwidth}
                \includegraphics[width=\linewidth]{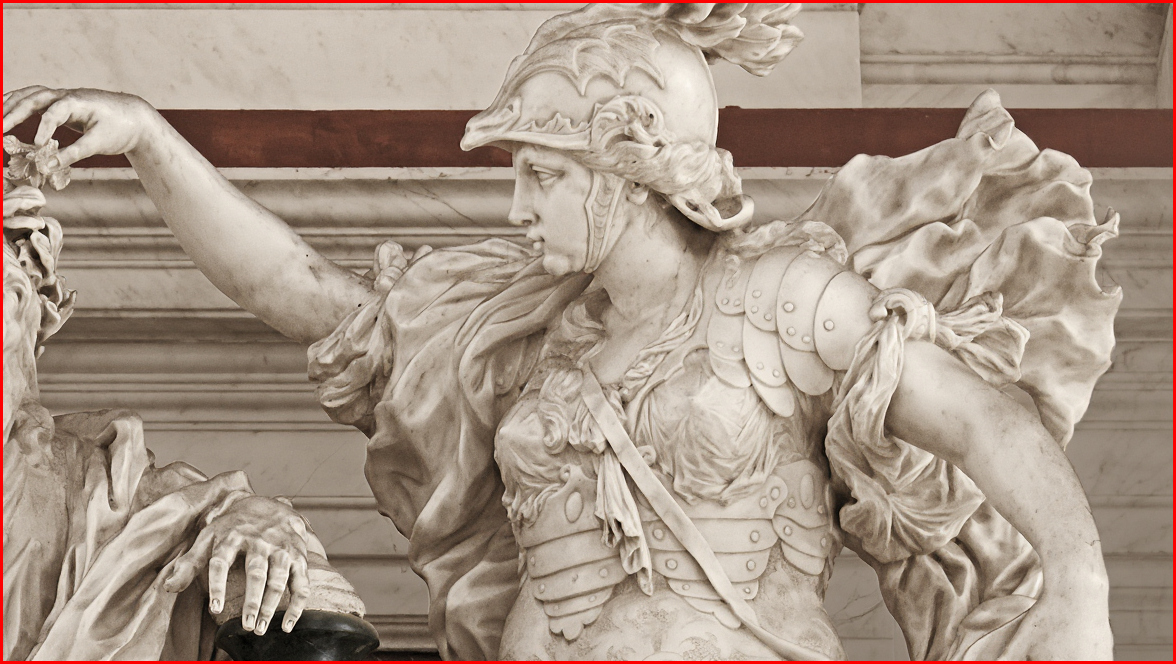}
                \caption*{\textbf{Ours}}
            \end{subfigure}
        \end{subfigure}
    \end{subfigure}
    \caption{\textbf{Qualitative comparisons on \textit{DIV2K-Val} (colorization).}
    Left: our full-resolution result with crop locations marked (red boxes).
    Right: grayscale input and outputs from InstColor, ColorFormer, BigColor, DDColor, and our method on the corresponding crops.
    Our method produces most material-consistent colors while preserving luminance and fine details.}
    \label{fig:ColorDIV2K}
\end{figure}

\newcommand{\colorbest}[1]{\textbf{\underline{#1}}}      
\newcommand{\colorsecond}[1]{\textbf{#1}} 
\newcommand{\NA}{\textemdash}                 

\begin{table}[t]
  \centering
  \caption{\textbf{Colorization on \textit{DIV2K-Val} and \textit{LSDIR-Val} (sRGB).}
  We report CF ($\uparrow$), $\Delta$CF ($\downarrow$), PSNR ($\uparrow$), and three no-reference perceptual metrics: MANIQA, CLIPIQA, and MUSIQ ($\uparrow$).
  Among the evaluated baselines under the stated resolution protocol, our separately trained colorization checkpoint attains the highest PSNR and no-reference scores. Colorfulness and no-reference quality do not certify recovery of the original colors.
  Best is \textbf{\underline{bold+underline}} and second best is \textbf{bold}.}
  \label{tab:colorization}
  \renewcommand{\arraystretch}{1.1}       
  \large
  \setlength{\tabcolsep}{4pt}             
  \begin{adjustbox}{max width=\linewidth}
  \begin{tabular}{l *{12}{S[table-format=2.4]} }
    \toprule
    \multirow{2}{*}{Method} &
    \multicolumn{6}{c}{\textit{DIV2K-Val}} &
    \multicolumn{6}{c}{\textit{LSDIR-Val}} \\
    \cmidrule(lr){2-7}\cmidrule(lr){8-13}
     & {CF$\uparrow$} & {$\Delta$CF$\downarrow$} & {PSNR$\uparrow$} & {MANIQA$\uparrow$} & {CLIPIQA$\uparrow$} & {MUSIQ$\uparrow$}
     & {CF$\uparrow$} & {$\Delta$CF$\downarrow$} & {PSNR$\uparrow$} & {MANIQA$\uparrow$} & {CLIPIQA$\uparrow$} & {MUSIQ$\uparrow$} \\
    \midrule
    \textit{InstColor}
      & 31.41392 & 19.83675 & 20.89596 & 0.388627 & 0.318372 & 31.92938
      & 31.59146 & 30.05199 & 19.41236 & 0.531498 & 0.410991 & 51.28417 \\

    \textit{ColorFormer}
      & 42.52021 & \mch{\best{12.91777}} & 20.61065 & 0.46678 & 0.370297 & 39.32474
      & \secbest{44.05395} & \secbest{21.94203} & 19.3988 & 0.575833 & 0.484854 & 58.07881 \\

    \textit{BigColor}
      & \secbest{43.05025} & 14.61973 & 20.45271 & 0.553216 & 0.423092 & 62.82568
      & 43.68117 & 25.68831 & 19.33953 & 0.625123 & 0.538906 & 70.76193 \\

    \textit{DDColor}
      & \mch{\best{48.87888}} & \secbest{14.09465} & \secbest{21.90095} & \secbest{0.55508} & \secbest{0.44811} & \mch{\secbest{64.18906}}
      & \mch{\best{53.91941}} & \mch{\best{17.9724}} & \mch{\secbest{20.54543}} & \mch{\secbest{0.631816}} & \mch{\secbest{0.565078}} & \mch{\secbest{71.47521}} \\

    \rowcolor{gray!8}
    \textbf{Ours}
      & 29.34042548 & 19.01019123 & \mch{\best{22.83509711}} & \mch{\best{0.613738337}} & \mch{\best{0.527565865}} & \best{64.44354279}
      & 37.64891811 & 25.02170636 & \mch{\best{20.82177372}} & \mch{\best{0.683847314}} & \mch{\best{0.64668608}} & \mch{\best{71.77491945}} \\
    \bottomrule
  \end{tabular}
\end{adjustbox}
\end{table}

\paragraph{Baselines and protocol.}
We compare to four recent automatic colorization systems with public code and checkpoints:
DDColor~\cite{DDColor}, BigColor~\cite{BigColor}, ColorFormer~\cite{ColorFormer}, and InstColor~\cite{InstColor}.
Some baselines enforce a low fixed test resolution (e.g., $256{\times}256$ in ColorFormer~\cite{ColorFormer} and InstColor~\cite{InstColor}).
To reduce resolution-induced bias while respecting their codebases, we run those models at their maximum stable resolution ($512{\times}512$) and bicubic-resize back to the input size for evaluation.

\noindent\textbf{Quantitative.}
Table~\ref{tab:colorization} shows the highest PSNR and all three no-reference scores for our separately trained colorization checkpoint among the evaluated baselines on both datasets under the stated resolution protocol. Some baselines have higher CF or smaller $\Delta$CF; these measures capture different aspects of colorfulness and do not alone determine color accuracy. Grayscale-to-color reconstruction is intrinsically ambiguous, and no-reference metrics cannot establish recovery of the original colors. The results support applicability to colorization under this evaluation, not universal superiority across resolutions or datasets.

\noindent\textbf{Qualitative.}
Figure~\ref{fig:ColorDIV2K} shows visual comparisons on DIV2K-Val.
Our method generates realistic, material-consistent colors while preserving grayscale textures and lighting.
Other approaches may introduce over-saturated tones or unnatural hues (e.g., bluish casts on book spines or skin-tone tints on marble statues), which can compromise global coherence.
The zoomed regions highlight our better detail fidelity, restrained saturation, and more coherent scene appearance.
Additional examples are provided in Appendix~\ref{app:ColorLSDIR}.

\subsection{Ablation Studies}
\label{sec:ablation}

All ablations are conducted on \textit{DIV2K-Val} at the D2 level in the SR\&denoising setting, under strictly matched training/inference budgets and fixed random seeds.
Overall, the full model with pure noise generation, in-token alignment, and DLG with both \emph{Text Emb.} ($e_t$) and \emph{LQ Emb.} ($e_i$), offers the best results both on full-reference and no-reference metrics.

\begin{figure}[htbp]
    \centering
    \begin{subfigure}{0.9\textwidth}
        \centering
        \begin{subfigure}{0.19\textwidth}
            \includegraphics[width=\linewidth]{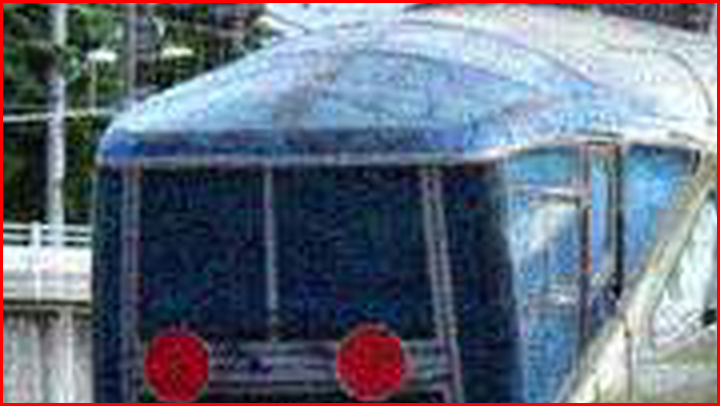}
            \caption*{LQ Input}
        \end{subfigure}
        \begin{subfigure}{0.19\textwidth}
            \includegraphics[width=\linewidth]{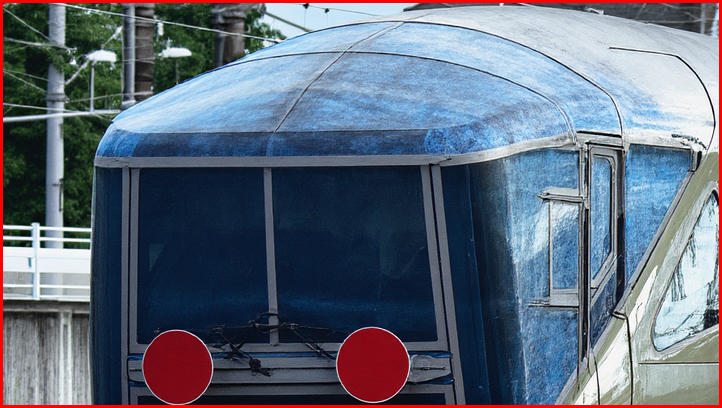}
            \caption*{Text \& LQ Emb.}
        \end{subfigure}
        \begin{subfigure}{0.19\textwidth}
            \includegraphics[width=\linewidth]{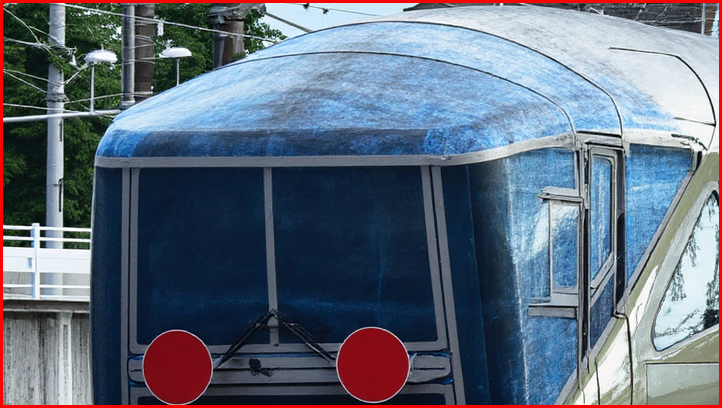}
            \caption*{Text Emb.}
        \end{subfigure}
        \begin{subfigure}{0.19\textwidth}
            \includegraphics[width=\linewidth]{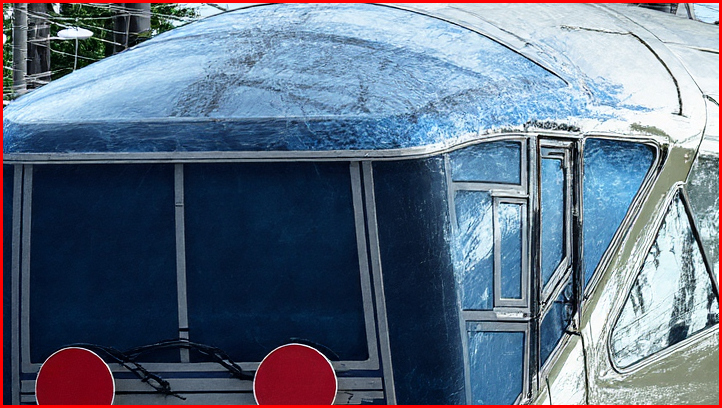}
            \caption*{LQ Emb.}
        \end{subfigure}
        \begin{subfigure}{0.19\textwidth}
            \includegraphics[width=\linewidth]{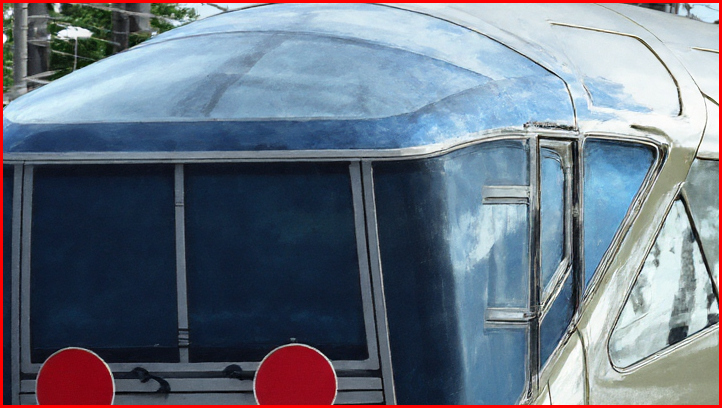}
            \caption*{Blank}
        \end{subfigure}
        \\
        \begin{subfigure}{0.19\textwidth}
            \includegraphics[width=\linewidth]{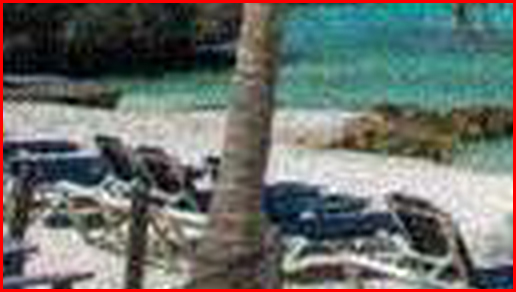}
            \caption*{LQ Input}
        \end{subfigure}
        \begin{subfigure}{0.19\textwidth}
            \includegraphics[width=\linewidth]{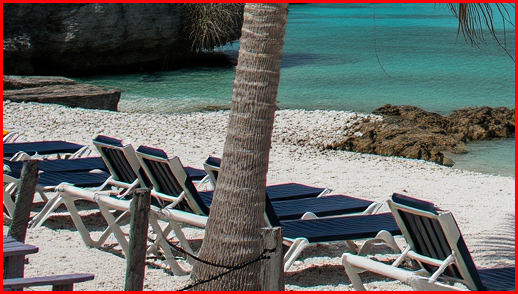}
            \caption*{Pure noise}
        \end{subfigure}
        \begin{subfigure}{0.19\textwidth}
            \includegraphics[width=\linewidth]{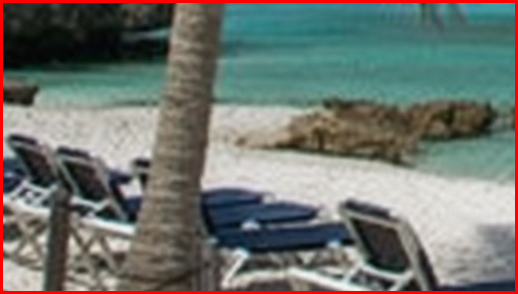}
            \caption*{Denoise 0.9}
        \end{subfigure}
        \begin{subfigure}{0.19\textwidth}
            \includegraphics[width=\linewidth]{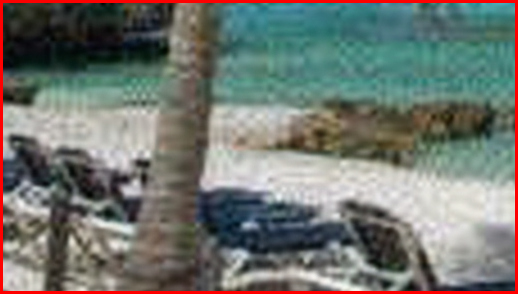}
            \caption*{Denoise 0.6}
        \end{subfigure}
        \begin{subfigure}{0.19\textwidth}
            \includegraphics[width=\linewidth]{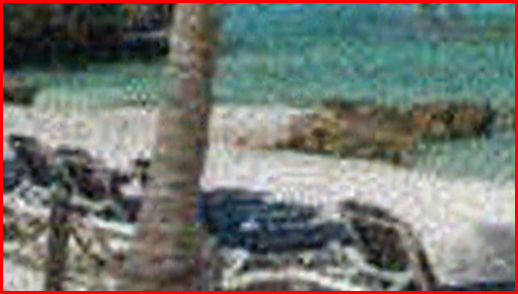}
            \caption*{Denoise 0.3}
        \end{subfigure}
        \\
        \begin{subfigure}{0.32\textwidth}
            \includegraphics[width=\linewidth]{figures/Ablation_Denoise/1/LQ.jpg}
            \caption*{LQ Input}
        \end{subfigure}
        \begin{subfigure}{0.32\textwidth}
            \includegraphics[width=\linewidth]{figures/Ablation_Denoise/1/Ours.jpg}
            \caption*{w/ In-Token}
        \end{subfigure}
        \begin{subfigure}{0.32\textwidth}
            \includegraphics[width=\linewidth]{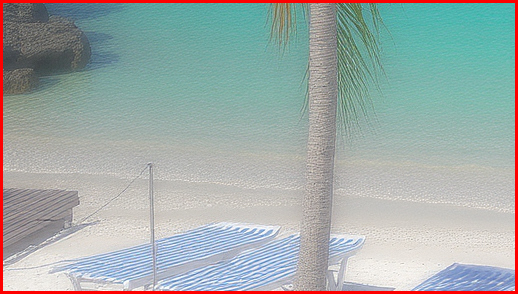}
            \caption*{w/o In-Token}
        \end{subfigure}
    \end{subfigure}
    \caption{\textbf{Ablations for SR \& denoising.}
    \emph{Top:} DLG variants show that combining text and LQ embeddings (Text \& LQ Emb.) yields the best structure and detail.
    \emph{Middle:} starting from pure noise performs better than denoising-initialization at varying strengths (0.9/0.6/0.3).
    \emph{Bottom:} removing \textbf{In-Token} alignment (w/o In-Token) degrades both structure and perceptual quality.}
    \label{fig:Ablation}
\end{figure}

\paragraph{Ablation of DLG.}
We decompose DLG into \emph{Text Emb.} (global task prior) and \emph{LQ Emb.} (instance evidence) and evaluate the $2{\times}2$ combinations (Fig.~\ref{fig:Ablation}, Tab.~\ref{tab:dlg_div2k}).
\textbf{Blank} yields blurry, incomplete results (the train wiper disappears), hurting both full-/no-reference metrics.
\textbf{LQ only} is sharper but “dirty”: noise in the LQ input bends the wiper and suppresses PSNR/SSIM, even as perceptual quality may rise (e.g., CLIPIQA $0.510{\to}0.598$).
\textbf{Text only} is clean yet over-smoothed; roof seams and the wiper remain implausible. Slightly improve full-/no-reference metrics.
\textbf{DLG (Text+LQ)} uses the text prior to constrain and denoise LQ evidence, anchoring structure while filtering artifacts; both full-/no-reference metrics improve (e.g., PSNR $21.98\,\mathrm{dB}$), producing the best results.

\paragraph{Ablation of Generation Regime: pure noise vs.\ denoising.}
Our in-token design allows generating the restored image directly from pure noise, bypassing the classical “denoise the LQ” route.
We vary denoising strength $\{0.30,0.60,0.90\}$ while keeping the schedule fixed (Fig.~\ref{fig:Ablation}, Tab.~\ref{tab:noise_denoise_div2k}).
Even with strong denoising ($0.90$), anchoring to flawed LQ inputs depresses perceptual quality (CLIPIQA $0.363$ vs.\ $0.629$ for pure noise) and leaves a large PSNR gap ($16.41$ vs.\ $21.98$\,dB).
Increasing the tested denoising strength moderately improves PSNR/SSIM and CLIPIQA within this ablation, but all three settings remain below pure-noise initialization. These observations apply to the reported checkpoint and sampling protocol, rather than establishing a general disadvantage of diffusion-based restoration.

\paragraph{Ablation of In-Token Alignment.}
Finally, removing token-level alignment during training and inference causes a substantial drop in fidelity and structural coherence, degrading both full-reference and no-reference metrics (Fig.~\ref{fig:Ablation}, Tab.~\ref{tab:token_alignment_div2k}).
With alignment enabled, the model preserves the fidelity, underscoring that token-aligned supervision is central to our approach.

\begin{table*}[t]
  \centering
  \renewcommand{\arraystretch}{1.05}

  \begin{minipage}[t]{0.48\linewidth}
    \captionsetup{type=table}
    \caption{\textbf{DLG ablation on \textit{DIV2K-Val} (D2).} We toggle \emph{Text Emb.} and \emph{LQ Emb.} in DLG.}
    \label{tab:dlg_div2k}
    \vspace{2pt}
    \begin{adjustbox}{max width=\linewidth}
      \begin{tabular}{cc
                      S[table-format=2.2]
                      S[table-format=1.4]
                      S[table-format=1.3]}
        \toprule
        Text Emb. & LQ Emb. & {PSNR$\uparrow$} & {SSIM$\uparrow$} & {CLIPIQA$\uparrow$} \\
        \midrule
        \xmark & \xmark & 15.5284 & 0.3811 & 0.5102 \\
        \cmark & \xmark & 16.2637 & 0.4271 & 0.5623 \\
        \xmark & \cmark & 15.0445 & 0.3408 & 0.5984 \\
        \cmark & \cmark & \textbf{21.9795} & \textbf{0.5698} & \textbf{0.6293} \\
        \bottomrule
      \end{tabular}
    \end{adjustbox}
  \end{minipage}
  \hfill
  \begin{minipage}[t]{0.48\linewidth}
    \captionsetup{type=table}
    \caption{\textbf{Generation regime on \textit{DIV2K-Val} (D2).} Pure noise vs.\ denoising initialization.}
    \label{tab:noise_denoise_div2k}
    \vspace{2pt}
    \begin{adjustbox}{max width=\linewidth}
      \begin{tabular}{c c
                      S[table-format=2.2]
                      S[table-format=1.4]
                      S[table-format=1.3]}
        \toprule
        Pure noise & Denoising & {PSNR$\uparrow$} & {SSIM$\uparrow$} & {CLIPIQA$\uparrow$} \\
        \midrule
        \xmark & 0.30        & 16.0334 & 0.3218 & 0.2072 \\
        \xmark & 0.60        & 16.2681 & 0.3565 & 0.2240 \\
        \xmark & 0.90        & 16.4119 & 0.4432 & 0.3633 \\
        \cmark & \textemdash{}  & \textbf{21.9795} & \textbf{0.5698} & \textbf{0.6293} \\
        \bottomrule
      \end{tabular}
    \end{adjustbox}
  \end{minipage}
\end{table*}

\begin{table*}[t]
  \centering
  \caption{\textbf{Effect of in-token alignment on \textit{DIV2K-Val} (D2).} Enabling alignment substantially improves both full-reference (PSNR/SSIM) and no-reference(CLIPIQA/MUSIQ/MANIQA) metrics.}
  \label{tab:token_alignment_div2k}
  \vspace{2pt}
  \renewcommand{\arraystretch}{1.05}
  \begin{adjustbox}{max width=0.72\linewidth}
    \begin{tabular}{c
                    S[table-format=2.4]
                    S[table-format=1.4]
                    S[table-format=1.4]
                    S[table-format=2.4]
                    S[table-format=1.4]}
      \toprule
      In-Token Alignment & {PSNR$\uparrow$} & {SSIM$\uparrow$} & {CLIPIQA$\uparrow$} & {MUSIQ$\uparrow$} & {MANIQA$\uparrow$}\\
      \midrule
      \xmark & 11.8003 & 0.2932 & 0.4538 & 47.1899 & 0.5417 \\
      \cmark & \textbf{21.9795} & \textbf{0.5698} & \textbf{0.6293}  & \textbf{66.8804} & \textbf{0.6176} \\
      \bottomrule
    \end{tabular}
  \end{adjustbox}
\end{table*}

\clearpage
\section{Conclusion}

In-Token Learning adapts a pretrained inpainting diffusion transformer with paired conditional flow supervision, channel-aligned evidence, and DLG. Two separately trained checkpoints demonstrate the same framework on SR/denoising and automatic colorization. The original results show competitive performance but not uniformly best scores, and RealLQ250 exposes a synthetic-to-real generalization limitation. Native QHD inference and a tiled 12K demonstration are distinct capabilities; attention cost increases with spatial resolution and neither conditioning nor tiling guarantees faithful detail recovery. This report preserves the broad early study, including colorization, that preceded the real-world SR-focused Fill2SR work. Better degradation coverage and evaluation of hallucinations and tile consistency remain important future directions.

\subsubsection*{Use of LLMs}
A large language model assisted with preparation of this technical report, including wording, mathematical notation, and consistency checks. No new experiments, experimental values, or image results were generated for this revision. The authors are responsible for verifying and approving the final content.


\subsubsection*{Reproducibility Statement}
All numerical results and comparison images in this report are retained from the early study. SR/denoising and colorization require different task-trained checkpoints. At preparation of this report, the original anonymous code/checkpoint repository was not publicly accessible, and the original external full-resolution demo link could not be verified. We therefore do not advertise either as a working reproduction entry point. A currently verified public release of the early training/inference code and both checkpoints remains unavailable. The separate Fill2SR project page, \url{https://github.com/Xingfu-Yi/Fill2SR}, concerns the later SR work and must not be interpreted as a release of this report's colorization model.

\subsubsection*{Ethics Statement}
This work uses only publicly available datasets obtained from open Internet sources, and does not involve any private or sensitive data.

\bibliography{references}
\bibliographystyle{report}

\appendix

\newpage
\section{Additional Theoretical Notes}
\label{app:theory}

\subsection{Rectified Flow Matching}
Our In-Token Learning framework is trained via rectified flow matching~\cite{lipman2022flow, liu2022flow, esser2024scaling}, 
which formulates image restoration as learning a time-dependent velocity field 
that models transport from noise to a distribution of clean latents conditioned on the degraded observation. For a fixed observation and initial noise, the ODE gives a deterministic sampling trajectory; different seeds can give different plausible restorations, not a unique inverse of the degradation.

\paragraph{Forward Process.}
Let $x_0 \sim p_\text{data}$ be the clean target latent and $\epsilon \sim \mathcal{N}(0, \mathrm{I})$ Gaussian noise. 
Rectified flow defines a linear interpolation path between $x_0$ and $\epsilon$:
\begin{equation}
    z_t = (1-t) x_0 + t \epsilon, \quad t \in [0, 1],
\end{equation}
where $z_0 = x_0$ and $z_1 = \epsilon$. 
This deterministic path simplifies the generative process into an ODE trajectory in latent space~\cite{lipman2022flow}.

\paragraph{Velocity Field and Training Target.}
The model $v_\theta(z_t, t, h_t, e_f)$ predicts the velocity that transports $z_t$ towards $x_0$.
The ground-truth velocity is:
\begin{equation}
    \frac{d}{dt} z_t = \epsilon - x_0.
\end{equation}

Hence, the training objective is the $\ell_2$ regression:
\begin{equation}
    \mathcal{L}_\theta 
    = \mathbb{E}_{t,z_0,\epsilon,I_{\mathrm{lq}}}
    \Big[ 
        \| v_\theta(z_t, t, h_t, e_f) - (\epsilon - x_0) \|_2^2
    \Big].
    \label{eq:flow_matching_loss}
\end{equation}

\paragraph{Conditioning with In-Token Learning.}
Our model conditions the velocity field on two sources of information:  
(1) the in-token fusion $h_t = [x_t; y]$, where degraded tokens $y$ are concatenated with noisy latent tokens $x_t$ along the channel dimension; and  
(2) the fused embedding $e_f$ of the system prompt and low-quality image introduced by DLG.  
Formally, the conditional velocity predictor is defined as
\begin{equation}
    v_\theta(z_t, t, h_t, e_f) \approx v^*(z_t, t\mid I_{\mathrm{lq}}),
\end{equation}
where the population squared-loss optimum is $v^*(z_t,t\mid I_{\mathrm{lq}})=\mathbb{E}[\epsilon-x_0\mid z_t,t,I_{\mathrm{lq}}]$, with the task prompt fixed for the checkpoint. The training target is a paired-sample path velocity, whereas the learned field approximates its conditional expectation. Structural and semantic cues are supplied at all timesteps without guaranteeing exact recovery.

\subsection{In-Token Conditioning with RoPE}
We use Rotary Positional Embeddings (RoPE)~\cite{su2024roformer} in spatial form $(h,w)$ for both $x_t$ and $y$, 
since channel-wise fusion keeps the sequence length unchanged:
\begin{equation}
    h_t = [x_t; y] \in \mathbb{R}^{N \times 2d}.
\end{equation}
This preserves the spatial grid size while temporal information is still provided by the standard timestep embedding.  
As a result, the Transformer can naturally attend to location-wise correspondences, 
while channel fusion provides redundancy for detail preservation.

\subsection{Training Dynamics}
The rectified flow loss in Eq.~\ref{eq:flow_matching_loss} yields gradients:
\begin{align}
    \nabla_\theta \mathcal{L}_\theta 
    &= \mathbb{E}_{t,z_0,\epsilon,I_{\mathrm{lq}}}
    \Big[ 
        2 \big(v_\theta(z_t, t, h_t, e_f) - (\epsilon - x_0)\big) 
        \nabla_\theta v_\theta(z_t, t, h_t, e_f)
    \Big].
\end{align}

\noindent
Intuitively:  
- Small $t$ emphasizes fine detail reconstruction from $x_0$;  
- Large $t$ encourages noise suppression and structural guidance from $y$.  

This is an intuition about the noise level, not a demonstrated separation of learned functions across timesteps. During backward sampling, early steps have large $t$ and later steps small $t$; both $y$ and $e_f$ are supplied throughout.

\section{Inference Details}
\label{app:inference}

\subsection{Direct QHD Inference}
For resolutions up to QHD ($2560\times1440$), we run an ODE sampler on full images with the same in-token fusion at each step.

\subsection{Tiled Ultra-High Resolution Inference}
We detail overlapping-tile inference for 4K/8K/12K outputs without modifying the backbone architecture. This is not full-image ultra-high-resolution attention.
For higher resolutions we tile $z_{\mathrm{lq}}$ in latent space and tile $I_{\mathrm{lq}}$ in image space to obtain the corresponding tiled embeddings $e_i^{(\mathrm{tile})}$. 
At each step $t$:  
(1) split $z_t$ and $z_{\mathrm{lq}}$ into overlapping crops (stride $s$, overlap $o$), yielding $\{z_t^{(\mathrm{tile})}, z_{\mathrm{lq}}^{(\mathrm{tile})}\}$;  
(2) crop $I_{\mathrm{lq}}$ accordingly to compute $e_i^{(\mathrm{tile})}$;  
(3) build a fused embedding $e_f^{(\mathrm{tile})}=[e_t; e_i^{(\mathrm{tile})}]$, where $e_t$ is the global system prompt shared across tiles;  
(4) patch $z_t^{(\mathrm{tile})}$ and $z_{\mathrm{lq}}^{(\mathrm{tile})}$ into tokens $\{x_t^{(\mathrm{tile})}, y^{(\mathrm{tile})}\}$, forming $h_t^{(\mathrm{tile})}=[x_t^{(\mathrm{tile})}; y^{(\mathrm{tile})}]$;  
(5) update each crop with one denoising step and blend them with smooth overlaps to assemble the global latent $\tilde z_{t-\Delta t}$.  

After reaching $t{=}0$, we decode with $D(\cdot)$.  
Global $e_t$ and tile-local $e_i^{(\mathrm{tile})}$ supply task and local semantic cues. Overlap blending reduces visible boundaries, but seams and inconsistencies across distant tiles can remain; no quantitative guarantee of tile consistency is claimed.

\section{Complexity Analysis}
\label{app:complexity}

Let the latent spatial grid have size $H_\ell{\times}W_\ell$ and token patch size $p_\ell$, so $N=H_\ell W_\ell/p_\ell^2$. Let $h$ be the number of heads, $d_k$ the per-head backbone width, $n_f=n_t+n_i$ the semantic-token count, and $L$ the layer count. In the following idealized dense-attention comparison, backbone widths, layers, and precision are held fixed. It concerns the quadratic attention term, not the complete network's FLOPs or peak VRAM.

\subsection{In-Context vs. In-Token Conditioning}
We compare the FLOPs and memory scaling of sequence-level and channel-level fusion. 

\paragraph{In-context visual conditioning (sequence concat).}
The degraded image is appended as another $N$ tokens (plus $n_f$ prompt tokens), giving length $2N{+}n_f$:
\[
\mathrm{FLOPs}_{\text{ctx}}\ \propto\ L\,h\,(2N{+}n_f)^2\,d_k,\qquad
\mathrm{Mem}_{\text{ctx}}\ \propto\ (2N{+}n_f)^2 .
\]

\paragraph{In-token learning (channel concat).}
Channel fusion keeps $N$ image tokens and projects the fused input into the existing backbone width. The $2d$ pre-projection input width should not be substituted for the backbone attention width:
\[
\mathrm{FLOPs}_{\text{token}}\ \propto\ L\,h\,(N{+}n_f)^2\,d_k,\qquad
\mathrm{Mem}_{\text{token}}\ \propto\ (N{+}n_f)^2 .
\]

\subsection{Implications for Scalability}
For $n_f\ll N$, this idealized attention-arithmetic ratio simplifies to
\[
\frac{\mathrm{FLOPs}_{\text{ctx}}}{\mathrm{FLOPs}_{\text{token}}}\ \approx\ \frac{(2N)^2\,d_k}{N^2\,d_k}\ =\ 4.
\]
The number of entries in explicitly materialized attention matrices has the same approximate ratio. Memory-efficient attention need not store those matrices, so this does not establish a fourfold peak-memory saving. Input projections, MLPs, frozen encoders, and implementation choices also contribute to measured costs.
At a fixed pixel-to-latent conversion, $N$ grows with image area; neither fusion removes the resolution dependence. Tiled inference bounds each forward pass's spatial token count at the cost of multiple passes and weaker global context. Reported QHD and tiled 12K demonstrations should be interpreted within those limits.

\section{Additional Analyses and Results}
\label{app:result}

\paragraph{Inference precision.}
We study the quality--efficiency trade-off at inference time by comparing \textbf{bf16} and \textbf{qfloat8}~\cite{quanto}. 
For each precision, we report restoration quality (PSNR/SSIM, CLIPIQA, MUSIQ, MANIQA) together with the peak GPU memory requirement (VRAM) measured at \textbf{QHD (1440p)} inference. 
On \textit{DIV2K-Val} (D2), qfloat8 reduces peak VRAM by $\sim$43\% (30$\rightarrow$17\,GB) but consistently lowers quality across all metrics (Tab.~\ref{app:precision}). 
This clarifies when ultra-low precision is attractive—memory-bound scenarios—and when bf16 is preferable for quality-critical use cases.

\paragraph{SuperResolution\&denoising.} We first show the detailed per-level quantitative result, showing more baselines. We then show more visualizations of the validation set of DIV2K on degradation level D3. We therefore show case the low quality input at different level of degradation, and give a vivide comparison of the methods ability when degradation increases. Finally, we give more detailed visualizations on RealPhoto60, showing our methods ability.

\paragraph{Colorization.}
We present extended side-by-side visual comparisons against recent methods. 
Our approach produces sharper structures and more faithful colors while avoiding common artifacts such as over-saturation and hue shifts.

\newcommand{\blocksep}{\cmidrule(lr){2-14}} 

\begin{table*}[h]
  \centering
  \caption{
    \textbf{Efficiency comparison with recent diffusion-based restoration models.}
    Our method achieves the \textbf{lowest trainable parameter count} while maintaining competitive performance.
    Inference times at $1536^2$ use RTX 5880 Ada GPUs: SeeSR and ours use one GPU; SUPIR, DreamClear, and FaithDiff use two. These implementation-specific configurations are not a matched-hardware runtime comparison.
  }
  \label{tab:efficiency}
  \vspace{4pt}
  \renewcommand{\arraystretch}{1.10}
  \begin{adjustbox}{max width=0.82\linewidth}
  \begin{tabular}{lccccc}
    \toprule
    \textbf{Method} 
      & \textbf{SeeSR} 
      & \textbf{SUPIR} 
      & \textbf{DreamClear} 
      & \textbf{FaithDiff} 
      & \textbf{Ours} \\
    \midrule
    Trainable Params (B) $\downarrow$ 
      & 2.3 & 1.3 & 2.41 & 2.4 & \textbf{1.0} \\
    Inference Speed (s, $1536^2$) $\downarrow$
      & 34 & \textbf{26} & 107 & 35 & 75 \\
    \bottomrule
  \end{tabular}
  \end{adjustbox}
\end{table*}

\begin{table*}[h]
  \centering
  \caption{\textbf{Inference precision ablation on \textit{DIV2K-Val} (D2).}
  At \textbf{QHD (1440p)} inference, switching from \textbf{bf16} to \textbf{qfloat8} cuts the \emph{peak GPU memory} from 30 to 17\,GB ($\approx$43\%$\downarrow$), 
  but yields lower generation quality across fidelity (PSNR/SSIM) and perceptual metrics (CLIPIQA, MUSIQ, MANIQA).}
  \label{app:precision}
  \vspace{2pt}
  \renewcommand{\arraystretch}{1.05}
  \begin{adjustbox}{max width=0.8\linewidth}
    \begin{tabular}{c
                    S[table-format=2.4]
                    S[table-format=1.4]
                    S[table-format=1.4]
                    S[table-format=2.4]
                    S[table-format=1.4]
                    S[table-format=2.0]}
      \toprule
      Precision & {PSNR$\uparrow$} & {SSIM$\uparrow$} & {CLIPIQA$\uparrow$} & {MUSIQ$\uparrow$} & {MANIQA$\uparrow$} & {VRAM (GB)$\downarrow$}\\
      \midrule
      qfloat8 & 18.5053 & 0.4707 & 0.5096 & 63.8563 & 0.5886 & \textbf{17}\\
      \textbf{bf16} & \textbf{21.9795} & \textbf{0.5698} & \textbf{0.6293}  & \textbf{66.8804} & \textbf{0.6176} & {30} \\
      \bottomrule
    \end{tabular}
  \end{adjustbox}
\end{table*}

\begin{figure}[h]
    \centering
    \begin{subfigure}{0.7\textwidth}
        \centering
        \begin{subfigure}{0.32\textwidth}
            \includegraphics[width=\linewidth]{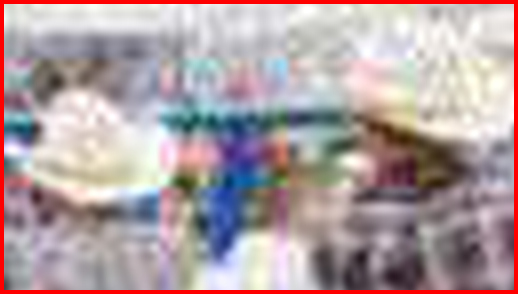}
            \caption*{LQ Input}
        \end{subfigure}
        \begin{subfigure}{0.32\textwidth}
            \includegraphics[width=\linewidth]{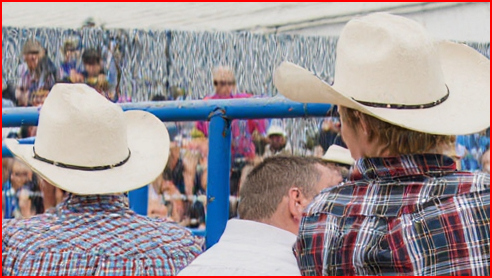}
            \caption*{qfloat8}
        \end{subfigure}
        \begin{subfigure}{0.32\textwidth}
            \includegraphics[width=\linewidth]{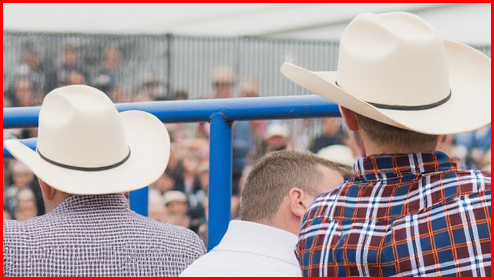}
            \caption*{\textbf{bf16}}
        \end{subfigure}
        \\
        \begin{subfigure}{0.32\textwidth}
            \includegraphics[width=\linewidth]{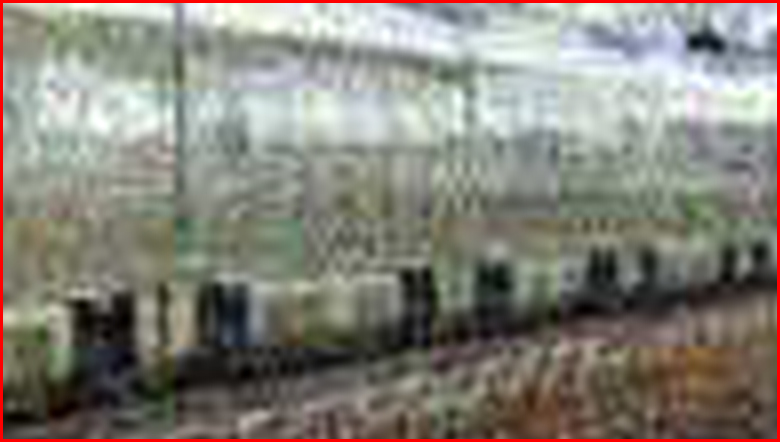}
            \caption*{LQ Input}
        \end{subfigure}
        \begin{subfigure}{0.32\textwidth}
            \includegraphics[width=\linewidth]{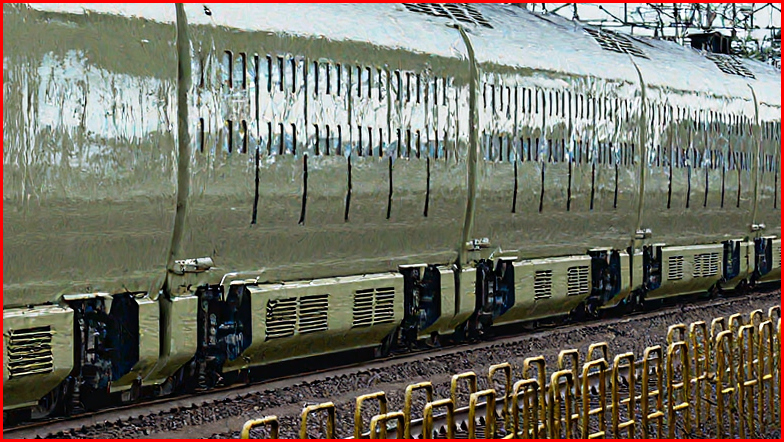}
            \caption*{qfloat8}
        \end{subfigure}
        \begin{subfigure}{0.32\textwidth}
            \includegraphics[width=\linewidth]{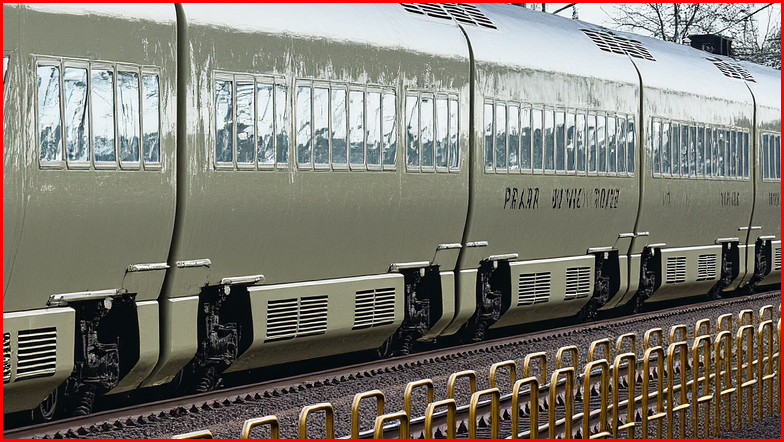}
            \caption*{\textbf{bf16}}
        \end{subfigure}
        \\
        \begin{subfigure}{0.32\textwidth}
            \includegraphics[width=\linewidth]{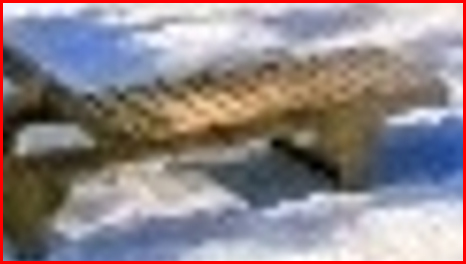}
            \caption*{LQ Input}
        \end{subfigure}
        \begin{subfigure}{0.32\textwidth}
            \includegraphics[width=\linewidth]{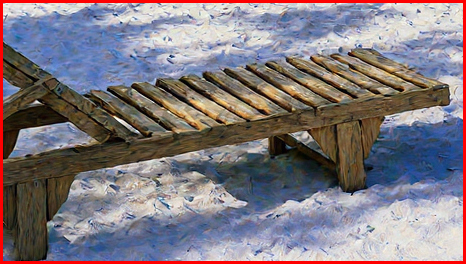}
            \caption*{qfloat8}
        \end{subfigure}
        \begin{subfigure}{0.32\textwidth}
            \includegraphics[width=\linewidth]{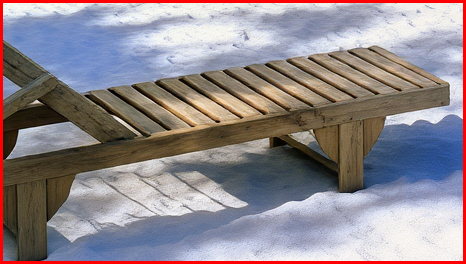}
            \caption*{\textbf{bf16}}
        \end{subfigure}
    \end{subfigure}
\caption{\textbf{Qualitative comparison under different inference precisions (qfloat8 vs. bf16).}
Each triplet shows the low-quality input, our model inferred with \textbf{qfloat8}, and with \textbf{bf16}. 
At \textbf{QHD (1440p)} inference, qfloat8 lowers the \emph{peak} VRAM from \textbf{30} to \textbf{17}\,GB, 
but under challenging settings, it more often suppresses fine structures and introduces spurious “dirty” textures.}
\end{figure}

\begin{table}[h]
  \centering
  \caption{Results on \textit{DIV2K\_VAL}, \textit{LSDIR\_VAL}, and \textit{FFHQ-face} across D1--D3.
  Higher is better for CLIPIQA/MUSIQ/MANIQA/PSNR/SSIM ($\uparrow$); lower is better for LPIPS ($\downarrow$).
  Best is \textbf{\underline{bold+underline}} and second best is \textbf{bold}.}
  \label{app:main_summary}
  \renewcommand{\arraystretch}{1.5}
  \setlength{\tabcolsep}{3.6pt}
  \small
  \begin{adjustbox}{max width=\linewidth}
  \begin{tabular}{l c l *{10}{S[table-format=2.4]}>{\columncolor{gray!8}}S[table-format=2.4]}
      \toprule
      \multirow{2}{*}{Datasets} & \multirow{2}{*}{D-Level} & \multirow{2}{*}{Metrics} & \multicolumn{11}{c}{Methods} \\
      \cmidrule(lr){4-14}
      &  &  &
      \mch{BSRGAN} & \mch{Real-ESRGAN} & \mch{SwinIR} & \mch{DASR} &
      \mch{StableSR} & \mch{DiffBIR} & \mch{SeeSR} & \mch{SUPIR} &
      \mch{DreamClear} & \mch{FaithDiff} & \mch{\textbf{Ours}} \\
      \midrule

      \multirow{18}{*}{\textit{DIV2K-Val}}
        & \multirow{6}{*}{D1} & CLIPIQA$\uparrow$ & 0.492997 & 0.578496 & 0.358100 & 0.566349 & 0.504783 & \best{0.625420} & \secbest{0.610119} & 0.518557 & 0.582970 & 0.575468 & 0.574520 \\
        &  & MUSIQ$\uparrow$  & 65.17959 & 67.78659 & 48.58408 & 67.86055 & 65.59607 & \best{69.34330} & \secbest{69.33327} & 65.53267 & 68.11756 & 68.07780 & 64.71648 \\
        &  & MANIQA$\uparrow$ & 0.573384 & 0.586981 & 0.558631 & 0.570533 & 0.594935 & 0.592093 & 0.602788 & 0.606075 & 0.594615 & \best{0.617272} & \secbest{0.611077} \\
        &  & PSNR$\uparrow$   & \best{27.97127} & 26.06688 & \secbest{26.32204} & 26.29519 & 24.18853 & 25.39995 & 24.51775 & 25.32686 & 23.38142 & 24.37231 & 25.27063 \\
        &  & SSIM$\uparrow$   & \best{0.806798} & 0.767768 & 0.682353 & \secbest{0.779781} & 0.704375 & 0.703634 & 0.673811 & 0.702545 & 0.641350 & 0.660903 & 0.705749 \\
        &  & LPIPS$\downarrow$& \best{0.283191} & \secbest{0.299735} & 0.449307 & 0.318349 & 0.318626 & 0.322640 & 0.343418 & 0.310729 & 0.346408 & 0.330226 & 0.308775 \\
      \blocksep
        & \multirow{6}{*}{D2} & CLIPIQA$\uparrow$ & 0.502481 & 0.475663 & 0.327683 & 0.415530 & 0.373426 & 0.568059 & \secbest{0.594843} & 0.511119 & 0.525974 & 0.581395 & \best{0.629325} \\
        &  & MUSIQ$\uparrow$  & 64.09235 & 62.01011 & 33.97698 & 59.93721 & 53.64849 & 66.04007 & \best{68.83204} & 65.26168 & 66.24704 & \secbest{68.53924} & 66.88035 \\
        &  & MANIQA$\uparrow$ & 0.506247 & 0.523418 & 0.339698 & 0.448943 & 0.481069 & 0.561315 & 0.597710 & 0.592560 & 0.575204 & \best{0.621938} & \secbest{0.617610} \\
        &  & PSNR$\uparrow$   & \best{23.9390} & \secbest{23.2248} & 22.7615 & 23.1437 & 23.6248 & 23.8260 & 23.1446 & 22.7877 & 22.1618 & 22.7779 & 21.9795 \\
        &  & SSIM$\uparrow$   & \best{0.6354} & \secbest{0.6308} & 0.5007 & 0.6210 & 0.6303 & 0.6173 & 0.5998 & 0.5791 & 0.5701 & 0.5692 & 0.5698 \\
        &  & LPIPS$\downarrow$& 0.4107 & 0.4020 & 0.5879 & 0.4258 & 0.3976 & 0.4144 & 0.4043 & 0.3994 & 0.4042 & \secbest{0.3959} & \best{0.3925} \\
        \blocksep
        & \multirow{6}{*}{D3} & CLIPIQA$\uparrow$ & 0.402386 & 0.408444 & 0.242032 & 0.349230 & 0.262123 & 0.527762 & \secbest{0.605608} & 0.485589 & 0.484020 & 0.572152 & \best{0.660179} \\
        &  & MUSIQ$\uparrow$  & 44.78789 & 47.11308 & 32.52163 & 41.50560 & 28.00328 & 63.08860 & \best{68.97384} & 63.04814 & 63.41077 & \secbest{67.71965} & 67.53725 \\
        &  & MANIQA$\uparrow$ & 0.419580 & 0.465234 & 0.183361 & 0.398369 & 0.257336 & 0.559441 & 0.604134 & 0.558746 & 0.553222 & \secbest{0.616063} & \best{0.636789} \\
        &  & PSNR$\uparrow$   & \best{21.3738} & 20.7643 & 20.1131 & 20.6904 & 21.0975 & \secbest{21.2616} & 20.1085 & 19.8682 & 19.5761 & 20.0535 & 19.0548 \\
        &  & SSIM$\uparrow$   & \best{0.5296} & \secbest{0.52547} & 0.4187 & 0.5117 & 0.4908 & 0.4985 & 0.4684 & 0.44267 & 0.4556 & 0.4420 & 0.4541 \\
        &  & LPIPS$\downarrow$& 0.5117 & 0.5031 & 0.6663 & 0.5202 & 0.5977 & 0.5132 & 0.5018 & 0.49927 & 0.4953 & \secbest{0.4892} & \best{0.4737} \\
        \midrule

      \multirow{18}{*}{\textit{LSDIR-Val}}
        & \multirow{6}{*}{D1} & CLIPIQA$\uparrow$ & 0.544298 & 0.640773 & 0.423107 & 0.630301 & 0.564541 & \secbest{0.675880} & 0.649585 & 0.613926 & 0.667404 & 0.643163 & \best{0.711231} \\
        &  & MUSIQ$\uparrow$  & 69.81489 & 72.98232 & 58.12021 & \secbest{73.03713} & 69.77098 & \best{73.07045} & 73.00904 & 71.74105 & 72.30856 & 72.02255 & 72.76724 \\
        &  & MANIQA$\uparrow$ & 0.623156 & 0.653814 & 0.600291 & 0.635968 & 0.639204 & 0.649660 & 0.645121 & 0.669970 & 0.647809 & \secbest{0.672102} & \best{0.688642} \\
        &  & PSNR$\uparrow$   & \best{24.75207} & 23.28721 & \secbest{23.61545} & 23.54955 & 21.34583 & 22.30817 & 21.55450 & 22.24058 & 21.09951 & 21.16418 & 22.733831 \\
        &  & SSIM$\uparrow$   & \best{0.756412} & 0.722017 & 0.649435 & \secbest{0.729450} & 0.624108 & 0.635460 & 0.585941 & 0.643138 & 0.602913 & 0.570211 & 0.680924 \\
        &  & LPIPS$\downarrow$& \best{0.273294} & \secbest{0.280520} & 0.438258 & 0.306198 & 0.316495 & 0.305282 & 0.334964 & 0.301743 & 0.319037 & 0.325596 & 0.281987 \\
      \blocksep
        & \multirow{6}{*}{D2} & CLIPIQA$\uparrow$ & 0.551750 & 0.538725 & 0.326486 & 0.469973 & 0.431033 & 0.631029 & \secbest{0.633736} & 0.595429 & 0.619217 & 0.633718 & \best{0.725827} \\
        &  & MUSIQ$\uparrow$  & 68.74689 & 68.70039 & 37.96491 & 64.65536 & 58.68908 & 70.52980 & \secbest{72.49195} & 70.65330 & 70.85945 & 71.87447 & \best{72.63834} \\
        &  & MANIQA$\uparrow$ & 0.536327 & 0.572817 & 0.329409 & 0.468586 & 0.528721 & 0.608507 & 0.639268 & 0.645247 & 0.622983 & \secbest{0.671575} & \best{0.688535} \\
        &  & PSNR$\uparrow$   & \best{21.1313} & 20.4738 & 20.4985 & 20.4978 & 20.7961 & \secbest{20.9779} & 20.5045 & 19.8372 & 19.5996 & 20.0026 & 19.5960 \\
        &  & SSIM$\uparrow$   & \best{0.5547} & \secbest{0.5496} & 0.4406 & 0.5340 & 0.5378 & 0.5387 & 0.5089 & 0.4797 & 0.5000 & 0.4780 & 0.5131 \\
        &  & LPIPS$\downarrow$& 0.4176 & 0.4092 & 0.5989 & 0.4389 & 0.4046 & 0.4076 & \secbest{0.3977} & 0.4117 & 0.4015 & \best{0.3935} & 0.4863 \\
        \blocksep
        & \multirow{6}{*}{D3} & CLIPIQA$\uparrow$ & 0.426734 & 0.447657 & 0.231604 & 0.382112 & 0.263999 & 0.585838 & \secbest{0.640609} & 0.501722 & 0.549667 & 0.615126 & \best{0.728166} \\
        &  & MUSIQ$\uparrow$  & 48.64420 & 51.99749 & 34.42582 & 46.20688 & 30.17084 & 66.74164 & \secbest{72.54838} & 61.45802 & 67.89442 & 70.99523 & \best{72.57192} \\
        &  & MANIQA$\uparrow$ & 0.417848 & 0.475715 & 0.175508 & 0.379575 & 0.268378 & 0.587838 & 0.631922 & 0.554881 & 0.573894 & \secbest{0.652183} & \best{0.686727} \\
        &  & PSNR$\uparrow$   & \best{18.9619} & 18.42731 & 18.2784 & 18.4354 & \secbest{18.9172} & 18.8488 & 18.1164 & 17.7413 & 17.5175 & 17.7367 & 17.5267 \\
        &  & SSIM$\uparrow$   & \best{0.4303} & \secbest{0.4292} & 0.3398 & 0.4078 & 0.3934 & 0.4119 & 0.3751 & 0.3312 & 0.3740 & 0.3381 & 0.3889 \\
        &  & LPIPS$\downarrow$& 0.5371 & 0.5266 & 0.6896 & 0.5482 & 0.6114 & 0.5139 & \best{0.5051} & 0.5472 & 0.5095 & \secbest{0.5057} & 0.54345 \\
        \midrule

      \multirow{18}{*}{\textit{FFHQ-face}}
        & \multirow{6}{*}{D1} & CLIPIQA$\uparrow$ & 0.541178 & 0.556741 & 0.236626 & 0.499058 & \secbest{0.596170} & \best{0.643611} & 0.586410 & 0.530332 & 0.568995 & 0.563856 & 0.502632 \\
        &  & MUSIQ$\uparrow$  & 74.89174 & 72.23323 & 48.72375 & 71.74989 & \secbest{75.52720} & 75.37485 & 75.51161 & 73.15149 & 74.36445 & \best{75.68991} & 71.84016 \\
        &  & MANIQA$\uparrow$ & 0.572418 & 0.548266 & 0.539077 & 0.528210 & 0.595976 & 0.593524 & 0.600250 & \secbest{0.601134} & 0.588360 & \best{0.633883} & 0.599696 \\
        &  & PSNR$\uparrow$   & \best{32.02855} & 30.91520 & 29.76186 & \secbest{31.11275} & 28.89645 & 29.84026 & 29.43710 & 30.25784 & 28.28038 & 28.63189 & 30.33679 \\
        &  & SSIM$\uparrow$   & \secbest{0.846770} & 0.844782 & 0.703129 & \best{0.850663} & 0.791085 & 0.786584 & 0.785146 & 0.790143 & 0.746659 & 0.743394 & 0.793236 \\
        &  & LPIPS$\downarrow$& \best{0.297552} & 0.319671 & 0.454815 & 0.331761 & \secbest{0.298137} & 0.342617 & 0.318192 & 0.299125 & 0.330949 & 0.320151 & 0.316246 \\
      \blocksep
        & \multirow{6}{*}{D2} & CLIPIQA$\uparrow$ & 0.519826 & 0.426681 & 0.264441 & 0.340110 & 0.515410 & \best{0.625452} & 0.542456 & 0.543303 & 0.480328 & 0.571561 & \secbest{0.591490} \\
        &  & MUSIQ$\uparrow$  & 72.48447 & 64.29935 & 26.49761 & 60.71150 & 71.81948 & 73.94498 & 73.23125 & 74.47312 & 70.59531 & \best{76.18117} & \secbest{74.49006} \\
        &  & MANIQA$\uparrow$ & 0.527772 & 0.481015 & 0.362963 & 0.428620 & 0.534552 & 0.588728 & 0.585324 & 0.603976 & 0.563977 & \best{0.641102} & \secbest{0.615376} \\
        &  & PSNR$\uparrow$   & 28.5038 & 28.2503 & 25.8472 & 28.1374 & \secbest{28.5570} & \best{28.6004} & 27.5172 & 27.7102 & 27.4778 & 27.5357 & 26.9644 \\
        &  & SSIM$\uparrow$   & \secbest{0.7839} & \best{0.7887} & 0.5741 & 0.78308 & 0.7699 & 0.7675 & 0.7591 & 0.7403 & 0.7346 & 0.7232 & 0.7329 \\
        &  & LPIPS$\downarrow$& 0.3797 & 0.3699 & 0.5804 & 0.3801 & 0.3602 & 0.3793 & 0.3783 & \secbest{0.3593} & 0.3763 & 0.3673 & \best{0.3559} \\
        \blocksep
        & \multirow{6}{*}{D3} & CLIPIQA$\uparrow$ & 0.381209 & 0.305809 & 0.236654 & 0.252401 & 0.336460 & \secbest{0.615714} & 0.519018 & 0.516514 & 0.392825 & 0.580188 & \best{0.639518} \\
        &  & MUSIQ$\uparrow$  & 64.32358 & 53.72704 & 27.06953 & 47.15531 & 52.97259 & 72.48646 & 71.40988 & 73.18912 & 67.31593 & \best{76.71128} & \secbest{76.06122} \\
        &  & MANIQA$\uparrow$ & 0.509665 & 0.446793 & 0.183387 & 0.390901 & 0.378267 & 0.595134 & 0.576884 & 0.585848 & 0.543764 & \best{0.646684} & \secbest{0.630166} \\
        &  & PSNR$\uparrow$   & \best{25.9960} & \secbest{25.7447} & 23.0445 & 25.3520 & 24.8549 & 25.8737 & 24.8490 & 24.7313 & 24.4744 & 23.96516 & 24.2035 \\
        &  & SSIM$\uparrow$   & 0.6992 & \best{0.7144} & 0.5433 & \secbest{0.6966} & 0.6263 & 0.6710 & 0.6684 & 0.6210 & 0.62704 & 0.5930 & 0.6330 \\
        &  & LPIPS$\downarrow$& \best{0.4258} & 0.4495 & 0.6197 & 0.4641 & 0.5090 & 0.4584 & \secbest{0.4334} & 0.4351 & 0.4404 & 0.4349 & 0.4354 \\
        \bottomrule
    \end{tabular}
  \end{adjustbox}
\end{table}

\begin{figure}[h]
    \centering
    \begin{subfigure}{\textwidth}
        \centering
        \begin{subfigure}{0.99\textwidth}
            \centering
            \begin{subfigure}{0.32\textwidth}
                \includegraphics[width=\linewidth]{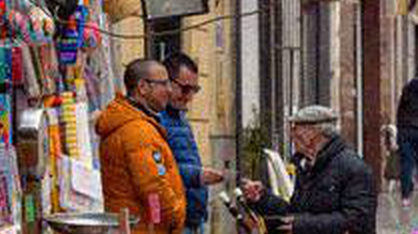}
                \caption*{LQ Input (D1)}
            \end{subfigure}
            \begin{subfigure}{0.32\textwidth}
                \includegraphics[width=\linewidth]{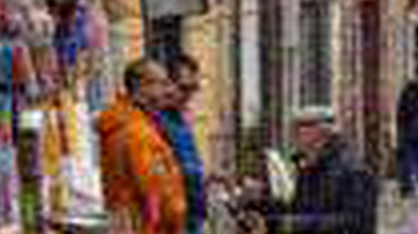}
                \caption*{LQ Input (D2)}
            \end{subfigure}
            \begin{subfigure}{0.32\textwidth}
                \includegraphics[width=\linewidth]{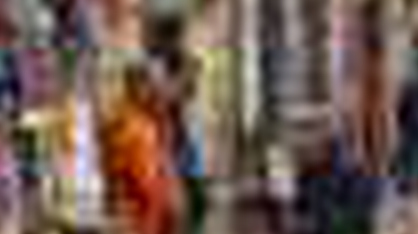}
                \caption*{LQ Input (D3)}
            \end{subfigure}
            \\
            \vspace{1em} 
            \begin{subfigure}{0.32\textwidth}
                \includegraphics[width=\linewidth]{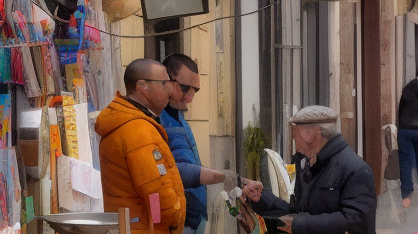}
                \caption*{DiffBIR (D1)}
            \end{subfigure}
            \begin{subfigure}{0.32\textwidth}
                \includegraphics[width=\linewidth]{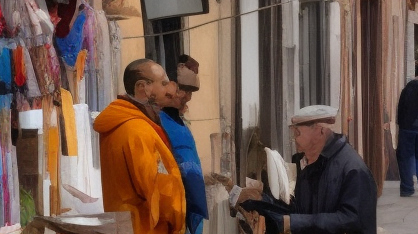}
                \caption*{DiffBIR (D2)}
            \end{subfigure}
            \begin{subfigure}{0.32\textwidth}
                \includegraphics[width=\linewidth]{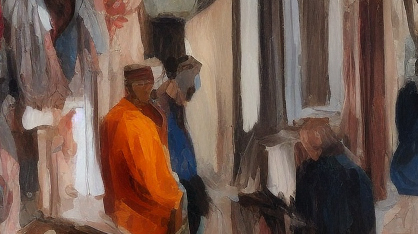}
                \caption*{DiffBIR (D3)}
            \end{subfigure}
            \\
            \vspace{1em} 
            \begin{subfigure}{0.32\textwidth}
                \includegraphics[width=\linewidth]{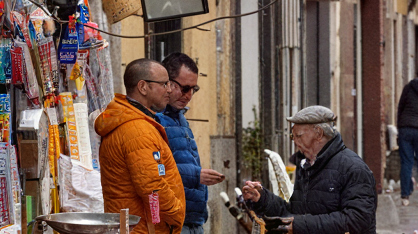}
                \caption*{SUPIR (D1)}
            \end{subfigure}
            \begin{subfigure}{0.32\textwidth}
                \includegraphics[width=\linewidth]{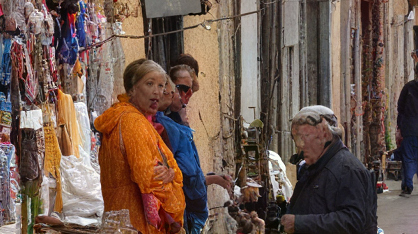}
                \caption*{SUPIR (D2)}
            \end{subfigure}
            \begin{subfigure}{0.32\textwidth}
                \includegraphics[width=\linewidth]{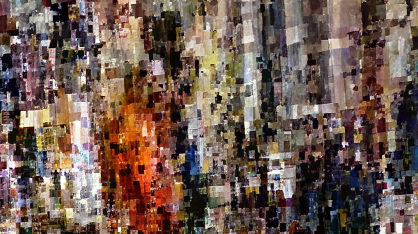}
                \caption*{SUPIR (D3)}
            \end{subfigure}
            \\
            \vspace{1em} 
            \begin{subfigure}{0.32\textwidth}
                \includegraphics[width=\linewidth]{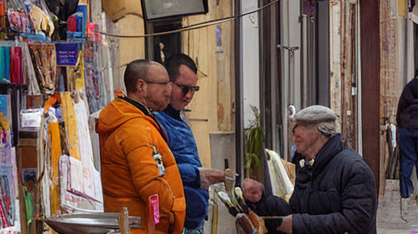}
                \caption*{FaithDiff (D1)}
            \end{subfigure}
            \begin{subfigure}{0.32\textwidth}
                \includegraphics[width=\linewidth]{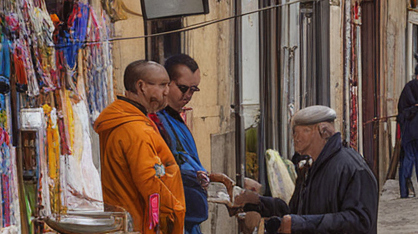}
                \caption*{FaithDiff (D2)}
            \end{subfigure}
            \begin{subfigure}{0.32\textwidth}
                \includegraphics[width=\linewidth]{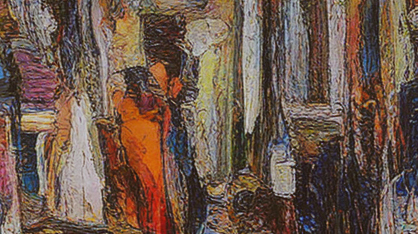}
                \caption*{FaithDiff (D3)}
            \end{subfigure}
            \\
            \vspace{1em} 
            \begin{subfigure}{0.32\textwidth}
                \includegraphics[width=\linewidth]{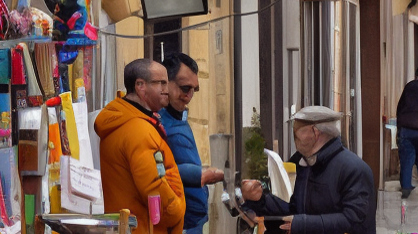}
                \caption*{SeeSR (D1)}
            \end{subfigure}
            \begin{subfigure}{0.32\textwidth}
                \includegraphics[width=\linewidth]{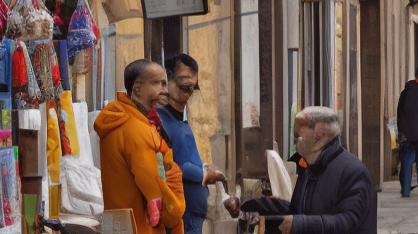}
                \caption*{SeeSR (D2)}
            \end{subfigure}
            \begin{subfigure}{0.32\textwidth}
                \includegraphics[width=\linewidth]{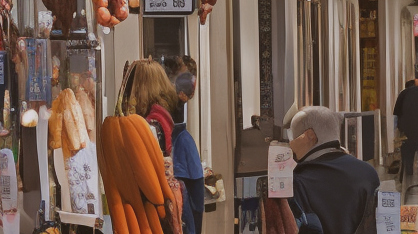}
                \caption*{SeeSR (D3)}
            \end{subfigure} 
            \\
            \vspace{1em} 
            \begin{subfigure}{0.32\textwidth}
                \includegraphics[width=\linewidth]{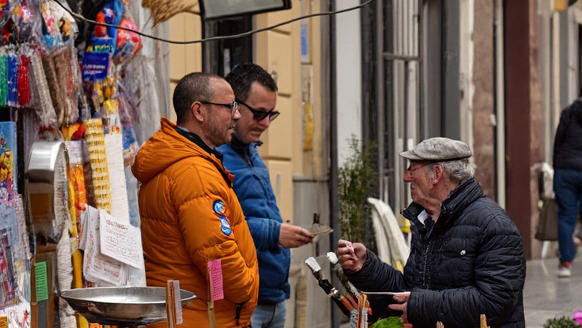}
                \caption*{\textbf{Ours} (D1)}
            \end{subfigure}
            \begin{subfigure}{0.32\textwidth}
                \includegraphics[width=\linewidth]{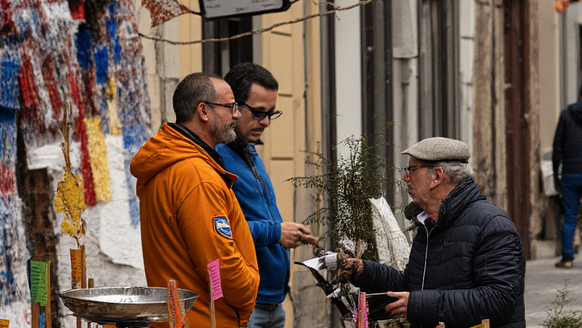}
                \caption*{\textbf{Ours} (D2)}
            \end{subfigure}
            \begin{subfigure}{0.32\textwidth}
                \includegraphics[width=\linewidth]{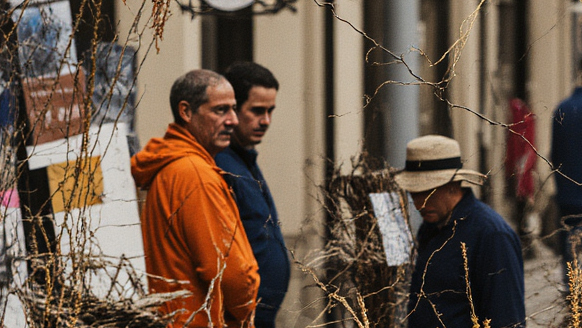}
                \caption*{\textbf{Ours} (D3)}
            \end{subfigure} 
        \end{subfigure}
    \end{subfigure}
    \caption{\textbf{Qualitative comparisons of different degradation level on LSDIR-Val.}
Selected examples across D1--D3 illustrate changes in structure, texture, and residual noise as degradation increases. Visual differences complement the per-level numerical results rather than establish a universal fidelity ranking.
}
    \label{app:Degradation_LSDIR}
\end{figure}

\begin{figure}[h]
    \centering

    \begin{subfigure}{\textwidth}
        \centering
        \begin{subfigure}{0.3\textwidth}
            \includegraphics[width=\linewidth]{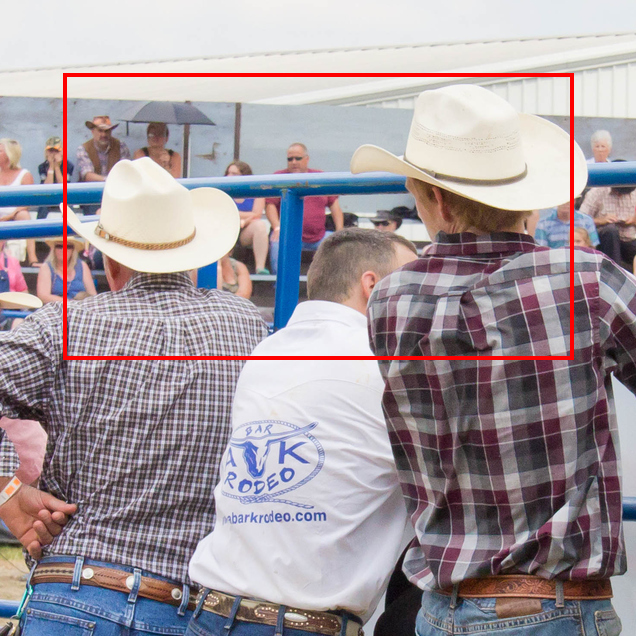}
            \caption*{GT}
        \end{subfigure}
        \begin{subfigure}{0.69\textwidth}
            \centering
            \begin{subfigure}{0.32\textwidth}
                \includegraphics[width=\linewidth]{figures/synthetic_2/LQ.jpg}
                \caption*{LQ Input}
            \end{subfigure}
            \begin{subfigure}{0.32\textwidth}
                \includegraphics[width=\linewidth]{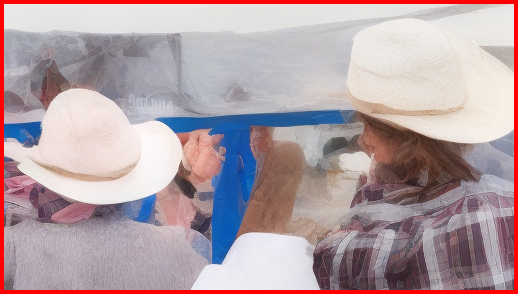}
                \caption*{DiffBIR}
            \end{subfigure}
            \begin{subfigure}{0.32\textwidth}
                \includegraphics[width=\linewidth]{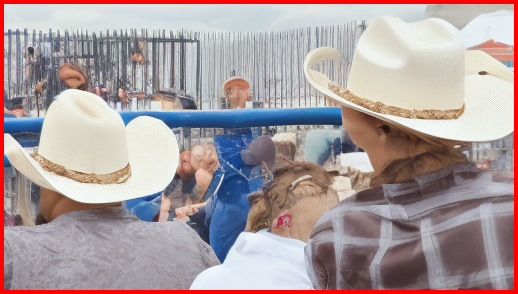}
                \caption*{SeeSR}
            \end{subfigure}
            \\
            \begin{subfigure}{0.32\textwidth}
                \includegraphics[width=\linewidth]{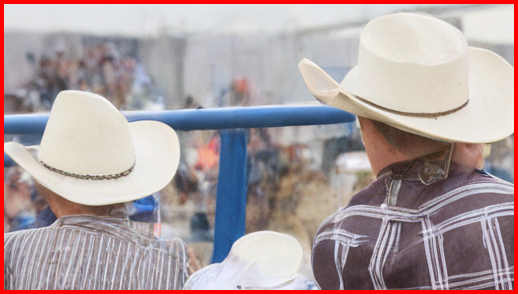}
                \caption*{FaithDiff}
            \end{subfigure}
            \begin{subfigure}{0.32\textwidth}
                \includegraphics[width=\linewidth]{figures/synthetic_2/SeeSR.jpg}
                \caption*{SeeSR}
            \end{subfigure}
            \begin{subfigure}{0.32\textwidth}
                \includegraphics[width=\linewidth]{figures/synthetic_2/Ours.jpg}
                \caption*{\textbf{Ours}}
            \end{subfigure}
        \end{subfigure}
    \end{subfigure}

    \begin{subfigure}{\textwidth}
        \centering
        \begin{subfigure}{0.3\textwidth}
            \includegraphics[width=\linewidth]{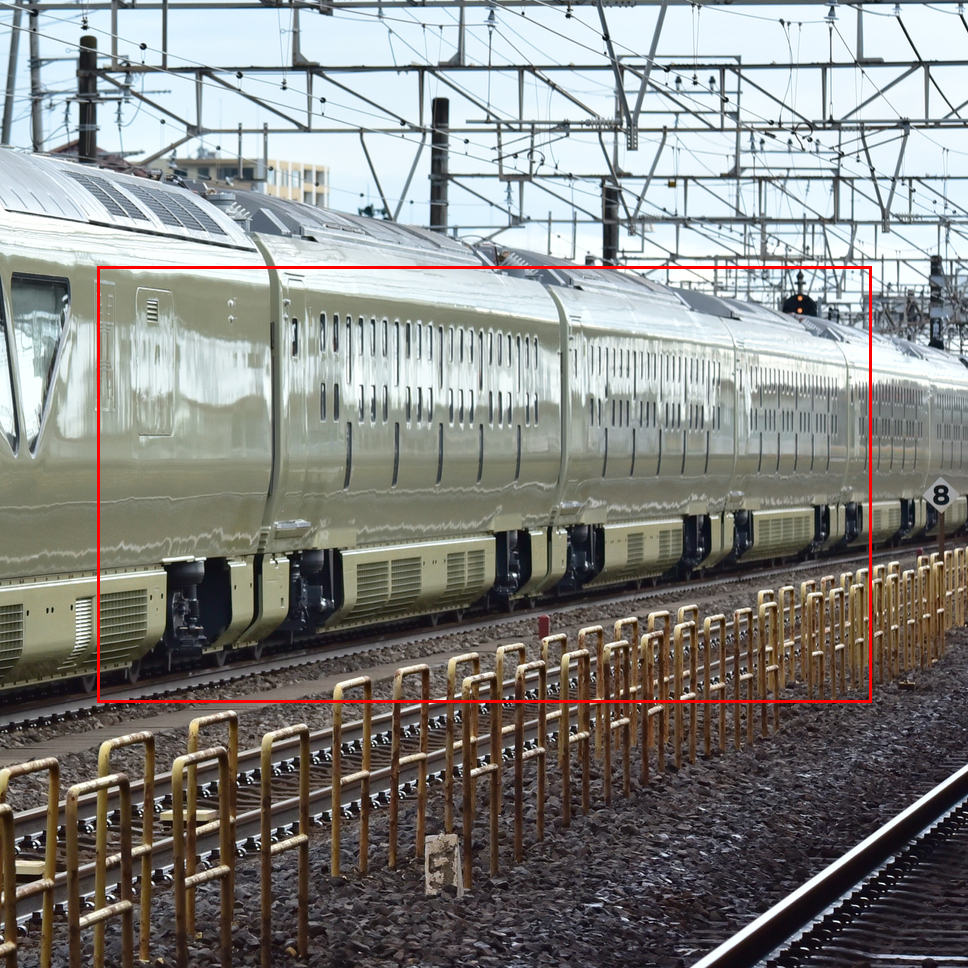}
            \caption*{GT}
        \end{subfigure}
        \begin{subfigure}{0.69\textwidth}
            \centering
            \begin{subfigure}{0.32\textwidth}
                \includegraphics[width=\linewidth]{figures/synthetic_3/LQ.jpg}
                \caption*{LQ Input}
            \end{subfigure}
            \begin{subfigure}{0.32\textwidth}
                \includegraphics[width=\linewidth]{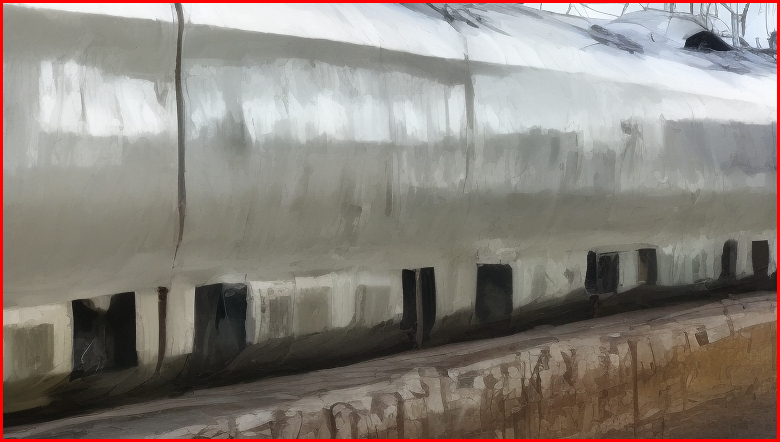}
                \caption*{DiffBIR}
            \end{subfigure}
            \begin{subfigure}{0.32\textwidth}
                \includegraphics[width=\linewidth]{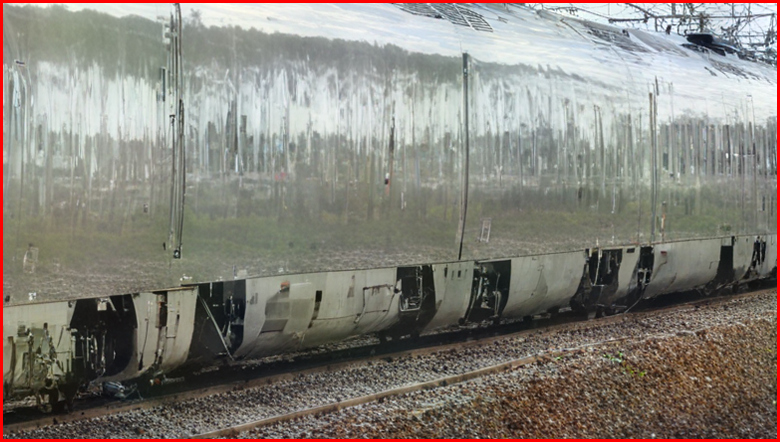}
                \caption*{SUPIR}
            \end{subfigure}
            \\
            \begin{subfigure}{0.32\textwidth}
                \includegraphics[width=\linewidth]{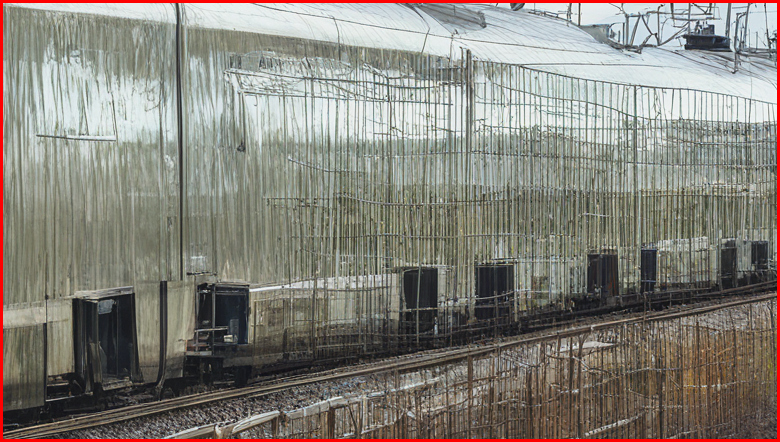}
                \caption*{FaithDiff}
            \end{subfigure}
            \begin{subfigure}{0.32\textwidth}
                \includegraphics[width=\linewidth]{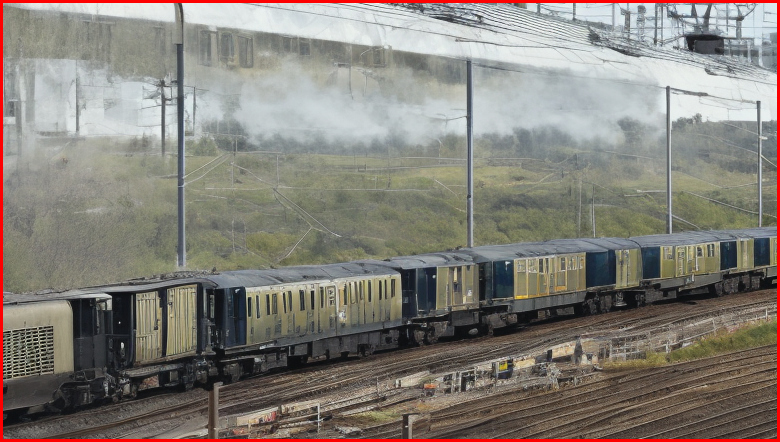}
                \caption*{SeeSR}
            \end{subfigure}
            \begin{subfigure}{0.32\textwidth}
                \includegraphics[width=\linewidth]{figures/synthetic_3/Ours.jpg}
                \caption*{\textbf{Ours}}
            \end{subfigure}
        \end{subfigure}
    \end{subfigure}

        \begin{subfigure}{\textwidth}
        \centering
        \begin{subfigure}{0.3\textwidth}
            \includegraphics[width=\linewidth]{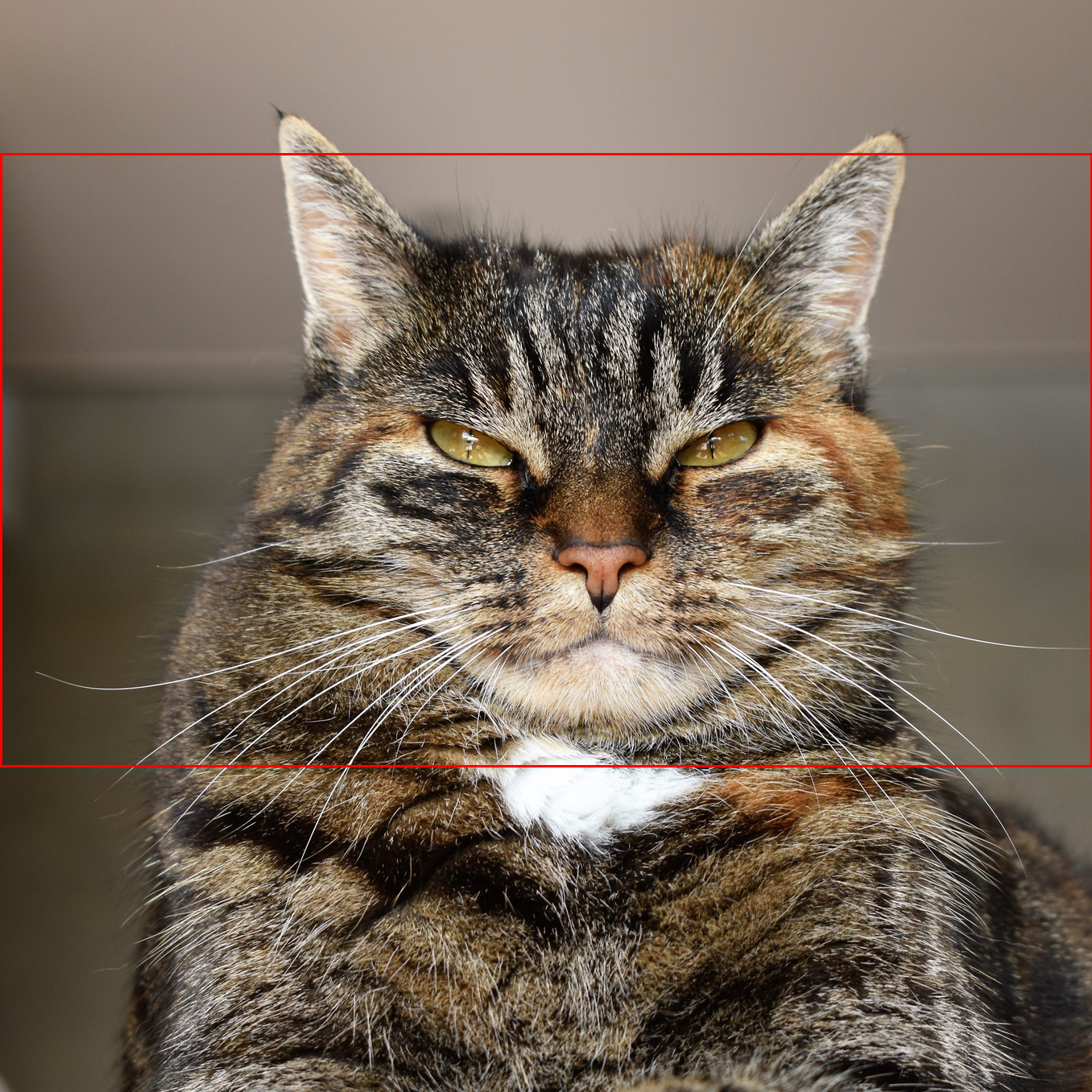}
            \caption*{GT}
        \end{subfigure}
        \begin{subfigure}{0.69\textwidth}
            \centering
            \begin{subfigure}{0.32\textwidth}
                \includegraphics[width=\linewidth]{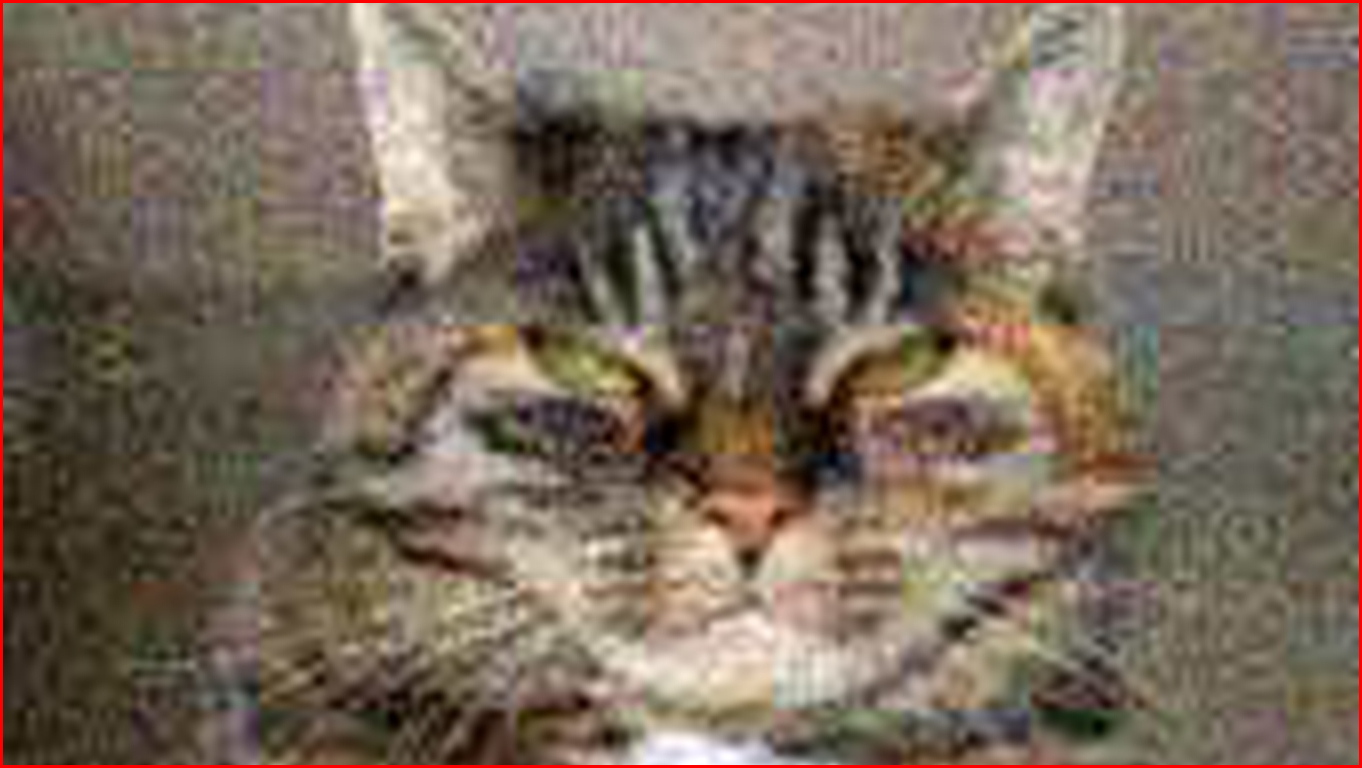}
                \caption*{LQ Input}
            \end{subfigure}
            \begin{subfigure}{0.32\textwidth}
                \includegraphics[width=\linewidth]{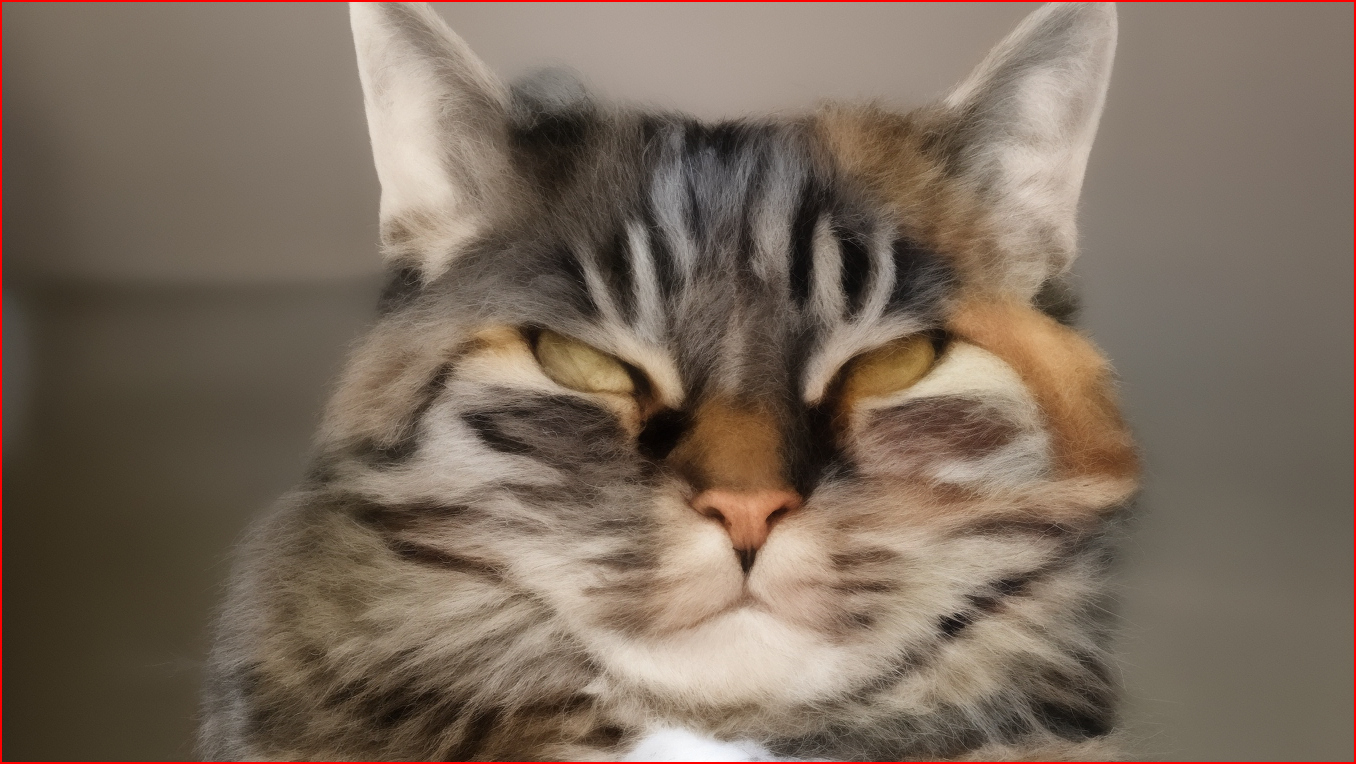}
                \caption*{DiffBIR}
            \end{subfigure}
            \begin{subfigure}{0.32\textwidth}
                \includegraphics[width=\linewidth]{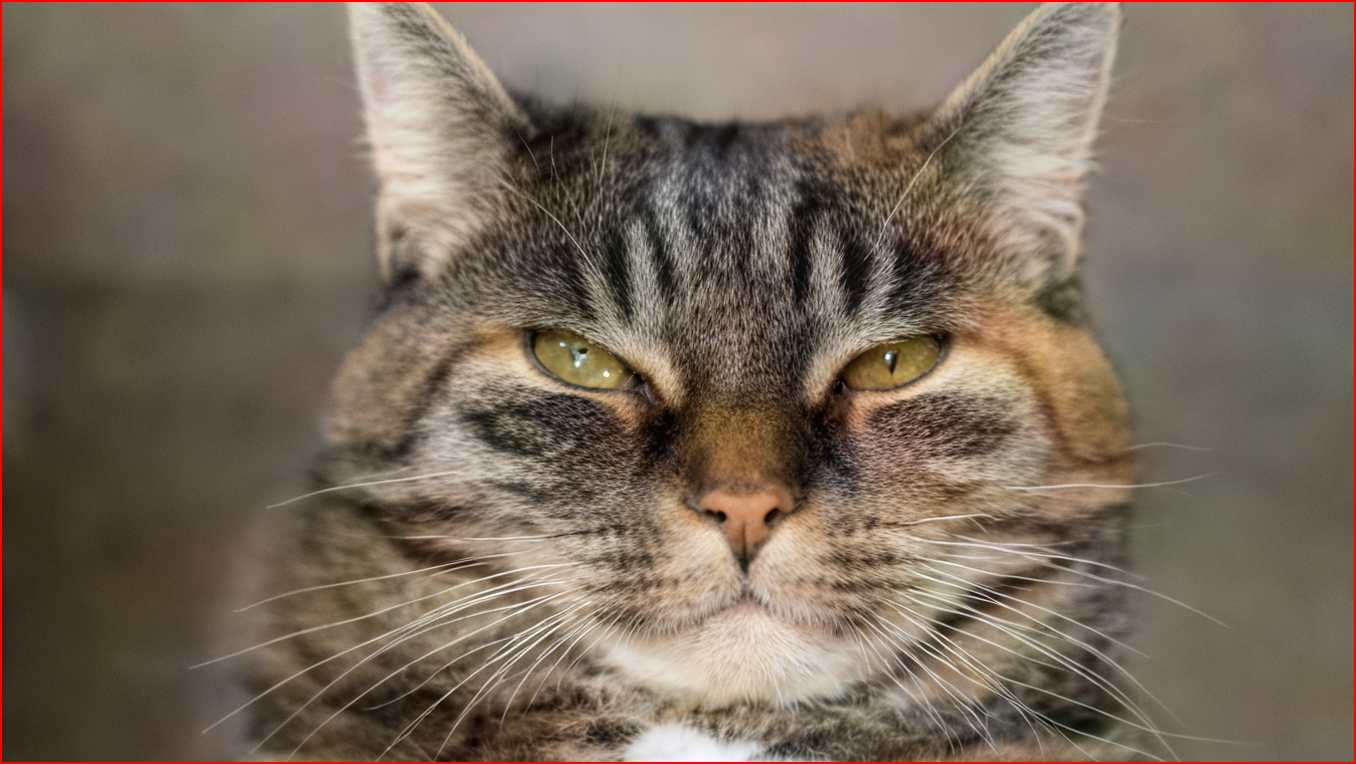}
                \caption*{SUPIR}
            \end{subfigure}
            \\
            \begin{subfigure}{0.32\textwidth}
                \includegraphics[width=\linewidth]{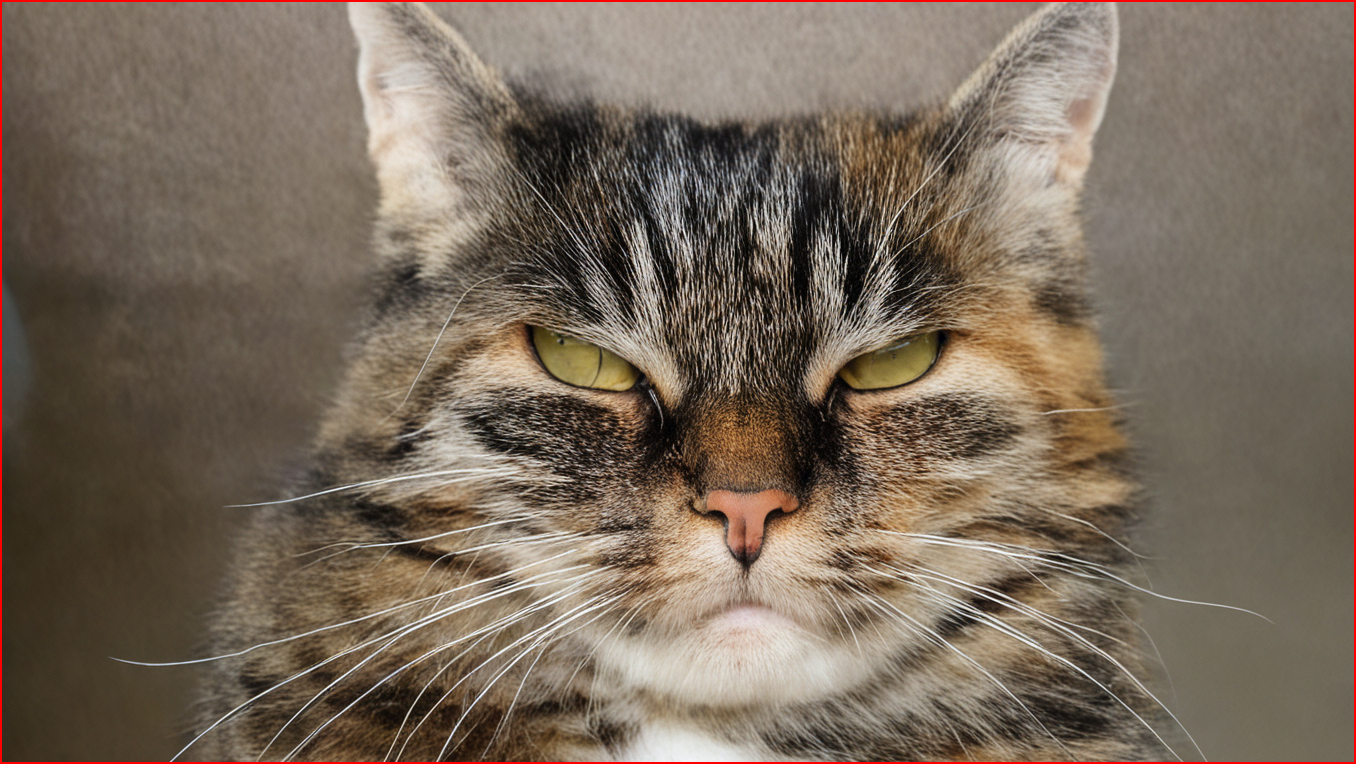}
                \caption*{FaithDiff}
            \end{subfigure}
            \begin{subfigure}{0.32\textwidth}
                \includegraphics[width=\linewidth]{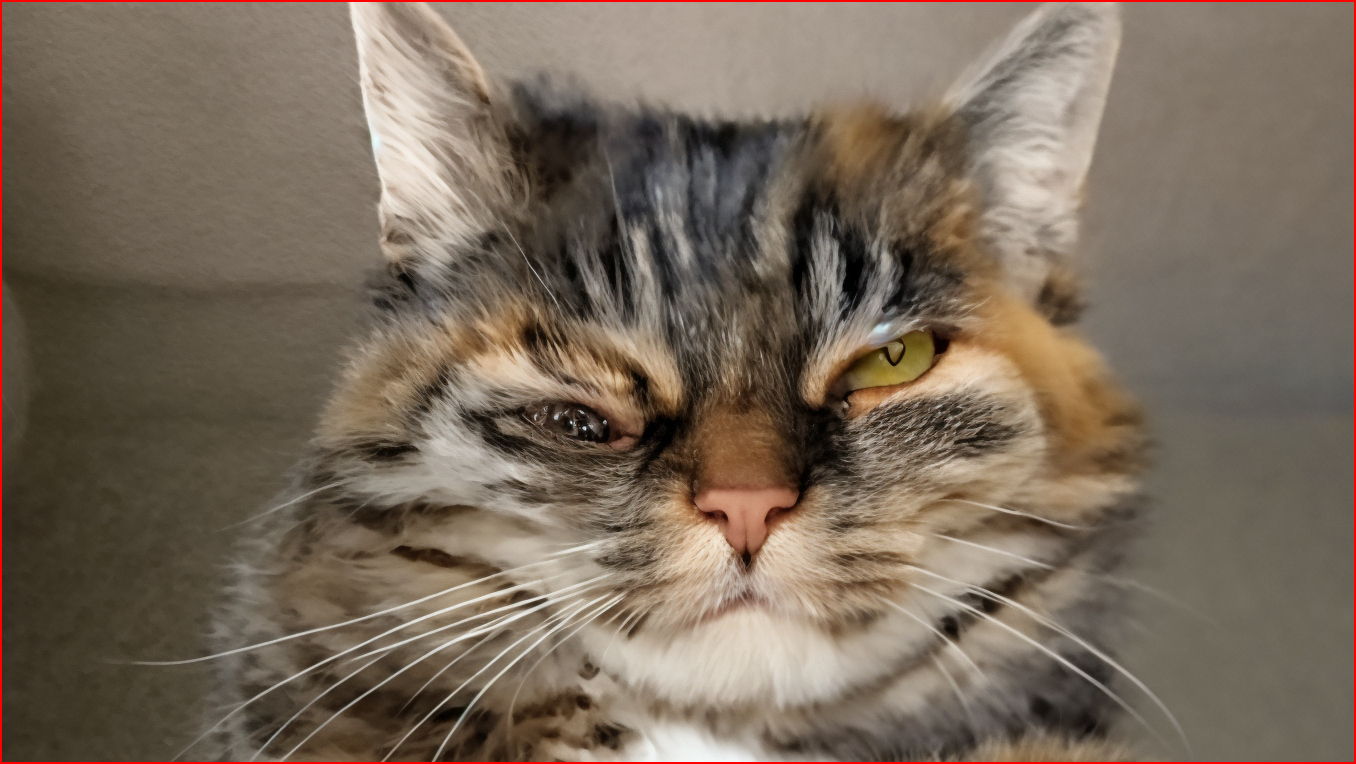}
                \caption*{SeeSR}
            \end{subfigure}
            \begin{subfigure}{0.32\textwidth}
                \includegraphics[width=\linewidth]{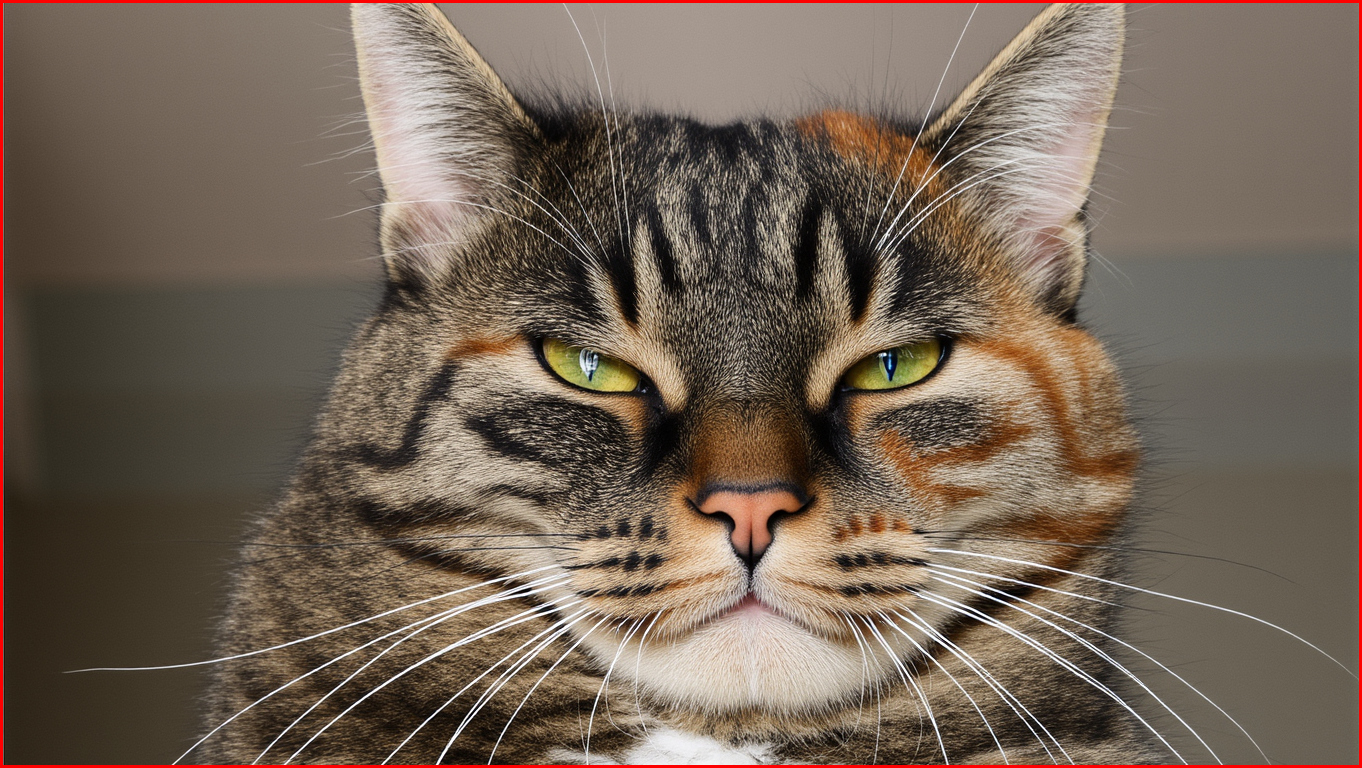}
                \caption*{\textbf{Ours}}
            \end{subfigure}
        \end{subfigure}
    \end{subfigure}

        \begin{subfigure}{\textwidth}
        \centering
        \begin{subfigure}{0.3\textwidth}
            \includegraphics[width=\linewidth]{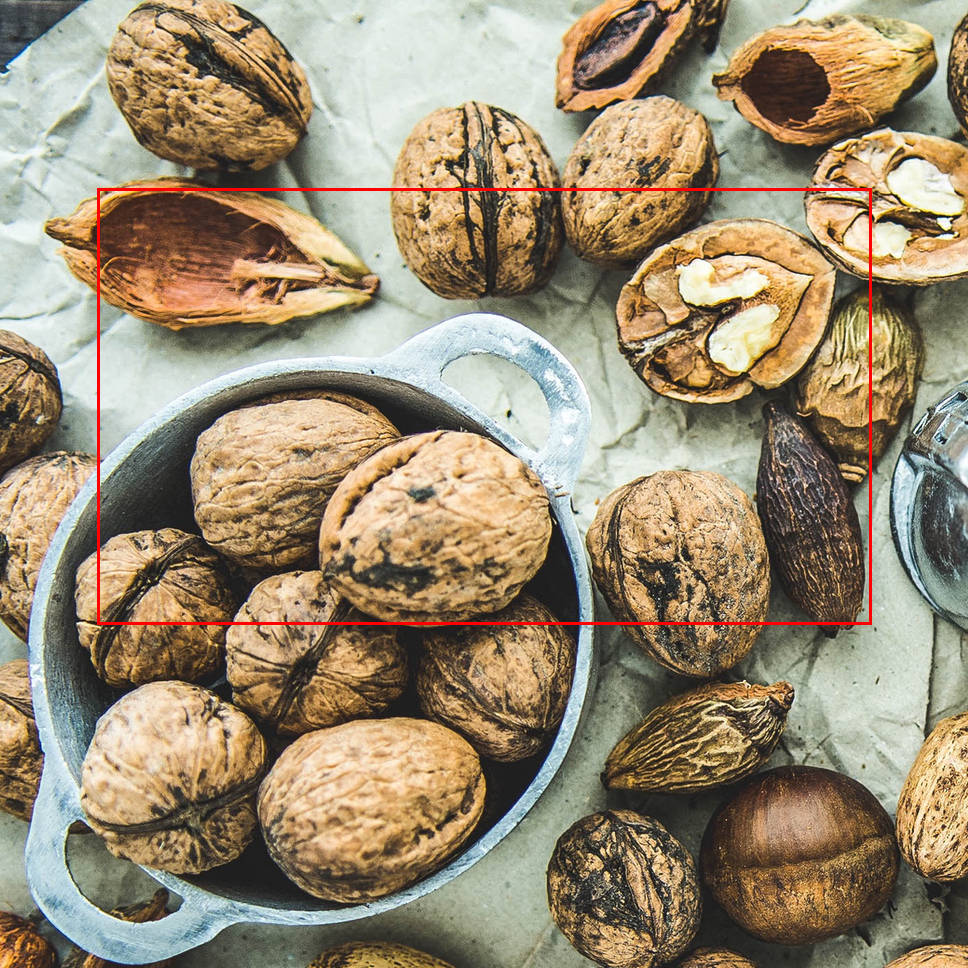}
            \caption*{GT}
        \end{subfigure}
        \begin{subfigure}{0.69\textwidth}
            \centering
            \begin{subfigure}{0.32\textwidth}
                \includegraphics[width=\linewidth]{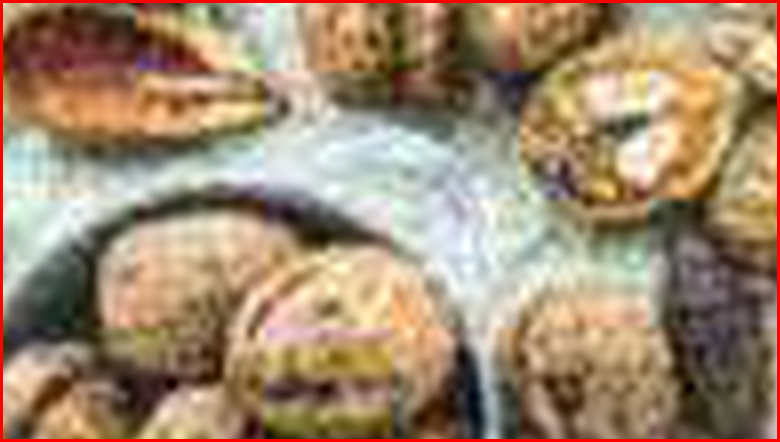}
                \caption*{LQ Input}
            \end{subfigure}
            \begin{subfigure}{0.32\textwidth}
                \includegraphics[width=\linewidth]{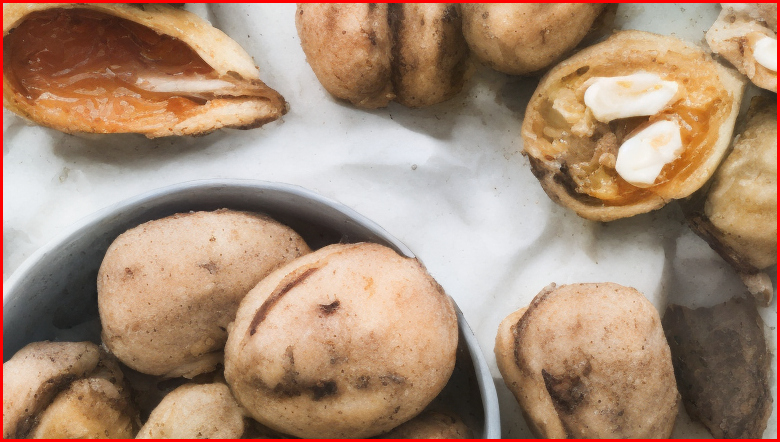}
                \caption*{DiffBIR}
            \end{subfigure}
            \begin{subfigure}{0.32\textwidth}
                \includegraphics[width=\linewidth]{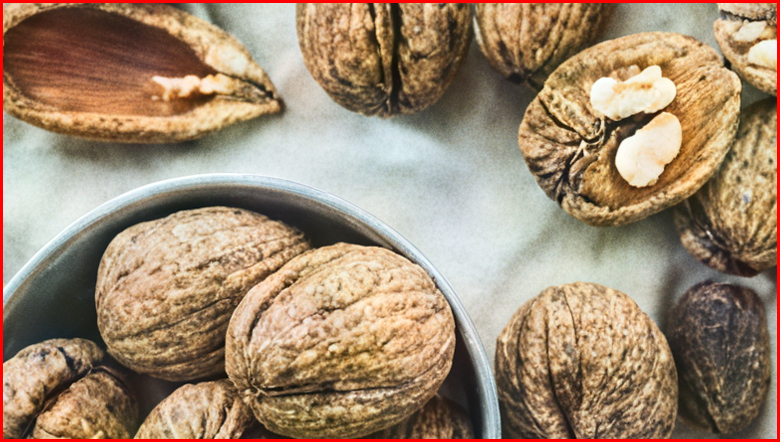}
                \caption*{SUPIR}
            \end{subfigure}
            \\
            \begin{subfigure}{0.32\textwidth}
                \includegraphics[width=\linewidth]{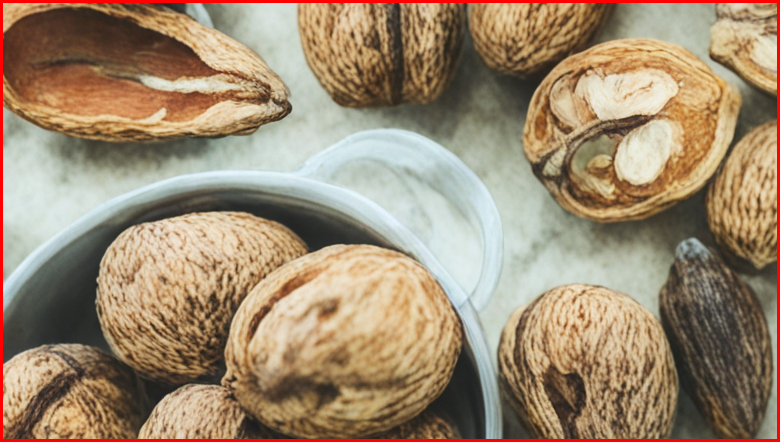}
                \caption*{FaithDiff}
            \end{subfigure}
            \begin{subfigure}{0.32\textwidth}
                \includegraphics[width=\linewidth]{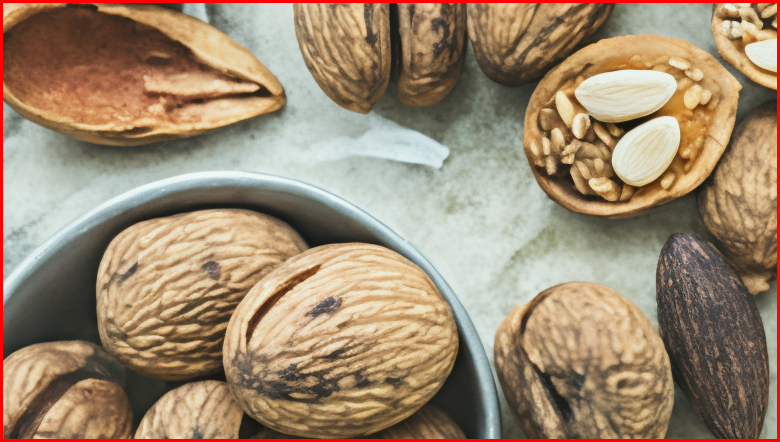}
                \caption*{SeeSR}
            \end{subfigure}
            \begin{subfigure}{0.32\textwidth}
                \includegraphics[width=\linewidth]{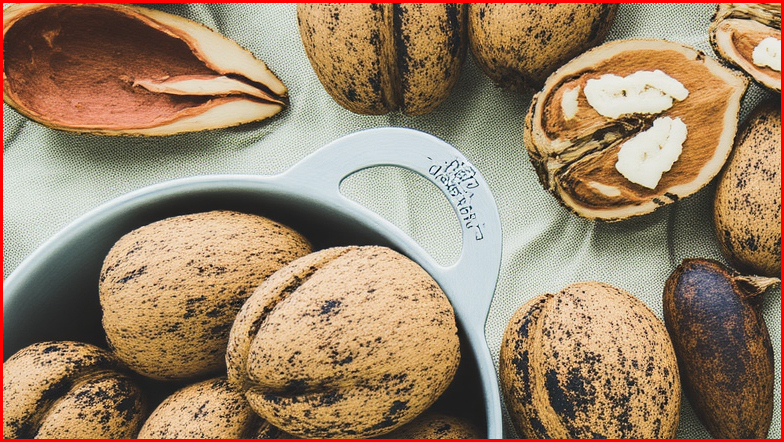}
                \caption*{\textbf{Ours}}
            \end{subfigure}
        \end{subfigure}
    \end{subfigure}

    \caption{\textbf{Qualitative comparisons on DIV2K-Val (D3).}
Selected examples illustrate structural and texture differences under severe degradation. The first example retains the duplicated SeeSR crop from the original comparison; no SUPIR crop is shown for that example. All original comparison image content is preserved.
}
    \label{app:Synthetic}
\end{figure}

\begin{figure}[h]
    \centering



    \begin{subfigure}{\textwidth}
        \centering
        \begin{subfigure}{0.3\textwidth}
            \includegraphics[width=\linewidth]{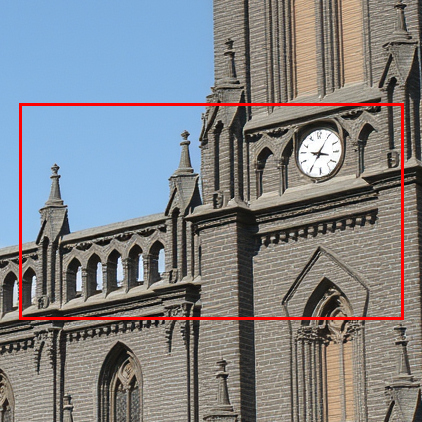}
            \caption*{\textbf{Ours}}
        \end{subfigure}
        \begin{subfigure}{0.69\textwidth}
            \centering
            \begin{subfigure}{0.32\textwidth}
                \includegraphics[width=\linewidth]{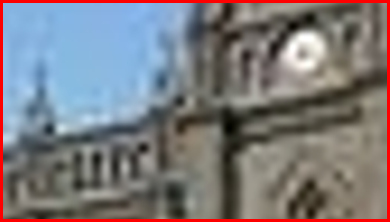}
                \caption*{LQ Input}
            \end{subfigure}
            \begin{subfigure}{0.32\textwidth}
                \includegraphics[width=\linewidth]{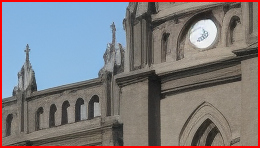}
                \caption*{DiffBIR}
            \end{subfigure}
            \begin{subfigure}{0.32\textwidth}
                \includegraphics[width=\linewidth]{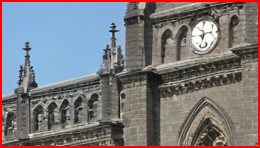}
                \caption*{SUPIR}
            \end{subfigure}
            \\
            \begin{subfigure}{0.32\textwidth}
                \includegraphics[width=\linewidth]{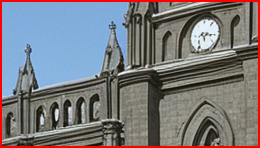}
                \caption*{FaithDiff}
            \end{subfigure}
            \begin{subfigure}{0.32\textwidth}
                \includegraphics[width=\linewidth]{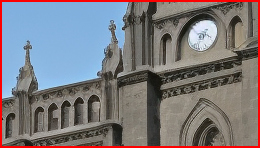}
                \caption*{SeeSR}
            \end{subfigure}
            \begin{subfigure}{0.32\textwidth}
                \includegraphics[width=\linewidth]{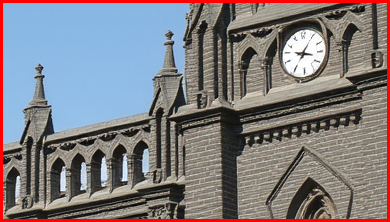}
                \caption*{\textbf{Ours}}
            \end{subfigure}
        \end{subfigure}
    \end{subfigure}

    \begin{subfigure}{\textwidth}
        \centering
        \begin{subfigure}{0.3\textwidth}
            \includegraphics[width=\linewidth]{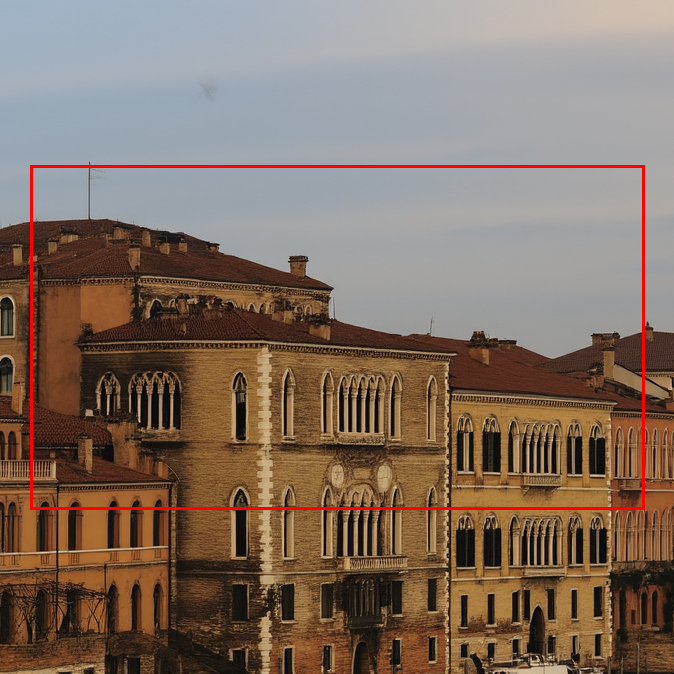}
            \caption*{\textbf{Ours}}
        \end{subfigure}
        \begin{subfigure}{0.69\textwidth}
            \centering
            \begin{subfigure}{0.32\textwidth}
                \includegraphics[width=\linewidth]{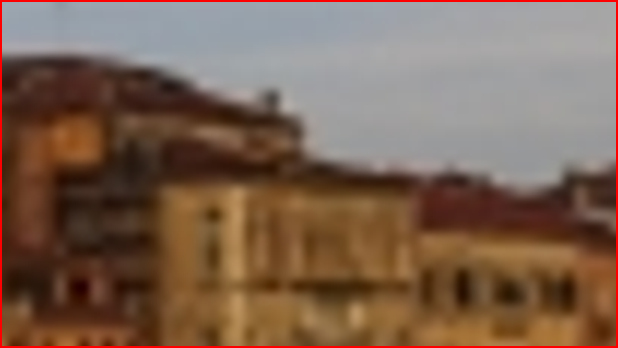}
                \caption*{LQ Input}
            \end{subfigure}
            \begin{subfigure}{0.32\textwidth}
                \includegraphics[width=\linewidth]{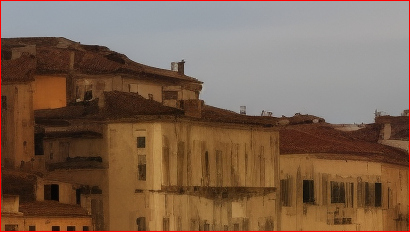}
                \caption*{DiffBIR}
            \end{subfigure}
            \begin{subfigure}{0.32\textwidth}
                \includegraphics[width=\linewidth]{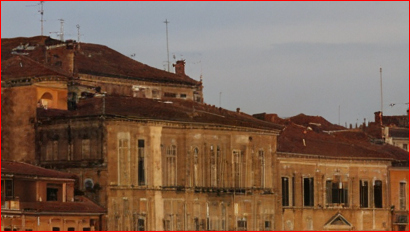}
                \caption*{SUPIR}
            \end{subfigure}
            \\
            \begin{subfigure}{0.32\textwidth}
                \includegraphics[width=\linewidth]{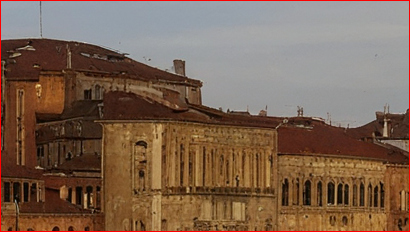}
                \caption*{FaithDiff}
            \end{subfigure}
            \begin{subfigure}{0.32\textwidth}
                \includegraphics[width=\linewidth]{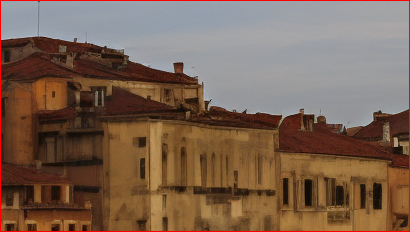}
                \caption*{SeeSR}
            \end{subfigure}
            \begin{subfigure}{0.32\textwidth}
                \includegraphics[width=\linewidth]{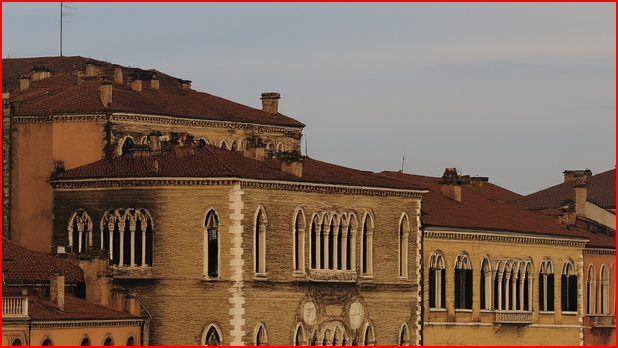}
                \caption*{\textbf{Ours}}
            \end{subfigure}
        \end{subfigure}
    \end{subfigure}

    \begin{subfigure}{\textwidth}
        \centering
        \begin{subfigure}{0.3\textwidth}
            \includegraphics[width=\linewidth]{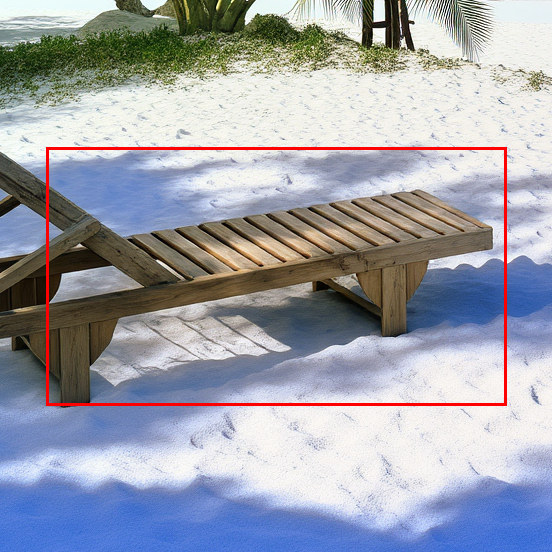}
            \caption*{\textbf{Ours}}
        \end{subfigure}
        \begin{subfigure}{0.69\textwidth}
            \centering
            \begin{subfigure}{0.32\textwidth}
                \includegraphics[width=\linewidth]{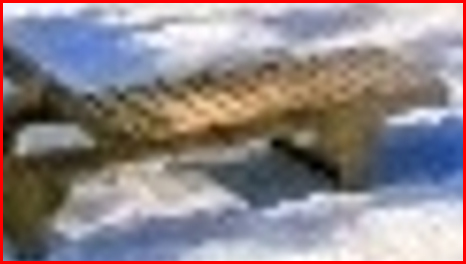}
                \caption*{LQ Input}
            \end{subfigure}
            \begin{subfigure}{0.32\textwidth}
                \includegraphics[width=\linewidth]{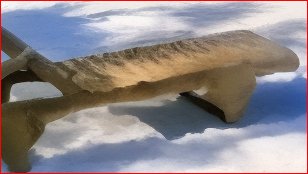}
                \caption*{DiffBIR}
            \end{subfigure}
            \begin{subfigure}{0.32\textwidth}
                \includegraphics[width=\linewidth]{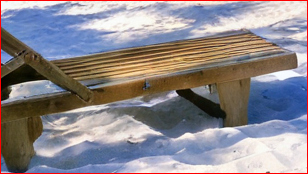}
                \caption*{SUPIR}
            \end{subfigure}
            \\
            \begin{subfigure}{0.32\textwidth}
                \includegraphics[width=\linewidth]{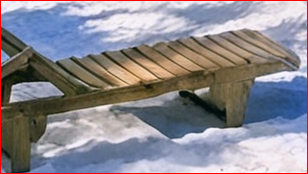}
                \caption*{FaithDiff}
            \end{subfigure}
            \begin{subfigure}{0.32\textwidth}
                \includegraphics[width=\linewidth]{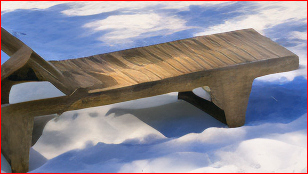}
                \caption*{SeeSR}
            \end{subfigure}
            \begin{subfigure}{0.32\textwidth}
                \includegraphics[width=\linewidth]{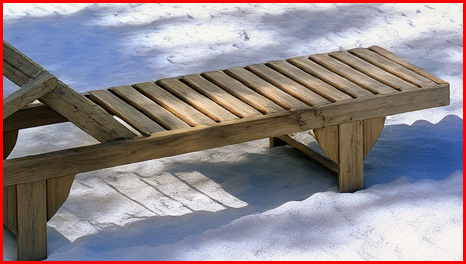}
                \caption*{\textbf{Ours}}
            \end{subfigure}
        \end{subfigure}
    \end{subfigure}

    \begin{subfigure}{\textwidth}
        \centering
        \begin{subfigure}{0.3\textwidth}
            \includegraphics[width=\linewidth]{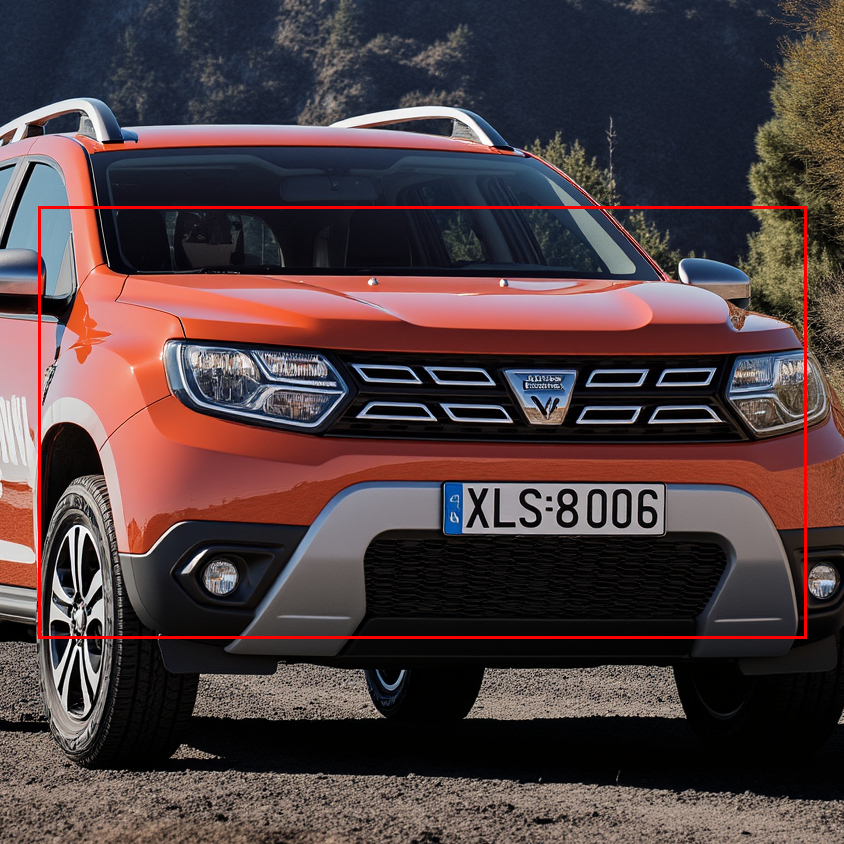}
            \caption*{\textbf{Ours}}
        \end{subfigure}
        \begin{subfigure}{0.69\textwidth}
            \centering
            \begin{subfigure}{0.32\textwidth}
                \includegraphics[width=\linewidth]{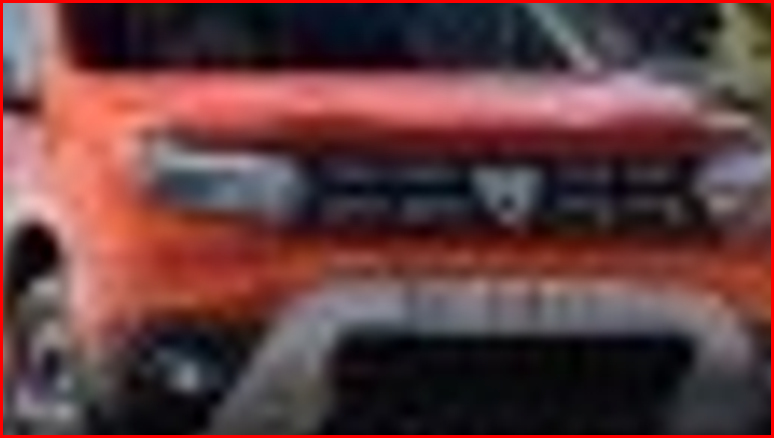}
                \caption*{LQ Input}
            \end{subfigure}
            \begin{subfigure}{0.32\textwidth}
                \includegraphics[width=\linewidth]{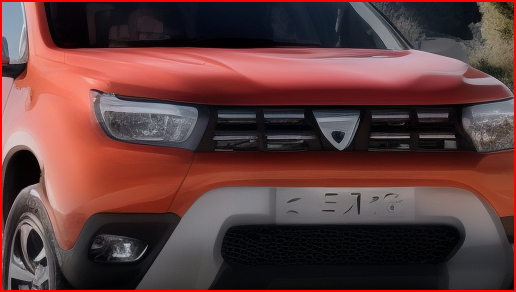}
                \caption*{DiffBIR}
            \end{subfigure}
            \begin{subfigure}{0.32\textwidth}
                \includegraphics[width=\linewidth]{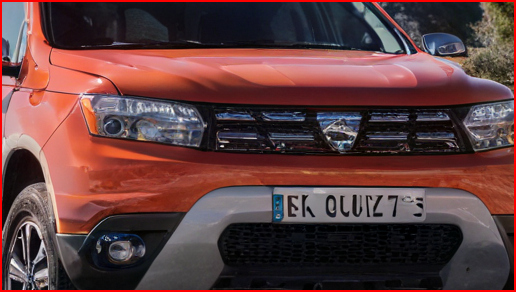}
                \caption*{SUPIR}
            \end{subfigure}
            \\
            \begin{subfigure}{0.32\textwidth}
                \includegraphics[width=\linewidth]{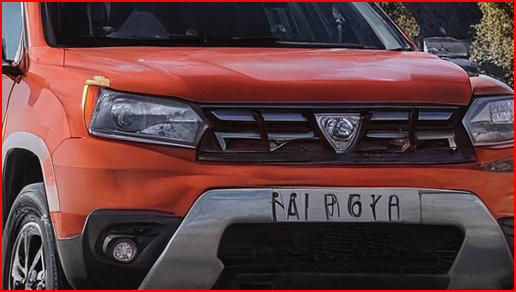}
                \caption*{FaithDiff}
            \end{subfigure}
            \begin{subfigure}{0.32\textwidth}
                \includegraphics[width=\linewidth]{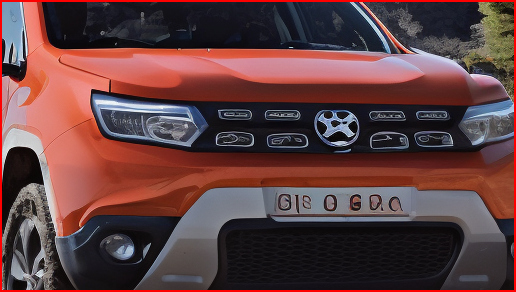}
                \caption*{SeeSR}
            \end{subfigure}
            \begin{subfigure}{0.32\textwidth}
                \includegraphics[width=\linewidth]{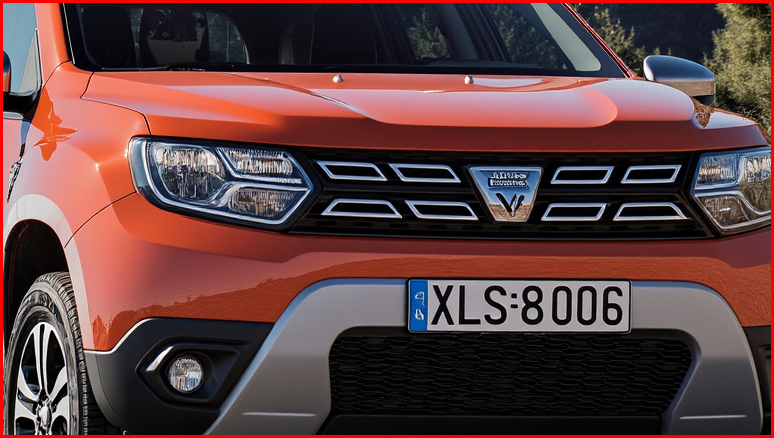}
                \caption*{\textbf{Ours}}
            \end{subfigure}
        \end{subfigure}
    \end{subfigure}

    \caption{\textbf{Qualitative comparisons on RealPhoto60 (2$\times$).}
    Left: our full-resolution result with crop locations marked (red boxes).
    Right: low-quality input and outputs from DiffBIR, SUPIR, FaithDiff, SeeSR, and ours on the corresponding crops.}
    \label{app:RealWorld}
\end{figure}

\begin{figure}[h]
    \centering
    \begin{subfigure}{0.99\textwidth}
        \centering
        \begin{subfigure}{0.32\textwidth}
            \includegraphics[width=\linewidth]{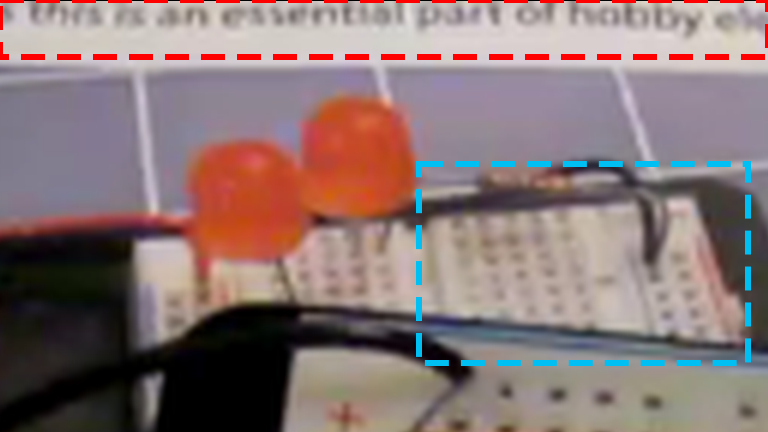}
            \caption*{LQ Input}
        \end{subfigure}
        \begin{subfigure}{0.32\textwidth}
            \includegraphics[width=\linewidth]{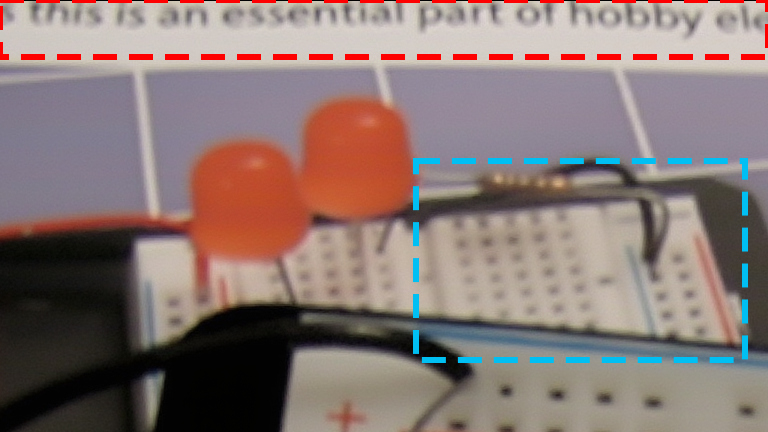}
            \caption*{Ours}
        \end{subfigure}
        \begin{subfigure}{0.32\textwidth}
            \includegraphics[width=\linewidth]{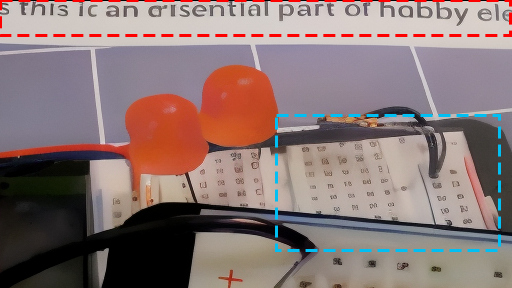}
            \caption*{DiffBIR}
        \end{subfigure}
        \\
        \begin{subfigure}{0.32\textwidth}
            \includegraphics[width=\linewidth]{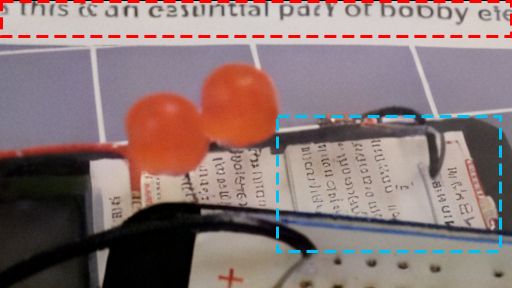}
            \caption*{SUPIR}
        \end{subfigure}
        \begin{subfigure}{0.32\textwidth}
            \includegraphics[width=\linewidth]{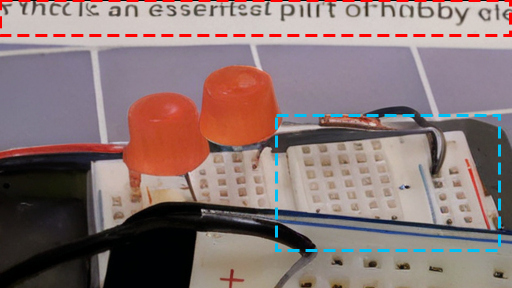}
            \caption*{FaithDiff}
        \end{subfigure}
        \begin{subfigure}{0.32\textwidth}
            \includegraphics[width=\linewidth]{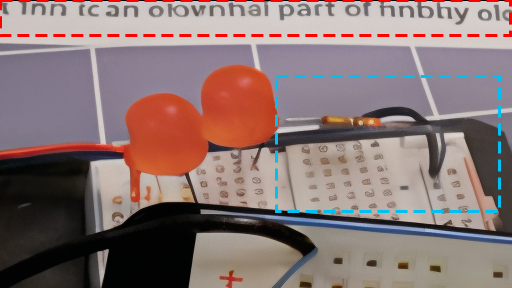}
            \caption*{SeeSR}
        \end{subfigure}
    \end{subfigure}
    \vspace{1em} 

    \begin{subfigure}{0.99\textwidth}
        \centering
        \begin{subfigure}{0.32\textwidth}
            \includegraphics[width=\linewidth]{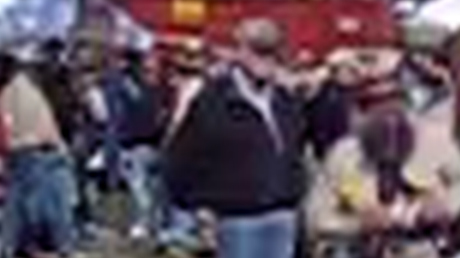}
            \caption*{LQ Input}
        \end{subfigure}
        \begin{subfigure}{0.32\textwidth}
            \includegraphics[width=\linewidth]{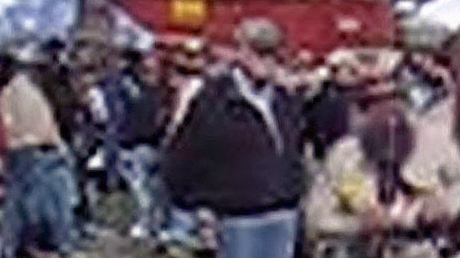}
            \caption*{Ours}
        \end{subfigure}
        \begin{subfigure}{0.32\textwidth}
            \includegraphics[width=\linewidth]{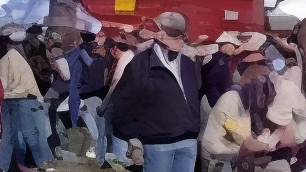}
            \caption*{DiffBIR}
        \end{subfigure}
        \\
        \begin{subfigure}{0.32\textwidth}
            \includegraphics[width=\linewidth]{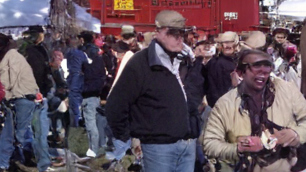}
            \caption*{SUPIR}
        \end{subfigure}
        \begin{subfigure}{0.32\textwidth}
            \includegraphics[width=\linewidth]{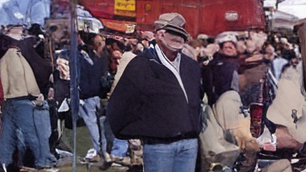}
            \caption*{FaithDiff}
        \end{subfigure}
        \begin{subfigure}{0.32\textwidth}
            \includegraphics[width=\linewidth]{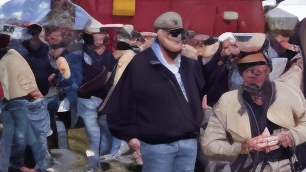}
            \caption*{SeeSR}
        \end{subfigure}
    \end{subfigure}
    \vspace{1em} 

    \begin{subfigure}{0.99\textwidth}
        \centering
        \begin{subfigure}{0.32\textwidth}
            \includegraphics[width=\linewidth]{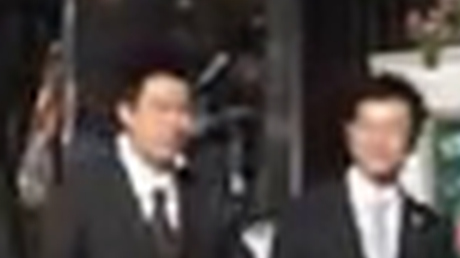}
            \caption*{LQ Input}
        \end{subfigure}
        \begin{subfigure}{0.32\textwidth}
            \includegraphics[width=\linewidth]{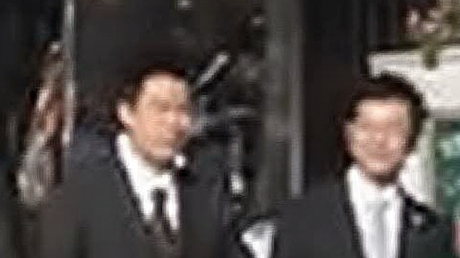}
            \caption*{Ours}
        \end{subfigure}
        \begin{subfigure}{0.32\textwidth}
            \includegraphics[width=\linewidth]{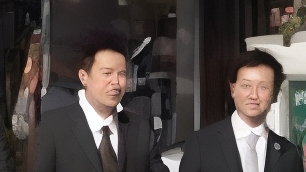}
            \caption*{DiffBIR}
        \end{subfigure}
        \\
        \begin{subfigure}{0.32\textwidth}
            \includegraphics[width=\linewidth]{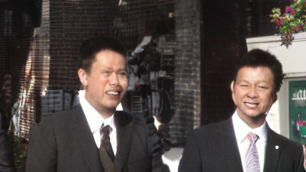}
            \caption*{SUPIR}
        \end{subfigure}
        \begin{subfigure}{0.32\textwidth}
            \includegraphics[width=\linewidth]{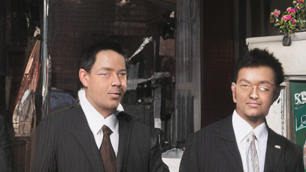}
            \caption*{FaithDiff}
        \end{subfigure}
        \begin{subfigure}{0.32\textwidth}
            \includegraphics[width=\linewidth]{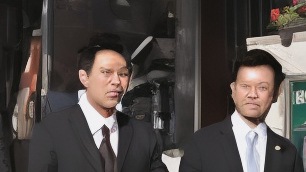}
            \caption*{SeeSR}
        \end{subfigure}
    \end{subfigure}

\caption{\textbf{Failure cases on RealLQ250.} 
These examples show residual degradation and lost details, consistent with limited generalization beyond the synthetic training recipe. Closeness to the low-quality input or lower sharpness does not certify correct geometry, reliability, or absence of hallucination. All generative restorations require caution when interpreting recovered content, especially historical artworks.}
\label{app:RealHQ250Fail}
\end{figure}

\begin{table*}[h]
  \centering
  \caption{\textbf{Sensitivity to input preprocessing on RealLQ250.}
We applied Gaussian blur with a standard deviation equal to 1\% of the image width. All reported no-reference scores increase after preprocessing. Without clean targets or a controlled causal analysis, this does not disprove overfitting or establish greater reconstruction fidelity.
  }
  \vspace{2pt}
  \renewcommand{\arraystretch}{1.05}
  \begin{adjustbox}{max width=0.7\linewidth}
    \begin{tabular}{c
                    S[table-format=1.4]
                    S[table-format=2.4]
                    S[table-format=1.4]}
      \toprule
      RealLQ250 & {CLIPIQA$\uparrow$} & {MUSIQ$\uparrow$} & {MANIQA$\uparrow$}\\
      \midrule
      Origin & 0.4792 & 54.3427 & 0.5397 \\
      Blurry & \textbf{0.5041} & \textbf{60.5381} & \textbf{0.5605} \\
      \bottomrule
    \end{tabular}
  \end{adjustbox}
\end{table*}

\begin{figure}[h]
    \centering

    \begin{subfigure}{0.99\textwidth}
        \centering
        \begin{subfigure}{0.32\textwidth}
            \includegraphics[width=\linewidth]{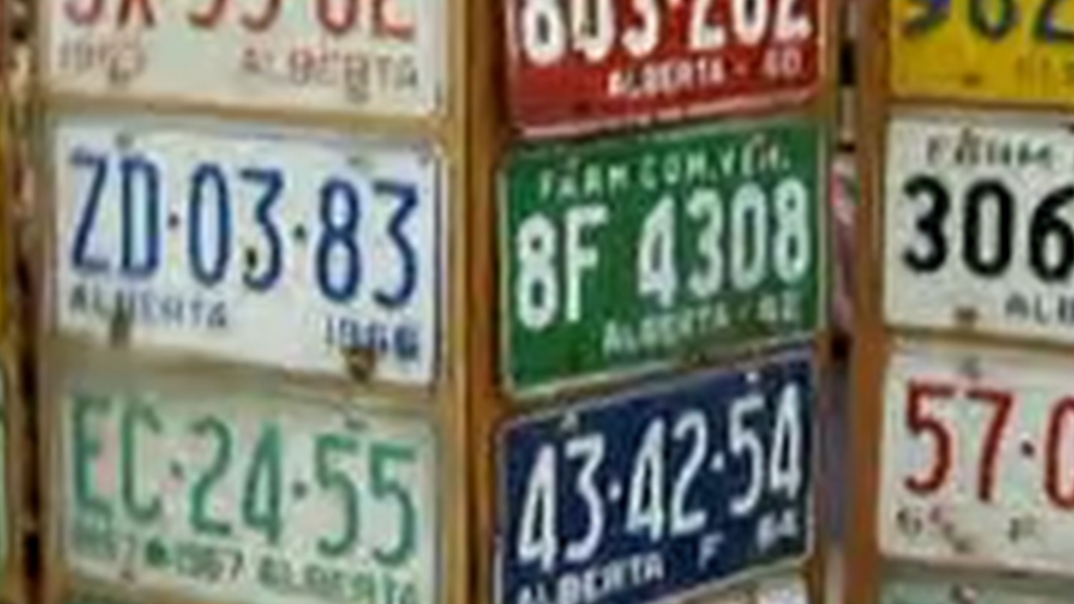}
            \caption*{Origin Input}
        \end{subfigure}
        \begin{subfigure}{0.32\textwidth}
            \includegraphics[width=\linewidth]{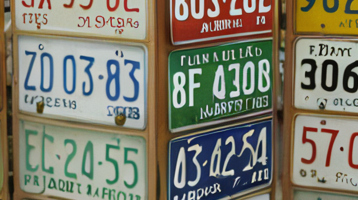}
            \caption*{FaithDiff}
        \end{subfigure}
        \begin{subfigure}{0.32\textwidth}
            \includegraphics[width=\linewidth]{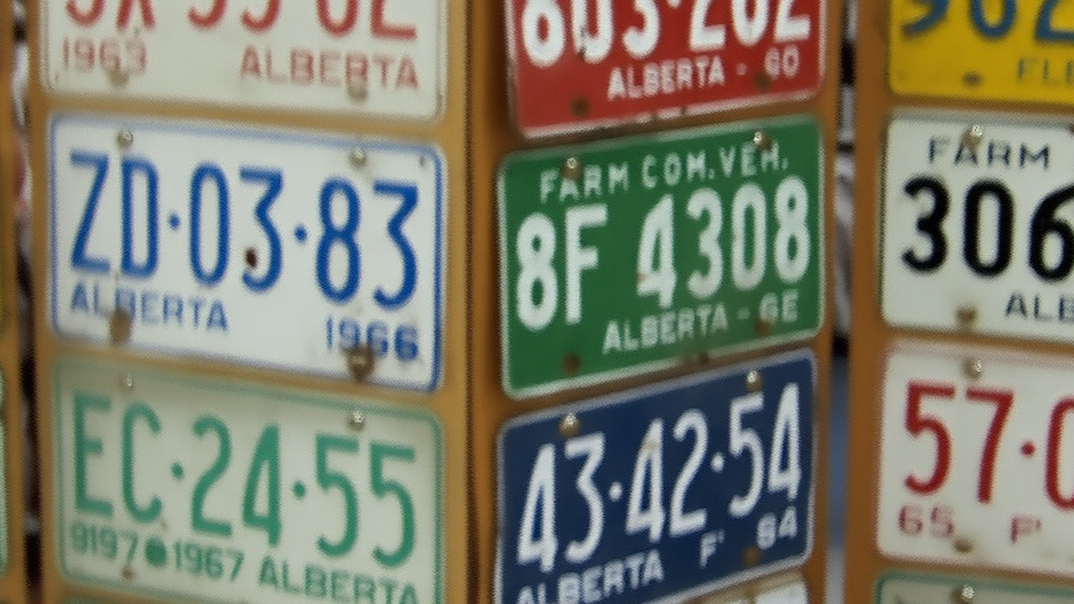}
            \caption*{Ours}
        \end{subfigure}
        \\
        \begin{subfigure}{0.32\textwidth}
            \includegraphics[width=\linewidth]{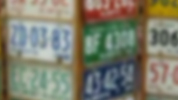}
            \caption*{Blurry Input}
        \end{subfigure}
        \begin{subfigure}{0.32\textwidth}
            \includegraphics[width=\linewidth]{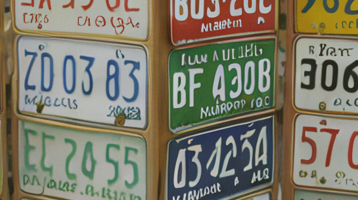}
            \caption*{FaithDiff}
        \end{subfigure}
        \begin{subfigure}{0.32\textwidth}
            \includegraphics[width=\linewidth]{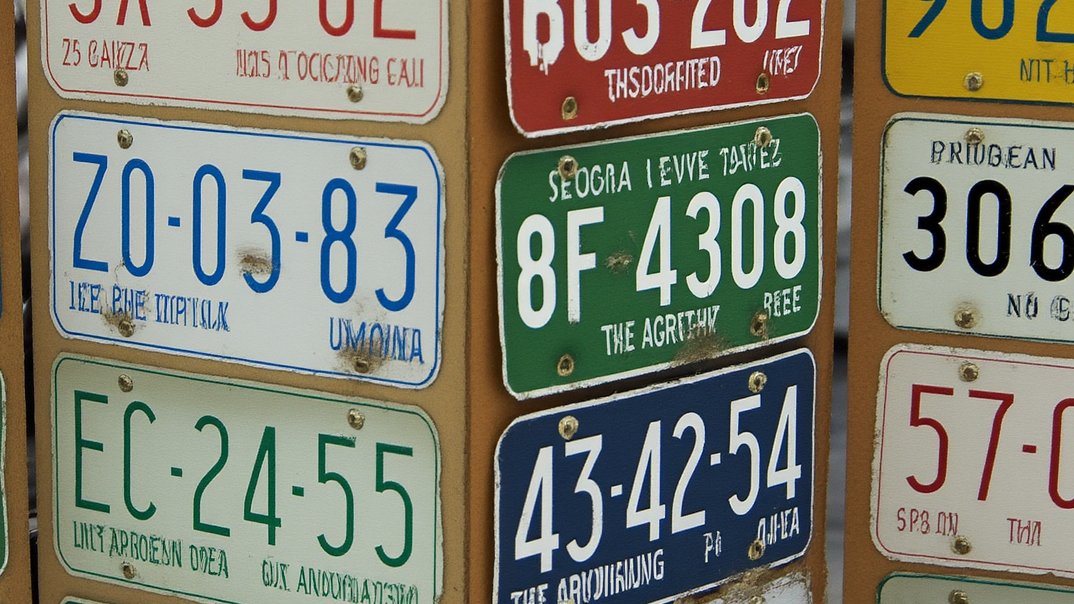}
            \caption*{Ours}
        \end{subfigure}
    \end{subfigure}
    \vspace{1em} 

    \begin{subfigure}{0.99\textwidth}
        \centering
        \begin{subfigure}{0.32\textwidth}
            \includegraphics[width=\linewidth]{figures/RealHQ250Fail/03/LQ.jpg}
            \caption*{Origin Input}
        \end{subfigure}
        \begin{subfigure}{0.32\textwidth}
            \includegraphics[width=\linewidth]{figures/RealHQ250Fail/03/FaithDiff.jpg}
            \caption*{FaithDiff}
        \end{subfigure}
        \begin{subfigure}{0.32\textwidth}
            \includegraphics[width=\linewidth]{figures/RealHQ250Fail/03/Ours.jpg}
            \caption*{Ours}
        \end{subfigure}
        \\
        \begin{subfigure}{0.32\textwidth}
            \includegraphics[width=\linewidth]{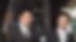}
            \caption*{Blurry Input}
        \end{subfigure}
        \begin{subfigure}{0.32\textwidth}
            \includegraphics[width=\linewidth]{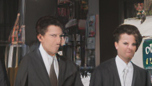}
            \caption*{FaithDiff}
        \end{subfigure}
        \begin{subfigure}{0.32\textwidth}
            \includegraphics[width=\linewidth]{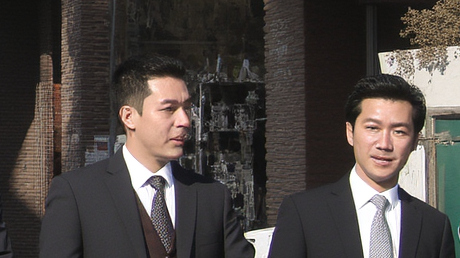}
            \caption*{Ours}
        \end{subfigure}
    \end{subfigure}

\caption{\textbf{Sensitivity to input preprocessing on RealLQ250.}
Gaussian blur with a standard deviation equal to 1\% of the image width changes the outputs in these selected examples. The corresponding no-reference scores improve, but original-detail fidelity cannot be established from these unpaired real-world images.
}
\end{figure}

\begin{figure}[h]
    \centering
    \begin{subfigure}{\textwidth}
        \centering
        \begin{subfigure}{0.3\textwidth}
            \includegraphics[width=\linewidth]{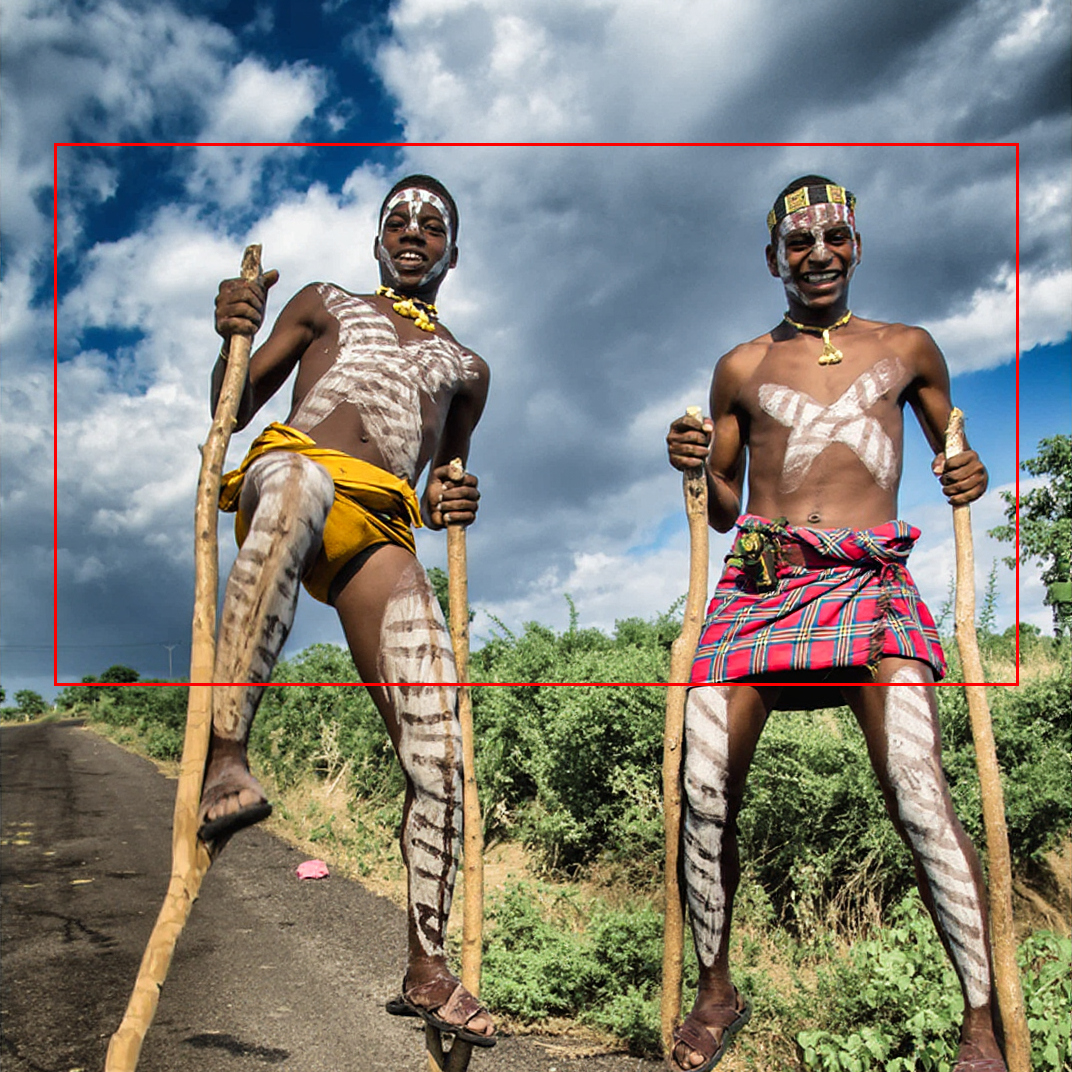}
            \caption*{\textbf{Ours}}
        \end{subfigure}
        \begin{subfigure}{0.69\textwidth}
            \centering
            \begin{subfigure}{0.32\textwidth}
                \includegraphics[width=\linewidth]{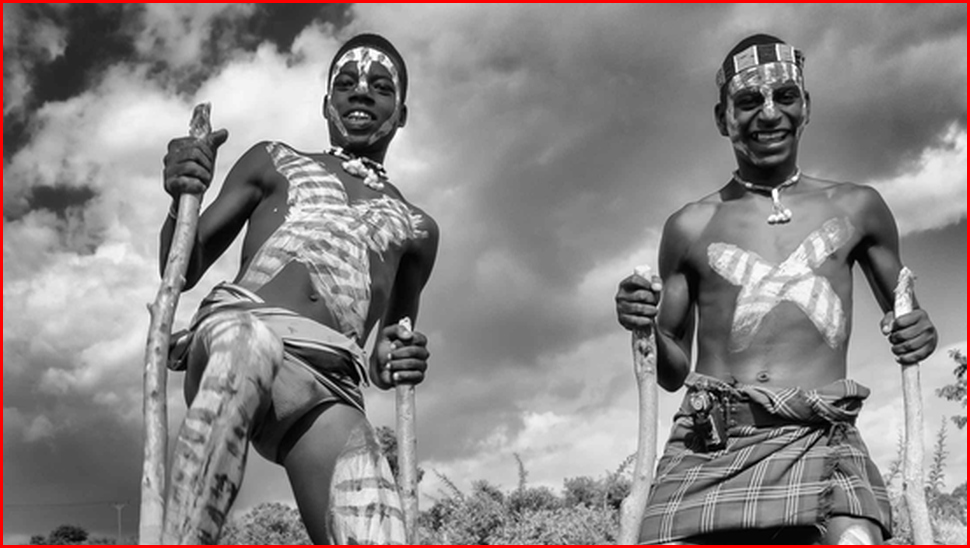}
                \caption*{LQ Input}
            \end{subfigure}
            \begin{subfigure}{0.32\textwidth}
                \includegraphics[width=\linewidth]{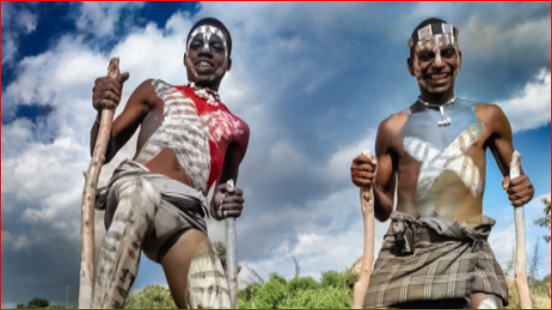}
                \caption*{InstColor}
            \end{subfigure}
            \begin{subfigure}{0.32\textwidth}
                \includegraphics[width=\linewidth]{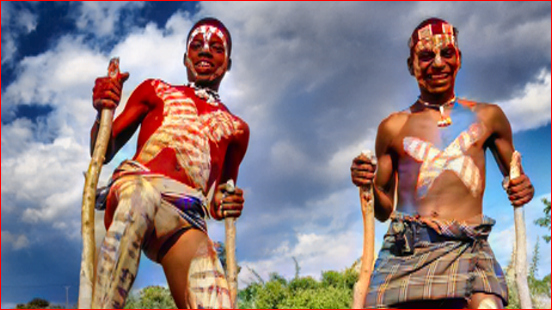}
                \caption*{ColorFormer}
            \end{subfigure}
            \\
            \begin{subfigure}{0.32\textwidth}
                \includegraphics[width=\linewidth]{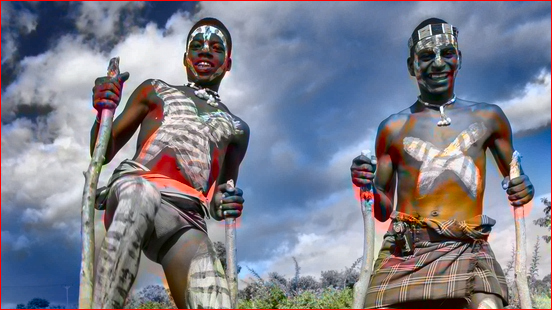}
                \caption*{BigColor}
            \end{subfigure}
            \begin{subfigure}{0.32\textwidth}
                \includegraphics[width=\linewidth]{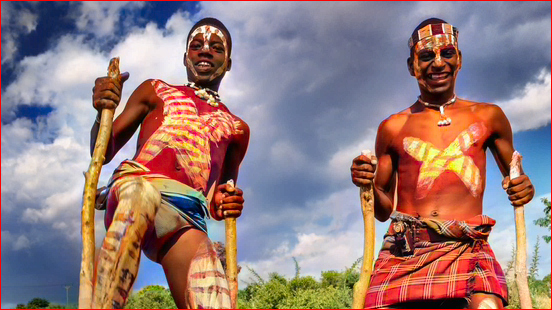}
                \caption*{DDColor}
            \end{subfigure}
            \begin{subfigure}{0.32\textwidth}
                \includegraphics[width=\linewidth]{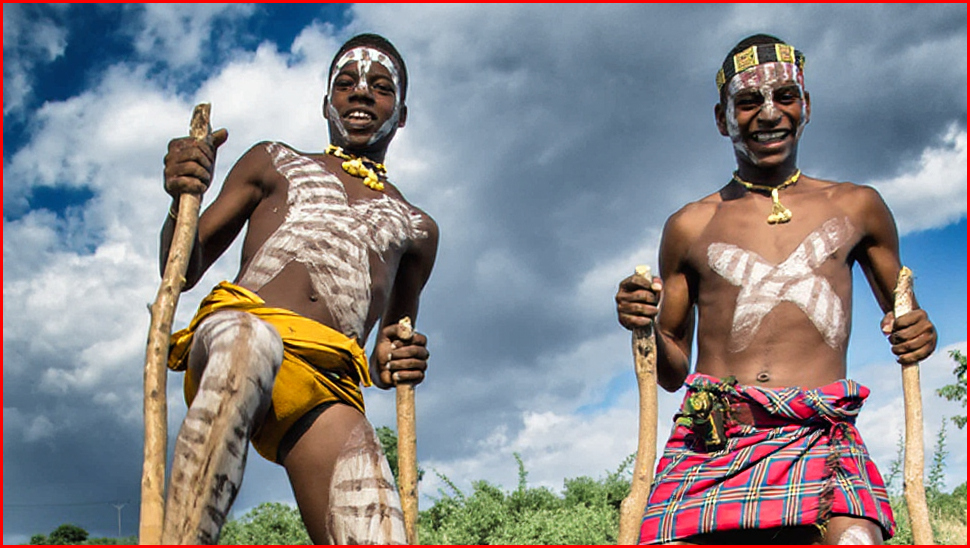}
                \caption*{\textbf{Ours}}
            \end{subfigure}
        \end{subfigure}
    \end{subfigure}

    \begin{subfigure}{\textwidth}
        \centering
        \begin{subfigure}{0.3\textwidth}
            \includegraphics[width=\linewidth]{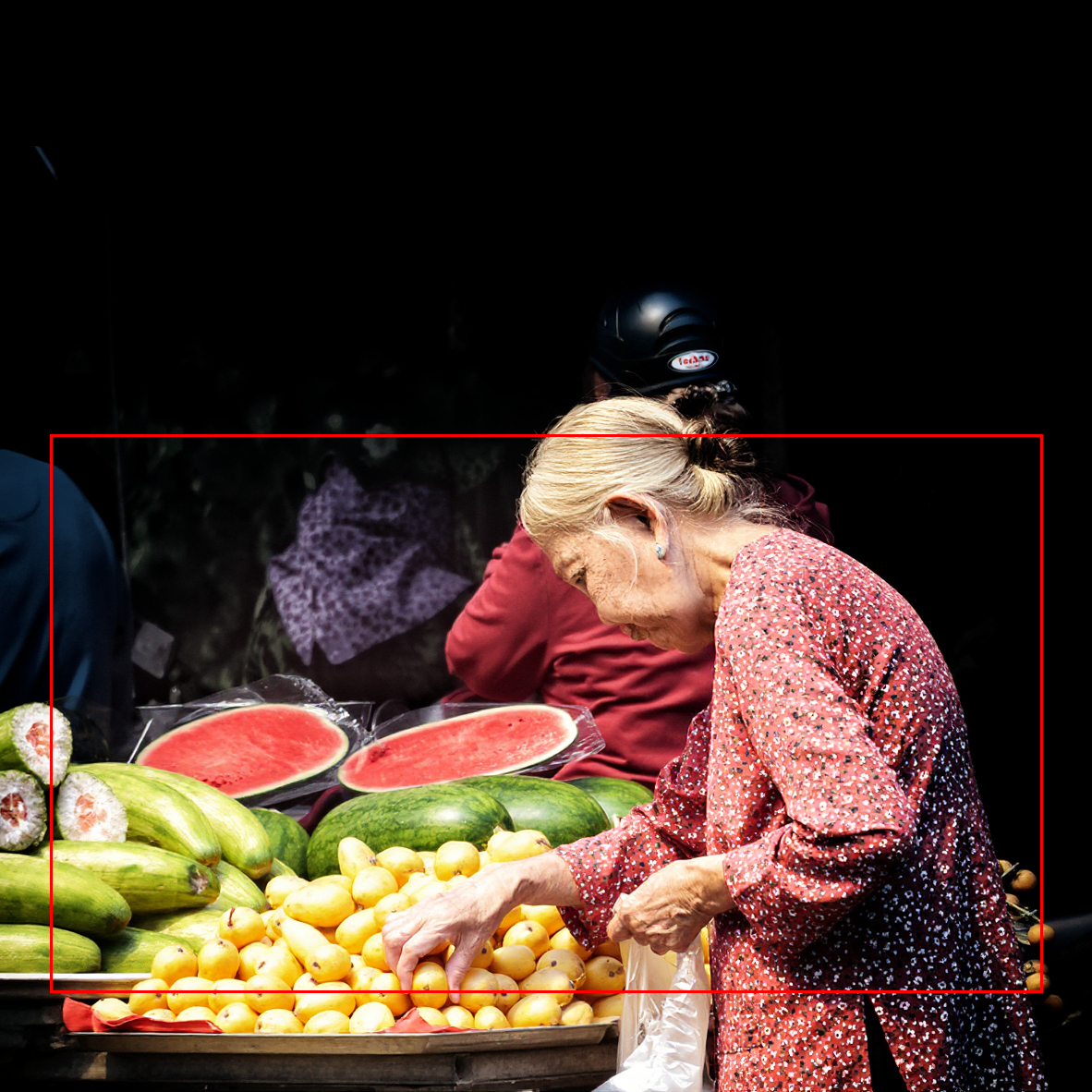}
            \caption*{\textbf{Ours}}
        \end{subfigure}
        \begin{subfigure}{0.69\textwidth}
            \centering
            \begin{subfigure}{0.32\textwidth}
                \includegraphics[width=\linewidth]{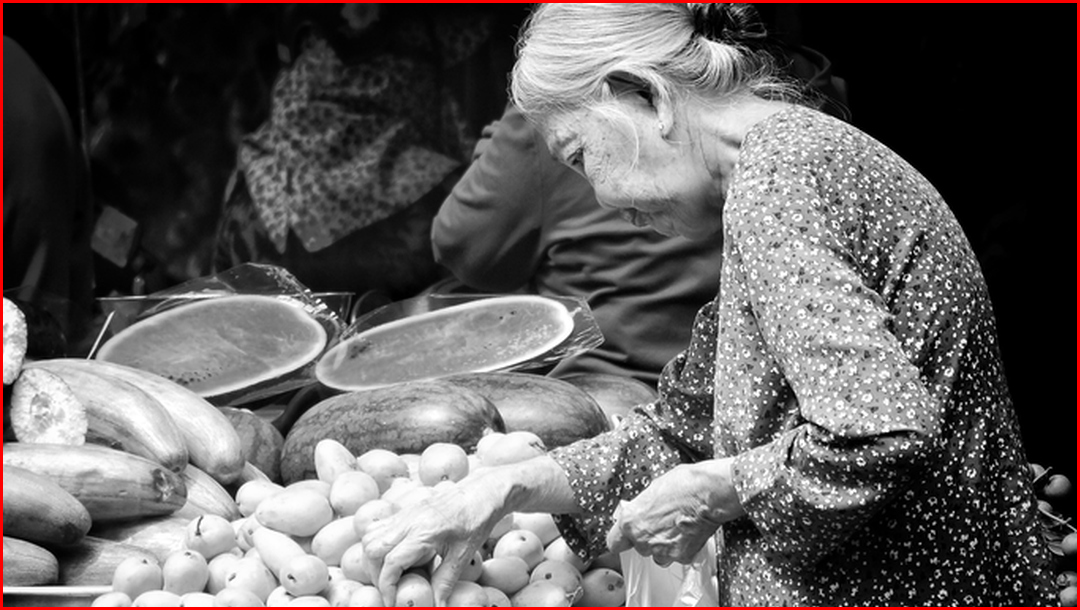}
                \caption*{LQ Input}
            \end{subfigure}
            \begin{subfigure}{0.32\textwidth}
                \includegraphics[width=\linewidth]{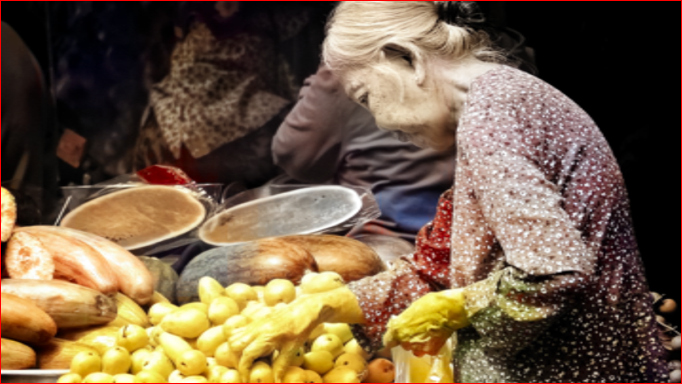}
                \caption*{InstColor}
            \end{subfigure}
            \begin{subfigure}{0.32\textwidth}
                \includegraphics[width=\linewidth]{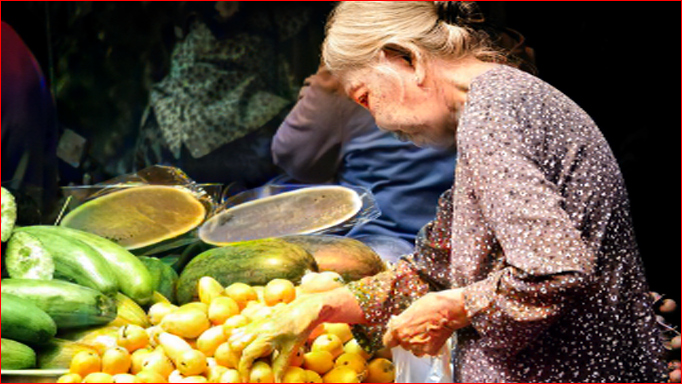}
                \caption*{ColorFormer}
            \end{subfigure}
            \\
            \begin{subfigure}{0.32\textwidth}
                \includegraphics[width=\linewidth]{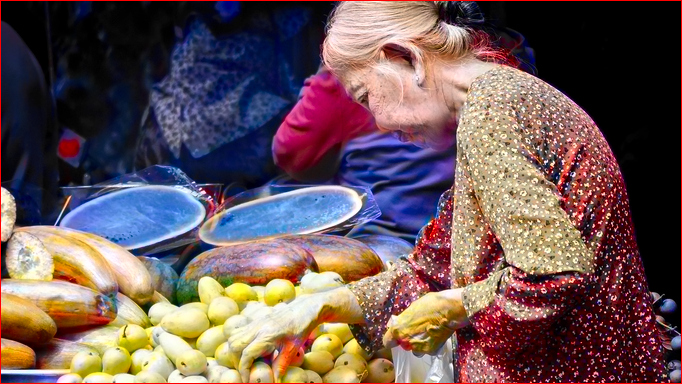}
                \caption*{BigColor}
            \end{subfigure}
            \begin{subfigure}{0.32\textwidth}
                \includegraphics[width=\linewidth]{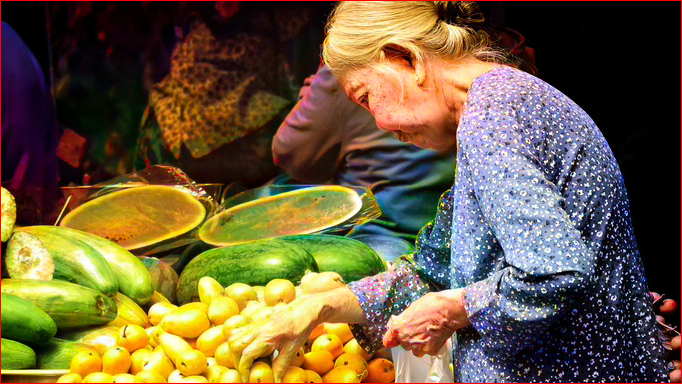}
                \caption*{DDColor}
            \end{subfigure}
            \begin{subfigure}{0.32\textwidth}
                \includegraphics[width=\linewidth]{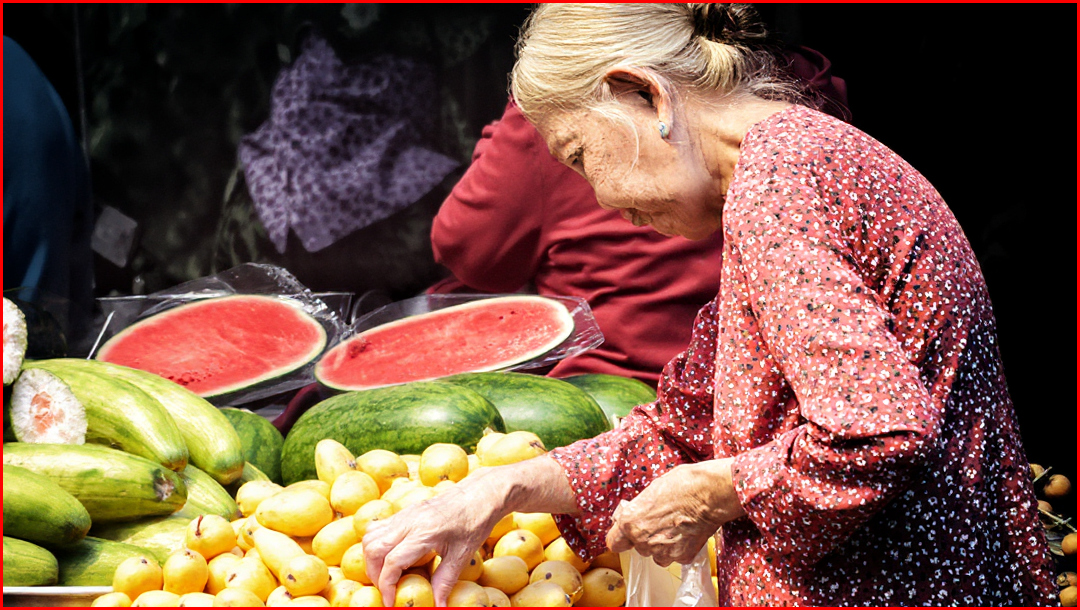}
                \caption*{\textbf{Ours}}
            \end{subfigure}
        \end{subfigure}
    \end{subfigure}

    \begin{subfigure}{\textwidth}
        \centering
        \begin{subfigure}{0.3\textwidth}
            \includegraphics[width=\linewidth]{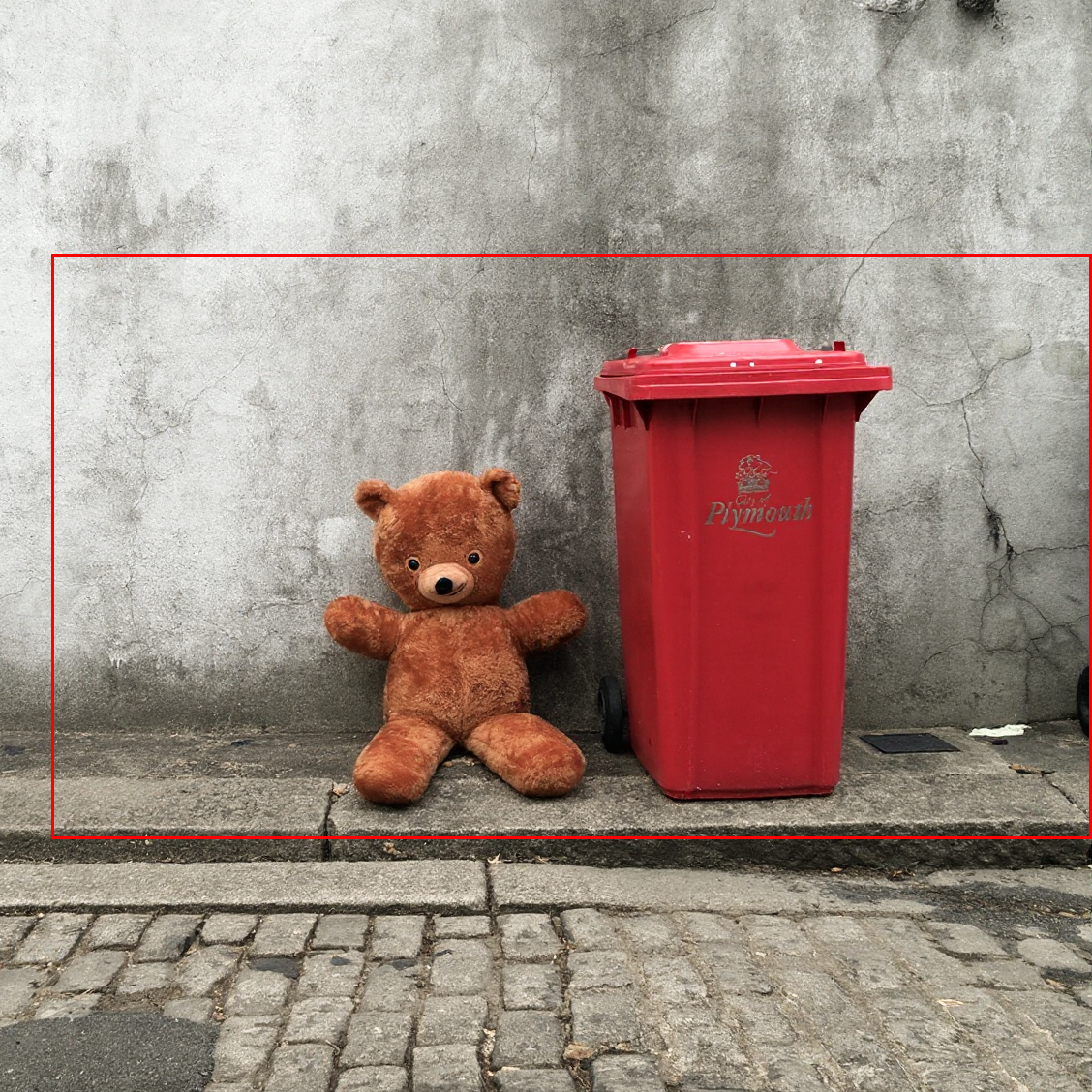}
            \caption*{\textbf{Ours}}
        \end{subfigure}
        \begin{subfigure}{0.69\textwidth}
            \centering
            \begin{subfigure}{0.32\textwidth}
                \includegraphics[width=\linewidth]{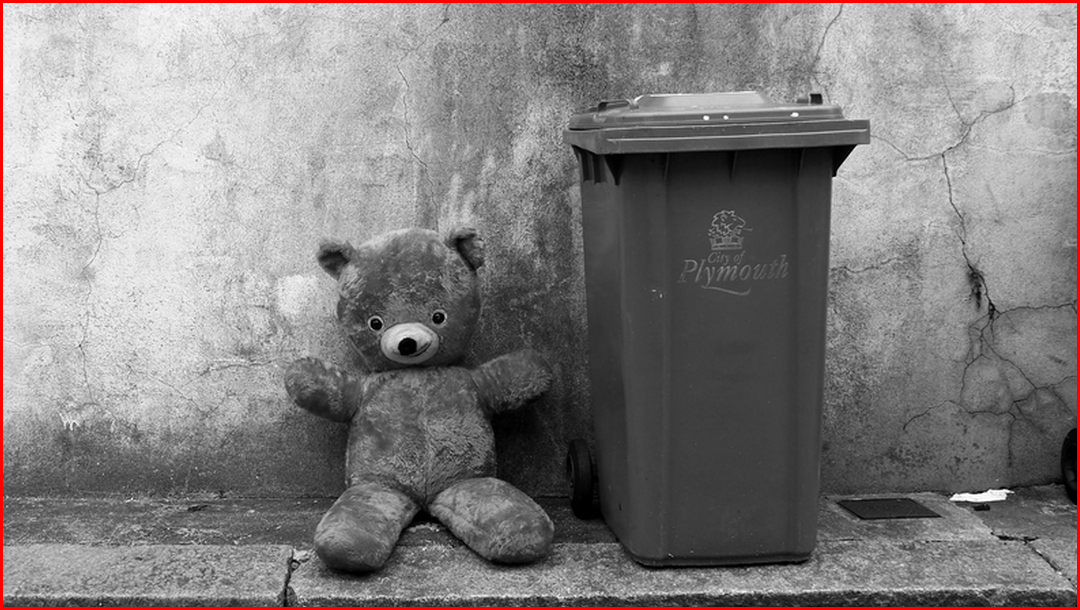}
                \caption*{LQ Input}
            \end{subfigure}
            \begin{subfigure}{0.32\textwidth}
                \includegraphics[width=\linewidth]{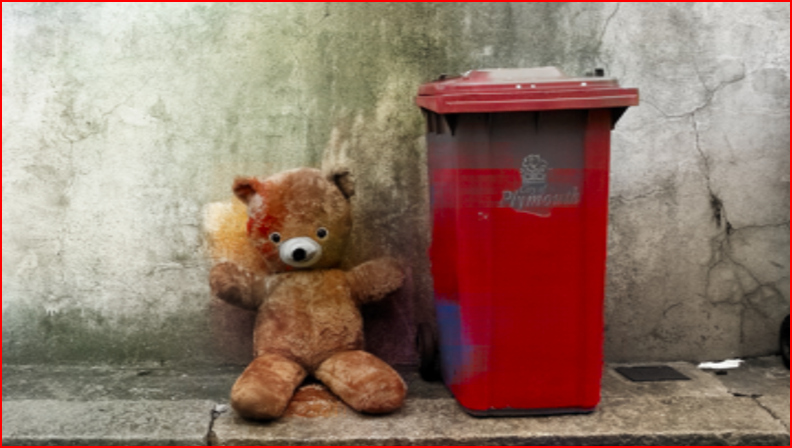}
                \caption*{InstColor}
            \end{subfigure}
            \begin{subfigure}{0.32\textwidth}
                \includegraphics[width=\linewidth]{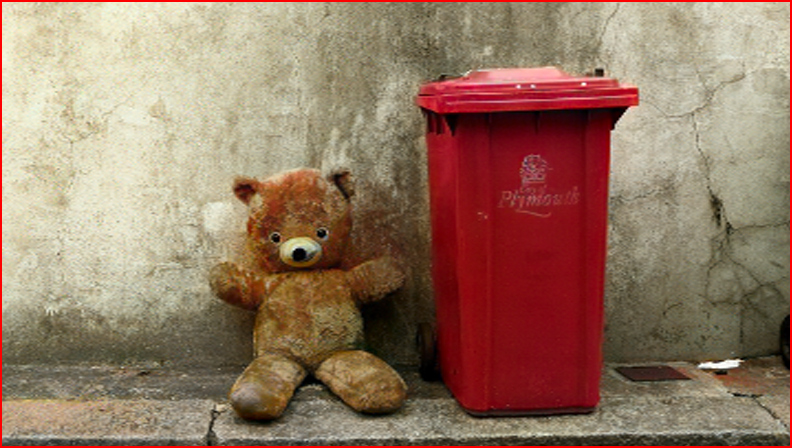}
                \caption*{ColorFormer}
            \end{subfigure}
            \\
            \begin{subfigure}{0.32\textwidth}
                \includegraphics[width=\linewidth]{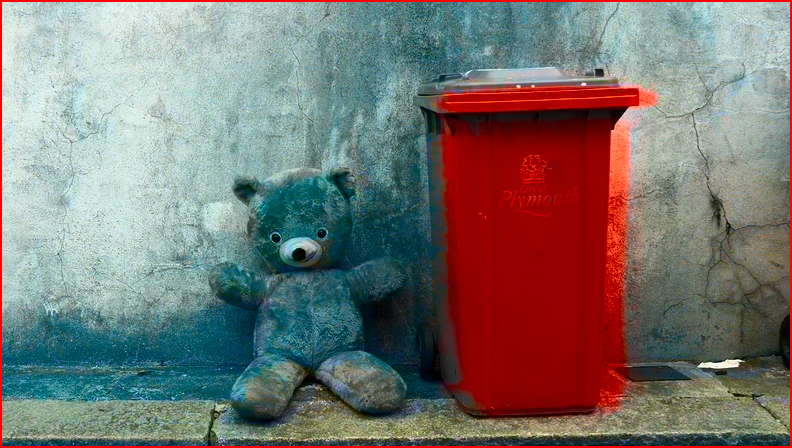}
                \caption*{BigColor}
            \end{subfigure}
            \begin{subfigure}{0.32\textwidth}
                \includegraphics[width=\linewidth]{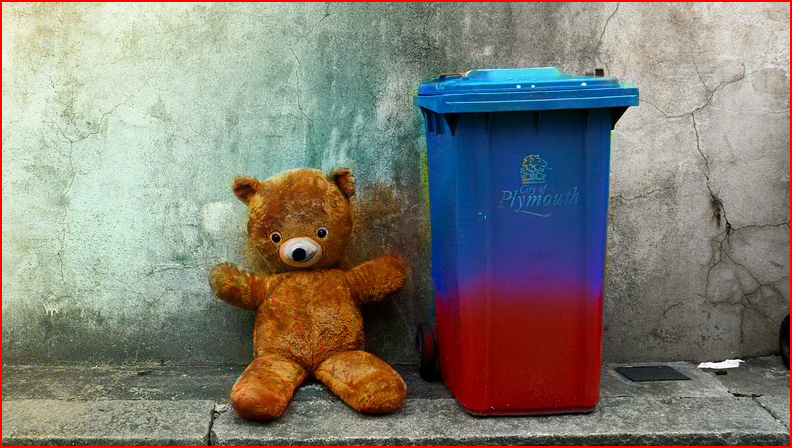}
                \caption*{DDColor}
            \end{subfigure}
            \begin{subfigure}{0.32\textwidth}
                \includegraphics[width=\linewidth]{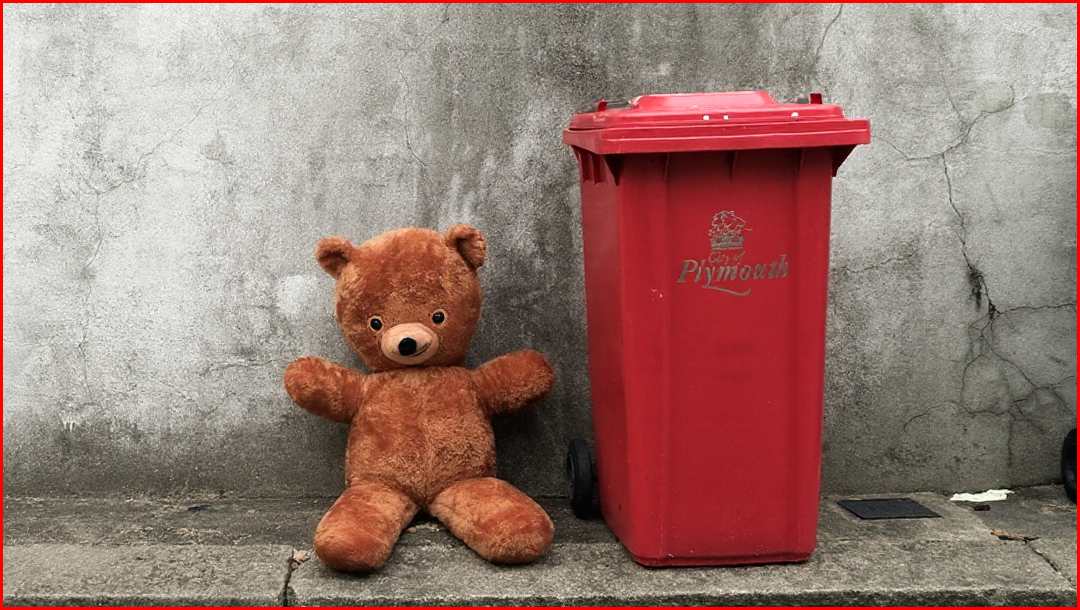}
                \caption*{\textbf{Ours}}
            \end{subfigure}
        \end{subfigure}
    \end{subfigure}

    \begin{subfigure}{\textwidth}
        \centering
        \begin{subfigure}{0.3\textwidth}
            \includegraphics[width=\linewidth]{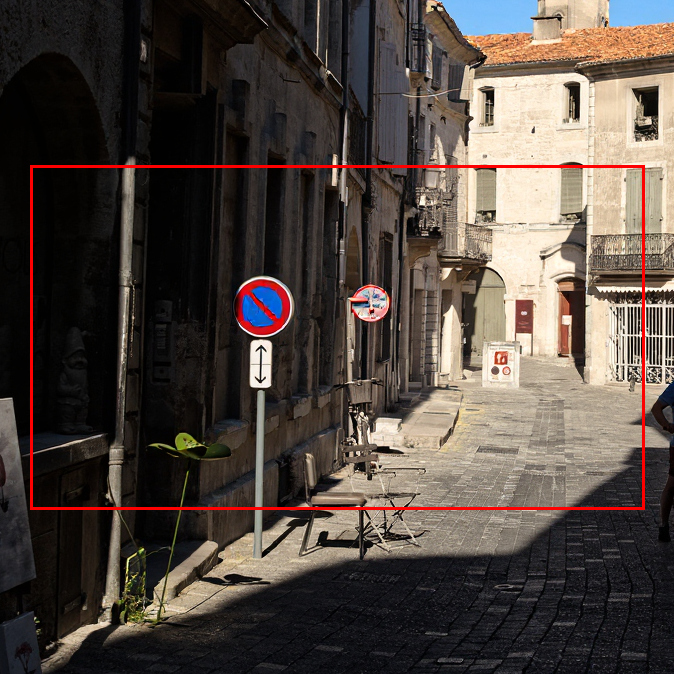}
            \caption*{\textbf{Ours}}
        \end{subfigure}
        \begin{subfigure}{0.69\textwidth}
            \centering
            \begin{subfigure}{0.32\textwidth}
                \includegraphics[width=\linewidth]{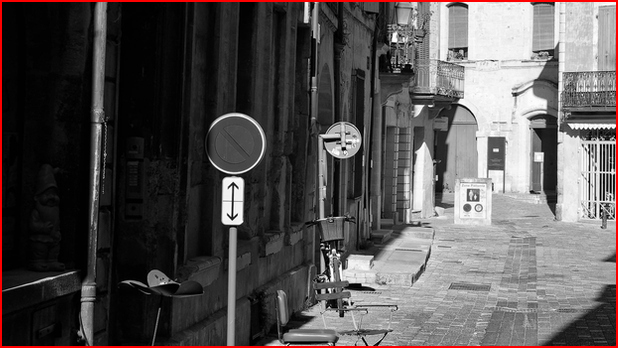}
                \caption*{LQ Input}
            \end{subfigure}
            \begin{subfigure}{0.32\textwidth}
                \includegraphics[width=\linewidth]{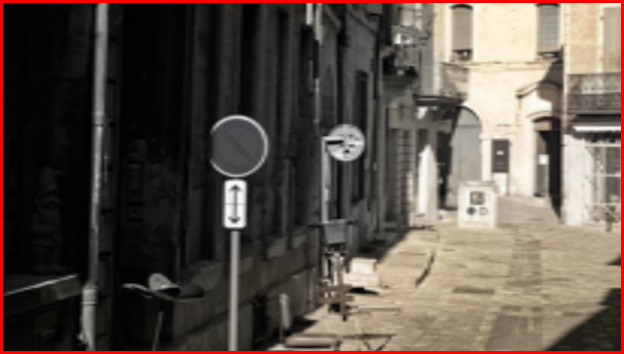}
                \caption*{InstColor}
            \end{subfigure}
            \begin{subfigure}{0.32\textwidth}
                \includegraphics[width=\linewidth]{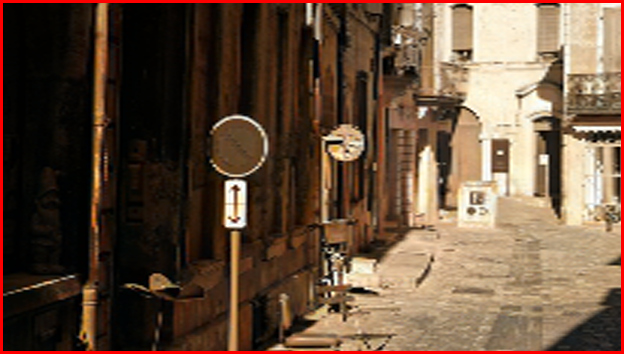}
                \caption*{ColorFormer}
            \end{subfigure}
            \\
            \begin{subfigure}{0.32\textwidth}
                \includegraphics[width=\linewidth]{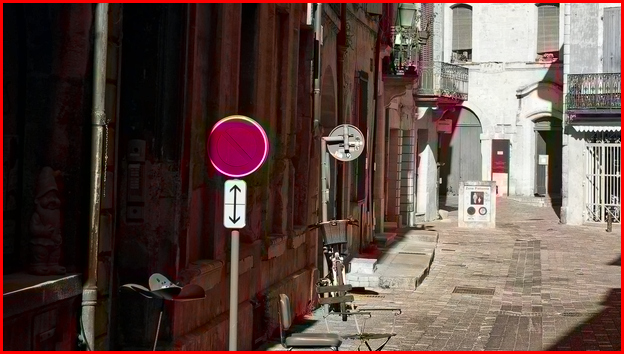}
                \caption*{BigColor}
            \end{subfigure}
            \begin{subfigure}{0.32\textwidth}
                \includegraphics[width=\linewidth]{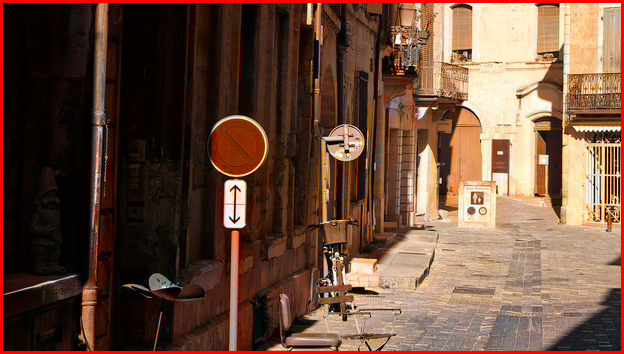}
                \caption*{DDColor}
            \end{subfigure}
            \begin{subfigure}{0.32\textwidth}
                \includegraphics[width=\linewidth]{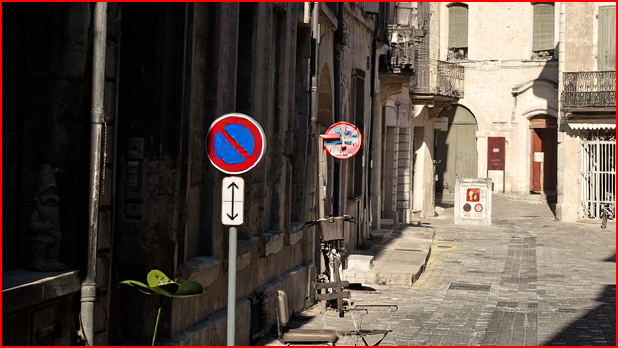}
                \caption*{\textbf{Ours}}
            \end{subfigure}
        \end{subfigure}
    \end{subfigure}

    \caption{\textbf{Qualitative comparisons on LSDIR (colorization).}
    Left: our full-resolution result with crop locations marked (red boxes).
    Right: grayscale input and outputs from InstColor, ColorFormer, BigColor, DDColor, and ours on the corresponding crops.
}
    \label{app:ColorLSDIR}
\end{figure}

\end{document}

%% file: math_commands.tex
\usepackage{amsmath,amsfonts,bm}

\def\eqref#1{equation~\ref{#1}}

\def\1{\bm{1}}

\DeclareMathAlphabet{\mathsfit}{\encodingdefault}{\sfdefault}{m}{sl}
\SetMathAlphabet{\mathsfit}{bold}{\encodingdefault}{\sfdefault}{bx}{n}













%% file: main.bbl
\begin{thebibliography}{42}
\providecommand{\natexlab}[1]{#1}
\providecommand{\url}[1]{\texttt{#1}}
\expandafter\ifx\csname urlstyle\endcsname\relax
  \providecommand{\doi}[1]{doi: #1}\else
  \providecommand{\doi}{doi: \begingroup \urlstyle{rm}\Url}\fi

\bibitem[Agustsson \& Timofte(2017)Agustsson and Timofte]{DIV2K}
Eirikur Agustsson and Radu Timofte.
\newblock Ntire 2017 challenge on single image super-resolution: Dataset and
  study.
\newblock In \emph{Proceedings of the IEEE/CVF Conference on Computer Vision
  and Pattern Recognition Workshops}, 2017.

\bibitem[Ai et~al.(2024)Ai, Zhou, Huang, Han, Chen, You, and Yang]{DreamClear}
Yuang Ai, Xiaoqiang Zhou, Huaibo Huang, Xiaotian Han, Zhengyu Chen, Quanzeng
  You, and Hongxia Yang.
\newblock Dreamclear: High-capacity real-world image restoration with
  privacy-safe dataset curation.
\newblock \emph{Advances in Neural Information Processing Systems}, pp.\
  55443--55469, 2024.

\bibitem[Blau \& Michaeli(2018)Blau and Michaeli]{Blau_2018_CVPR}
Yochai Blau and Tomer Michaeli.
\newblock The perception-distortion tradeoff.
\newblock In \emph{Proceedings of the IEEE/CVF Conference on Computer Vision
  and Pattern Recognition}, 2018.

\bibitem[Chen et~al.(2025)Chen, Pan, and Dong]{FaithDiff}
Junyang Chen, Jinshan Pan, and Jiangxin Dong.
\newblock Faithdiff: Unleashing diffusion priors for faithful image
  super-resolution.
\newblock In \emph{Proceedings of the IEEE/CVF Conference on Computer Vision
  and Pattern Recognition}, pp.\  28188--28197, 2025.

\bibitem[Deng et~al.(2009)Deng, Dong, Socher, Li, Li, and Fei-Fei]{ImageNet}
Jia Deng, Wei Dong, Richard Socher, Li-Jia Li, Kai Li, and Li~Fei-Fei.
\newblock Imagenet: A large-scale hierarchical image database.
\newblock In \emph{Proceedings of the IEEE/CVF Conference on Computer Vision
  and Pattern Recognition}, pp.\  248--255, 2009.

\bibitem[Deng et~al.(2025)Deng, Wu, Yang, Zhu, Wang, and Wu]{FluxIR}
Junyuan Deng, Xinyi Wu, Yongxing Yang, Congchao Zhu, Song Wang, and Zhenyao Wu.
\newblock Acquire and then adapt: Squeezing out text-to-image model for image
  restoration.
\newblock In \emph{Proceedings of the IEEE/CVF Conference on Computer Vision
  and Pattern Recognition}, pp.\  23195--23206, 2025.

\bibitem[Esser et~al.(2024{\natexlab{a}})Esser, Kulal, Blattmann, Entezari,
  M{\"u}ller, Saini, Levi, Lorenz, Sauer, Boesel, et~al.]{SD3}
Patrick Esser, Sumith Kulal, Andreas Blattmann, Rahim Entezari, Jonas
  M{\"u}ller, Harry Saini, Yam Levi, Dominik Lorenz, Axel Sauer, Frederic
  Boesel, et~al.
\newblock Scaling rectified flow transformers for high-resolution image
  synthesis.
\newblock In \emph{Proceedings of the International Conference on Machine
  Learning}, 2024{\natexlab{a}}.

\bibitem[Esser et~al.(2024{\natexlab{b}})Esser, Kulal, Blattmann, Entezari,
  M{\"u}ller, Saini, Levi, Lorenz, Sauer, Boesel, et~al.]{esser2024scaling}
Patrick Esser, Sumith Kulal, Andreas Blattmann, Rahim Entezari, Jonas
  M{\"u}ller, Harry Saini, Yam Levi, Dominik Lorenz, Axel Sauer, Frederic
  Boesel, et~al.
\newblock Scaling rectified flow transformers for high-resolution image
  synthesis.
\newblock In \emph{Proceedings of the International Conference on Machine
  Learning}, 2024{\natexlab{b}}.

\bibitem[Hasler \& Suesstrunk(2003)Hasler and Suesstrunk]{CF}
David Hasler and Sabine~E. Suesstrunk.
\newblock Measuring colorfulness in natural images.
\newblock In \emph{Human Vision and Electronic Imaging VIII}, volume 5007, pp.\
   87--95, 2003.

\bibitem[Ho et~al.(2020)Ho, Jain, and Abbeel]{DDPM}
Jonathan Ho, Ajay Jain, and Pieter Abbeel.
\newblock Denoising diffusion probabilistic models.
\newblock In \emph{Advances in Neural Information Processing Systems}, pp.\
  6840--6851, 2020.

\bibitem[Hu et~al.(2022)Hu, Shen, Wallis, Allen-Zhu, Li, Wang, Wang, Chen,
  et~al.]{Lora}
Edward~J. Hu, Yelong Shen, Phillip Wallis, Zeyuan Allen-Zhu, Yuanzhi Li, Shean
  Wang, Lu~Wang, Weizhu Chen, et~al.
\newblock Lora: Low-rank adaptation of large language models.
\newblock In \emph{Proceedings of the International Conference on Learning
  Representations}, 2022.

\bibitem[Huang et~al.(2024)Huang, Wang, Wu, Shi, Dou, Liang, Feng, Liu, and
  Zhou]{ICLoRA}
Lianghua Huang, Wei Wang, Zhi-Fan Wu, Yupeng Shi, Huanzhang Dou, Chen Liang,
  Yutong Feng, Yu~Liu, and Jingren Zhou.
\newblock In-context lora for diffusion transformers.
\newblock \emph{arXiv preprint arXiv:2410.23775}, 2024.

\bibitem[{Hugging Face Optimum Team}(2025)]{quanto}
{Hugging Face Optimum Team}.
\newblock Optimum-quanto: A {PyTorch} quantization backend for {Optimum}.
\newblock \url{https://github.com/huggingface/optimum-quanto}, 2025.
\newblock GitHub repository; version: v0.2.7; Accessed: 2025-08-30.

\bibitem[Ji et~al.(2022)Ji, Jiang, Luo, Tao, Chu, Xie, Wang, and
  Tai]{ColorFormer}
Xiaozhong Ji, Boyuan Jiang, Donghao Luo, Guangpin Tao, Wenqing Chu, Zhifeng
  Xie, Chengjie Wang, and Ying Tai.
\newblock Colorformer: Image colorization via color memory assisted
  hybrid-attention transformer.
\newblock In \emph{Proceedings of the European Conference on Computer Vision},
  pp.\  20--36, 2022.

\bibitem[Kang et~al.(2023)Kang, Yang, Ouyang, Ren, Li, and Xie]{DDColor}
Xiaoyang Kang, Tao Yang, Wenqi Ouyang, Peiran Ren, Lingzhi Li, and Xuansong
  Xie.
\newblock Ddcolor: Towards photo-realistic image colorization via dual
  decoders.
\newblock In \emph{Proceedings of the IEEE/CVF International Conference on
  Computer Vision}, pp.\  328--338, 2023.

\bibitem[Karras et~al.(2019)Karras, Laine, and Aila]{FFHQ}
Tero Karras, Samuli Laine, and Timo Aila.
\newblock A style-based generator architecture for generative adversarial
  networks.
\newblock In \emph{Proceedings of the IEEE/CVF Conference on Computer Vision
  and Pattern Recognition}, 2019.

\bibitem[Ke et~al.(2021)Ke, Wang, Wang, Milanfar, and Yang]{MUSIQ}
Junjie Ke, Qifei Wang, Yilin Wang, Peyman Milanfar, and Feng Yang.
\newblock Musiq: Multi-scale image quality transformer.
\newblock In \emph{Proceedings of the IEEE/CVF International Conference on
  Computer Vision}, pp.\  5148--5157, 2021.

\bibitem[Kim et~al.(2022)Kim, Kang, Kim, Lee, Kim, Kim, Baek, and
  Cho]{BigColor}
Geonung Kim, Kyoungkook Kang, Seongtae Kim, Hwayoon Lee, Sehoon Kim, Jonghyun
  Kim, Seung-Hwan Baek, and Sunghyun Cho.
\newblock Bigcolor: Colorization using a generative color prior for natural
  images.
\newblock In \emph{Proceedings of the European Conference on Computer Vision},
  pp.\  350--366, 2022.

\bibitem[Labs(2024{\natexlab{a}})]{Flux}
Black~Forest Labs.
\newblock Flux.
\newblock \url{https://github.com/black-forest-labs/flux}, 2024{\natexlab{a}}.

\bibitem[Labs(2024{\natexlab{b}})]{flux1_redux_dev}
Black~Forest Labs.
\newblock Redux.
\newblock \url{https://huggingface.co/black-forest-labs/FLUX.1-Redux-dev},
  2024{\natexlab{b}}.

\bibitem[Labs et~al.(2025)Labs, Batifol, Blattmann, Boesel, Consul, Diagne,
  Dockhorn, English, English, Esser, et~al.]{Kontext}
Black~Forest Labs, Stephen Batifol, Andreas Blattmann, Frederic Boesel, Saksham
  Consul, Cyril Diagne, Tim Dockhorn, Jack English, Zion English, Patrick
  Esser, et~al.
\newblock Flux.1 kontext: Flow matching for in-context image generation and
  editing in latent space.
\newblock \emph{arXiv preprint arXiv:2506.15742}, 2025.

\bibitem[Li et~al.(2023)Li, Zhang, Liang, Cao, Liu, Gong, Zhang, Tang, Liu,
  Demandolx, Ranjan, Timofte, and Van~Gool]{LSDIR}
Yawei Li, Kai Zhang, Jingyun Liang, Jiezhang Cao, Ce~Liu, Rui Gong, Yulun
  Zhang, Hao Tang, Yun Liu, Denis Demandolx, Rakesh Ranjan, Radu Timofte, and
  Luc Van~Gool.
\newblock Lsdir: A large scale dataset for image restoration.
\newblock In \emph{Proceedings of the IEEE/CVF Conference on Computer Vision
  and Pattern Recognition Workshops}, pp.\  1775--1787, 2023.

\bibitem[Lin et~al.(2024)Lin, He, Chen, Lyu, Dai, Yu, Qiao, Ouyang, and
  Dong]{DiffBIR}
Xinqi Lin, Jingwen He, Ziyan Chen, Zhaoyang Lyu, Bo~Dai, Fanghua Yu, Yu~Qiao,
  Wanli Ouyang, and Chao Dong.
\newblock Diffbir: Toward blind image restoration with generative diffusion
  prior.
\newblock In \emph{Proceedings of the European Conference on Computer Vision},
  pp.\  430--448, 2024.

\bibitem[Lipman et~al.(2022)Lipman, Chen, Ben-Hamu, Nickel, and
  Le]{lipman2022flow}
Yaron Lipman, Ricky T.~Q. Chen, Heli Ben-Hamu, Maximilian Nickel, and Matt Le.
\newblock Flow matching for generative modeling.
\newblock \emph{arXiv preprint arXiv:2210.02747}, 2022.

\bibitem[Liu et~al.(2023)Liu, Li, Wu, and Lee]{LLAVA}
Haotian Liu, Chunyuan Li, Qingyang Wu, and Yong~Jae Lee.
\newblock Visual instruction tuning.
\newblock In \emph{Advances in Neural Information Processing Systems},
  volume~36, pp.\  34892--34916, 2023.

\bibitem[Liu et~al.(2022)Liu, Gong, and Liu]{liu2022flow}
Xingchao Liu, Chengyue Gong, and Qiang Liu.
\newblock Flow straight and fast: Learning to generate and transfer data with
  rectified flow.
\newblock \emph{arXiv preprint arXiv:2209.03003}, 2022.

\bibitem[Peebles \& Xie(2023)Peebles and Xie]{DiT}
William Peebles and Saining Xie.
\newblock Scalable diffusion models with transformers.
\newblock In \emph{Proceedings of the IEEE/CVF International Conference on
  Computer Vision}, pp.\  4195--4205, 2023.

\bibitem[Podell et~al.(2023)Podell, English, Lacey, Blattmann, Dockhorn,
  M{\"u}ller, Penna, and Rombach]{SDXL}
Dustin Podell, Zion English, Kyle Lacey, Andreas Blattmann, Tim Dockhorn, Jonas
  M{\"u}ller, Joe Penna, and Robin Rombach.
\newblock Sdxl: Improving latent diffusion models for high-resolution image
  synthesis.
\newblock \emph{arXiv preprint arXiv:2307.01952}, 2023.

\bibitem[Rombach et~al.(2022)Rombach, Blattmann, Lorenz, Esser, and Ommer]{SD}
Robin Rombach, Andreas Blattmann, Dominik Lorenz, Patrick Esser, and Bj{\"o}rn
  Ommer.
\newblock High-resolution image synthesis with latent diffusion models.
\newblock In \emph{Proceedings of the IEEE/CVF Conference on Computer Vision
  and Pattern Recognition}, pp.\  10684--10695, 2022.

\bibitem[Ronneberger et~al.(2015)Ronneberger, Fischer, and Brox]{UNet}
Olaf Ronneberger, Philipp Fischer, and Thomas Brox.
\newblock U-net: Convolutional networks for biomedical image segmentation.
\newblock In \emph{International Conference on Medical Image Computing and
  Computer-Assisted Intervention}, pp.\  234--241. Springer, 2015.

\bibitem[Saharia et~al.(2023)Saharia, Ho, Chan, Salimans, Fleet, and
  Norouzi]{SR3}
Chitwan Saharia, Jonathan Ho, William Chan, Tim Salimans, David~J. Fleet, and
  Mohammad Norouzi.
\newblock Image super-resolution via iterative refinement.
\newblock \emph{IEEE Transactions on Pattern Analysis and Machine
  Intelligence}, 45\penalty0 (4):\penalty0 4713--4726, 2023.

\bibitem[Su et~al.(2020)Su, Chu, and Huang]{InstColor}
Jheng-Wei Su, Hung-Kuo Chu, and Jia-Bin Huang.
\newblock Instance-aware image colorization.
\newblock In \emph{Proceedings of the IEEE/CVF Conference on Computer Vision
  and Pattern Recognition}, 2020.

\bibitem[Su et~al.(2024)Su, Ahmed, Lu, Pan, Bo, and Liu]{su2024roformer}
Jianlin Su, Murtadha Ahmed, Yu~Lu, Shengfeng Pan, Wen Bo, and Yunfeng Liu.
\newblock Roformer: Enhanced transformer with rotary position embedding.
\newblock \emph{Neurocomputing}, 568:\penalty0 127063, 2024.

\bibitem[Wang et~al.(2023)Wang, Chan, and Loy]{CLIPIQA}
Jianyi Wang, Kelvin C.~K. Chan, and Chen~Change Loy.
\newblock Exploring {CLIP} for assessing the look and feel of images.
\newblock In \emph{Proceedings of the AAAI Conference on Artificial
  Intelligence}, volume~37, pp.\  2555--2563, 2023.

\bibitem[Wang et~al.(2024)Wang, Yue, Zhou, Chan, and Loy]{StableSR}
Jianyi Wang, Zongsheng Yue, Shangchen Zhou, Kelvin C.~K. Chan, and Chen~Change
  Loy.
\newblock Exploiting diffusion prior for real-world image super-resolution.
\newblock \emph{International Journal of Computer Vision}, 132\penalty0
  (12):\penalty0 5929--5949, 2024.

\bibitem[Wang et~al.(2021)Wang, Xie, Dong, and Shan]{wang2021realesrgan}
Xintao Wang, Liangbin Xie, Chao Dong, and Ying Shan.
\newblock Real-{ESRGAN}: Training real-world blind super-resolution with pure
  synthetic data.
\newblock In \emph{Proceedings of the IEEE/CVF International Conference on
  Computer Vision}, pp.\  1905--1914, 2021.

\bibitem[Wu et~al.(2024)Wu, Yang, Sun, Zhang, Li, and Zhang]{SeeSR}
Rongyuan Wu, Tao Yang, Lingchen Sun, Zhengqiang Zhang, Shuai Li, and Lei Zhang.
\newblock Seesr: Towards semantics-aware real-world image super-resolution.
\newblock In \emph{Proceedings of the IEEE/CVF Conference on Computer Vision
  and Pattern Recognition}, pp.\  25456--25467, 2024.

\bibitem[Yang et~al.(2022)Yang, Wu, Shi, Lao, Gong, Cao, Wang, and
  Yang]{MANIQA}
Sidi Yang, Tianhe Wu, Shuwei Shi, Shanshan Lao, Yuan Gong, Mingdeng Cao, Jiahao
  Wang, and Yujiu Yang.
\newblock Maniqa: Multi-dimension attention network for no-reference image
  quality assessment.
\newblock In \emph{Proceedings of the IEEE/CVF Conference on Computer Vision
  and Pattern Recognition Workshops}, pp.\  1191--1200, 2022.

\bibitem[Yi \& Yu(2026)Yi and Yu]{Yi2026Fill2SR}
Xingfu Yi and Xiaoxue Yu.
\newblock {Fill2SR}: Repurposing inpainting diffusion transformers for
  real-world super-resolution.
\newblock In \emph{Computer Vision -- ECCV 2026}, pp.\  441--457. Springer
  Nature Switzerland, 2026.
\newblock \doi{10.1007/978-3-032-37556-8_24}.
\newblock URL \url{https://doi.org/10.1007/978-3-032-37556-8_24}.

\bibitem[Yu et~al.(2024)Yu, Gu, Li, Hu, Kong, Wang, He, Qiao, and Dong]{SUPIR}
Fanghua Yu, Jinjin Gu, Zheyuan Li, Jinfan Hu, Xiangtao Kong, Xintao Wang,
  Jingwen He, Yu~Qiao, and Chao Dong.
\newblock Scaling up to excellence: Practicing model scaling for
  photo-realistic image restoration in the wild.
\newblock In \emph{Proceedings of the IEEE/CVF Conference on Computer Vision
  and Pattern Recognition}, pp.\  25669--25680, 2024.

\bibitem[Zhang et~al.(2021)Zhang, Liang, Van~Gool, and
  Timofte]{zhang2021bsrgan}
Kai Zhang, Jingyun Liang, Luc Van~Gool, and Radu Timofte.
\newblock Designing a practical degradation model for deep blind image
  super-resolution.
\newblock In \emph{Proceedings of the IEEE/CVF International Conference on
  Computer Vision}, pp.\  4791--4800, 2021.

\bibitem[Zhang et~al.(2023)Zhang, Rao, and Agrawala]{ControlNet}
Lvmin Zhang, Anyi Rao, and Maneesh Agrawala.
\newblock Adding conditional control to text-to-image diffusion models.
\newblock In \emph{Proceedings of the IEEE/CVF International Conference on
  Computer Vision}, pp.\  3836--3847, 2023.

\end{thebibliography}
